\documentclass[runningheads]{llncs}
\usepackage{graphicx}
\pdfoutput=1
\usepackage{tikz}

\usepackage{comment}
\usepackage{amsmath,amssymb} 
\usepackage{color}
\usepackage{makecell}

\usepackage{bm}
\usepackage{epsfig}
\usepackage{amsmath}
\usepackage{amssymb}
\usepackage{multirow}
\usepackage{color}
\usepackage{algorithmicx}
\usepackage{algorithm}
\usepackage{algpseudocode}
\usepackage{array}
\usepackage{caption}
\usepackage{bbding}
\usepackage{colortbl}
\usepackage{threeparttable}
\usepackage{booktabs}
\usepackage{float}
\usepackage{subcaption}
\usepackage{cite}
\usepackage{arydshln}
\usepackage[accsupp]{axessibility}  

\usepackage[pagebackref=true,breaklinks=true,letterpaper=true,colorlinks,bookmarks=false]{hyperref}

\newcommand{\ignore}[1]{}
\def\etal{\textit{et al}.}
\def\ie{\textit{i.e.}}
\def\eg{\textit{e.g.}}

\definecolor{Gray}{gray}{0.85}
\definecolor{White}{gray}{1.0}
\newcolumntype{a}{>{\columncolor{Gray}}c}
\newcolumntype{z}{>{\columncolor{White}}c}

\begin{document}\sloppy
\pagestyle{headings}
\mainmatter

\title{Editing Out-of-domain GAN Inversion via Differential Activations} 



\author{Haorui Song\inst{1} \and
Yong Du\inst{2}\thanks{The first two authors contribute equally.} \and
Tianyi Xiang\inst{1}\and
Junyu Dong\inst{2}\and
Jing Qin\inst{3}\and
Shengfeng He\inst{1}\orcidID{0000-0002-3802-4644}\thanks{Corresponding author (hesfe@scut.edu.cn).}}

\authorrunning{Song et al.}

\institute{South China University of Technology, Guangzhou, China \and
Ocean University of China, Qingdao, China \and
The Hong Kong Polytechnic University, Hong Kong SAR, China
}

\maketitle

\begin{abstract}
Despite the demonstrated editing capacity in the latent space of a pretrained GAN model, inverting real-world images is stuck in a dilemma that the reconstruction cannot be faithful to the original input. The main reason for this is that the distributions between training and real-world data are misaligned, and because of that, it is unstable of GAN inversion for real image editing. In this paper, we propose a novel GAN prior based editing framework to tackle the out-of-domain inversion problem with a composition-decomposition paradigm. In particular, during the phase of composition, we introduce a differential activation module for detecting semantic changes from a global perspective, \ie, the relative gap between the features of edited and unedited images. With the aid of the generated Diff-CAM mask, a coarse reconstruction can intuitively be composited by the paired original and edited images. In this way, the attribute-irrelevant regions can be survived in almost whole, while the quality of such an intermediate result is still limited by an unavoidable ghosting effect. Consequently, in the decomposition phase, we further present a GAN prior based deghosting network for separating the final fine edited image from the coarse reconstruction. Extensive experiments exhibit superiorities over the state-of-the-art methods, in terms of qualitative and quantitative evaluations. The robustness and flexibility of our method is also validated on both scenarios of single attribute and multi-attribute manipulations. Code is available at \url{https://github.com/HaoruiSong622/Editing-Out-of-Domain}.
\end{abstract}

\section{Introduction}\label{sec:intro}
\begin{figure}
	\centering	
	\rotatebox[origin=l]{90}{\hspace{-12mm}Beard \hspace{12mm}Eyebrows}
	\begin{subfigure}{.18\linewidth}
		\centering
		\captionsetup{justification=centering}
		\includegraphics[width=\linewidth]{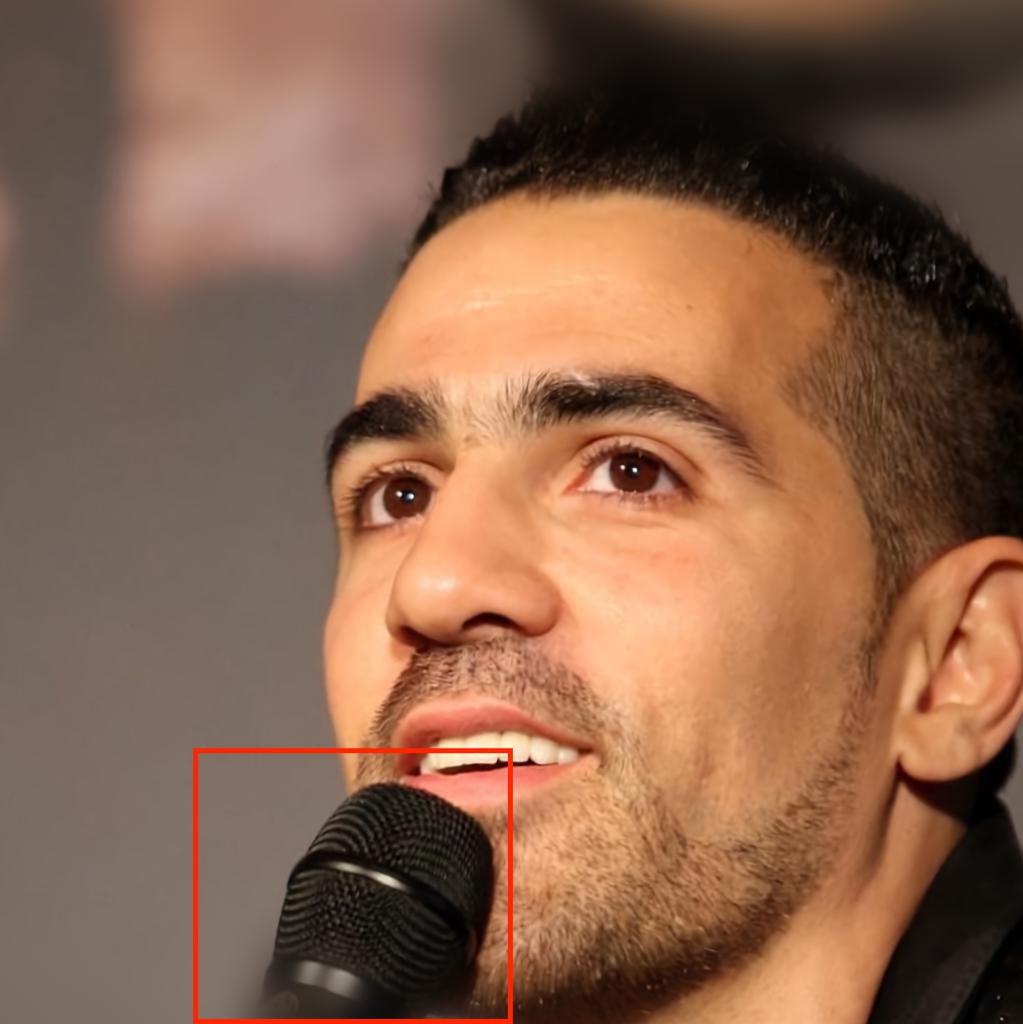}\\
		\vspace{.7mm}
		\includegraphics[width=\linewidth]{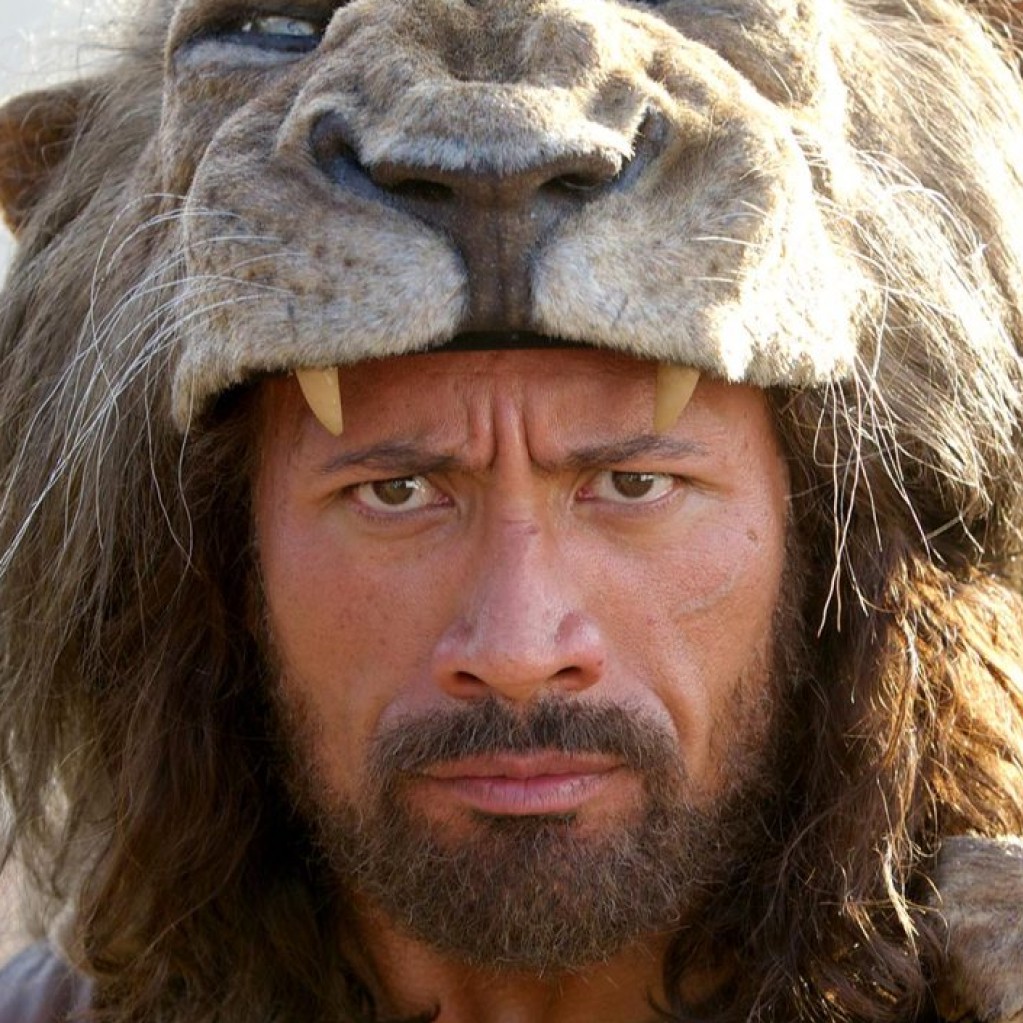}\\
		\caption{Original \protect\\ Image}\label{fig:teaser_a}
	\end{subfigure}
	\begin{subfigure}{.18\linewidth}
		\centering
		\captionsetup{justification=centering}
		\includegraphics[width=\linewidth]{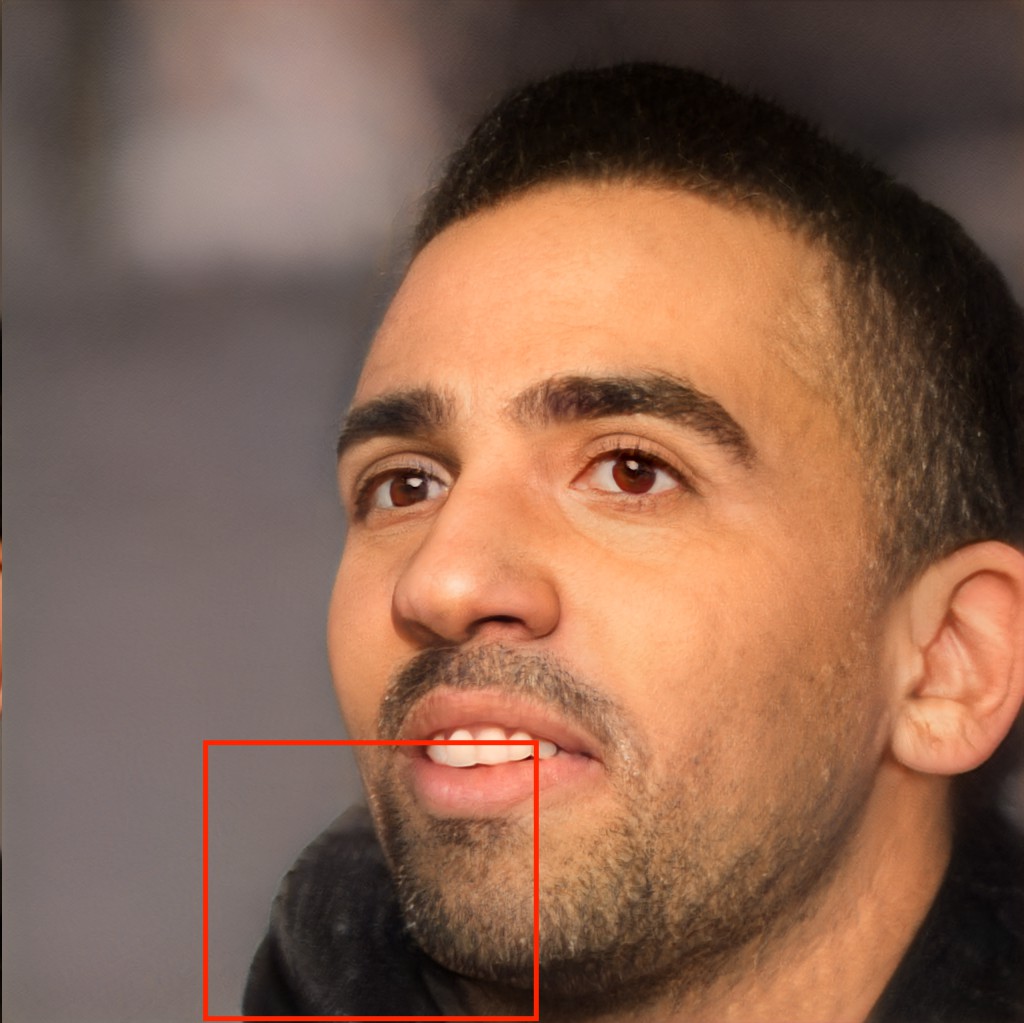}\\
		\vspace{.7mm}
		\includegraphics[width=\linewidth]{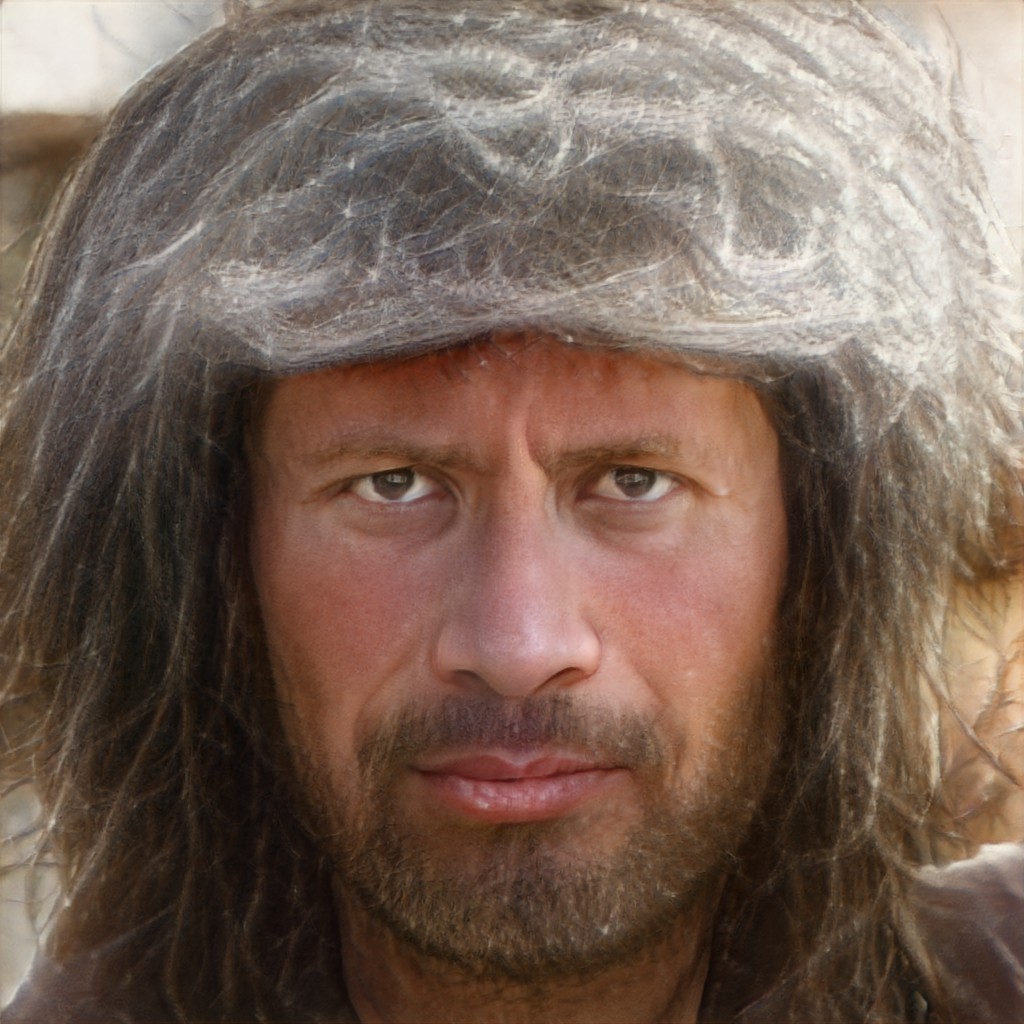}\\
		\caption{GAN \protect\\ Inversion}\label{fig:teaser_b}
	\end{subfigure}
	\begin{subfigure}{.18\linewidth}
		\centering
		\captionsetup{justification=centering}
		\includegraphics[width=\linewidth]{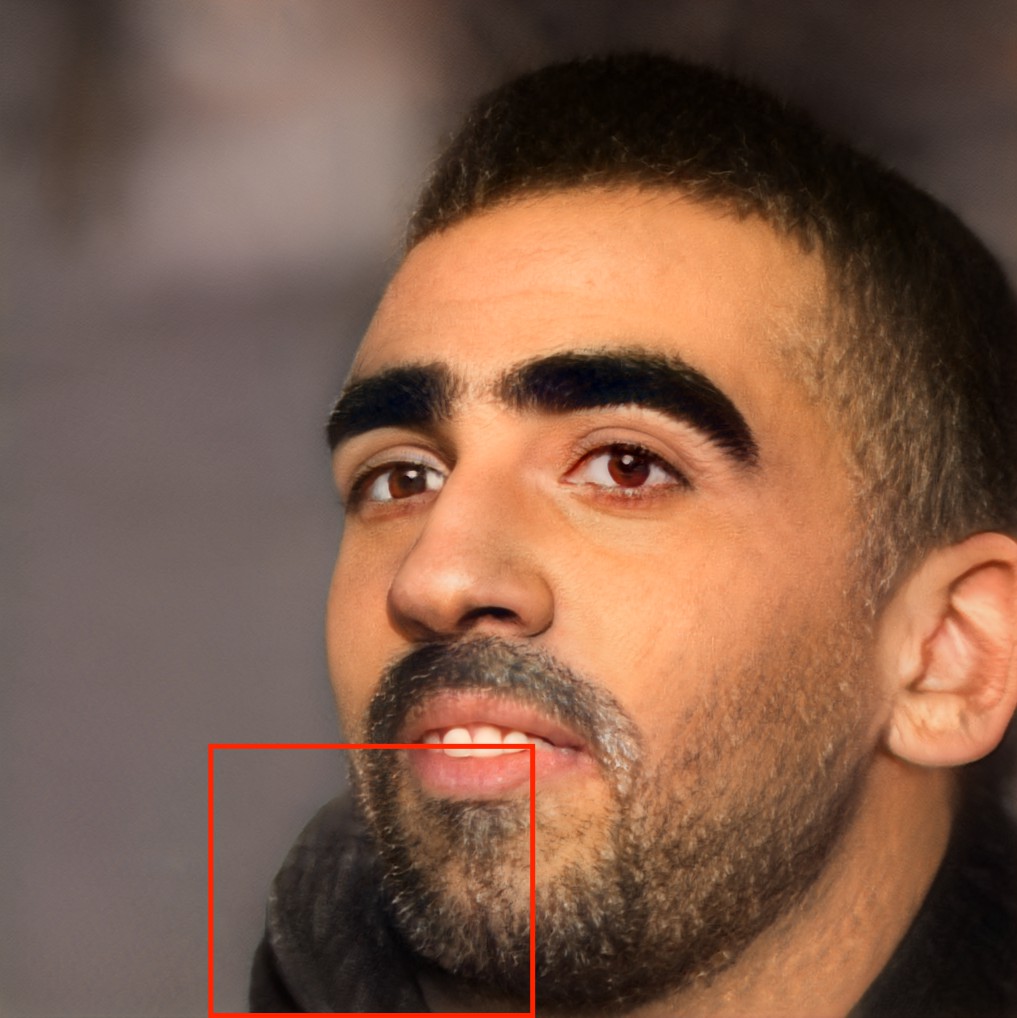}\\
		\vspace{.7mm}
		\includegraphics[width=\linewidth]{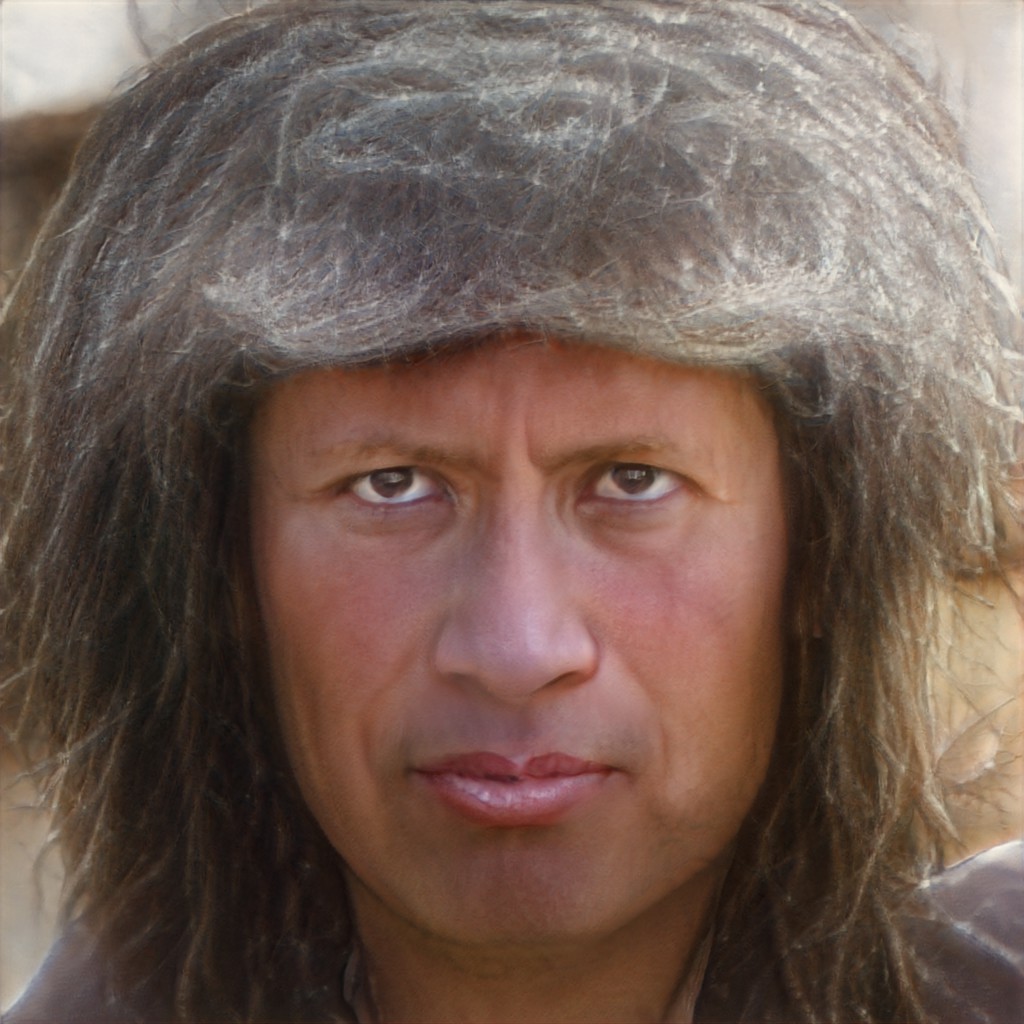}\\
		\caption{Latent \protect\\ Code Editing}\label{fig:teaser_c}
	\end{subfigure}
	\begin{subfigure}{.18\linewidth}
		\centering
		\captionsetup{justification=centering}
		\includegraphics[width=\linewidth]{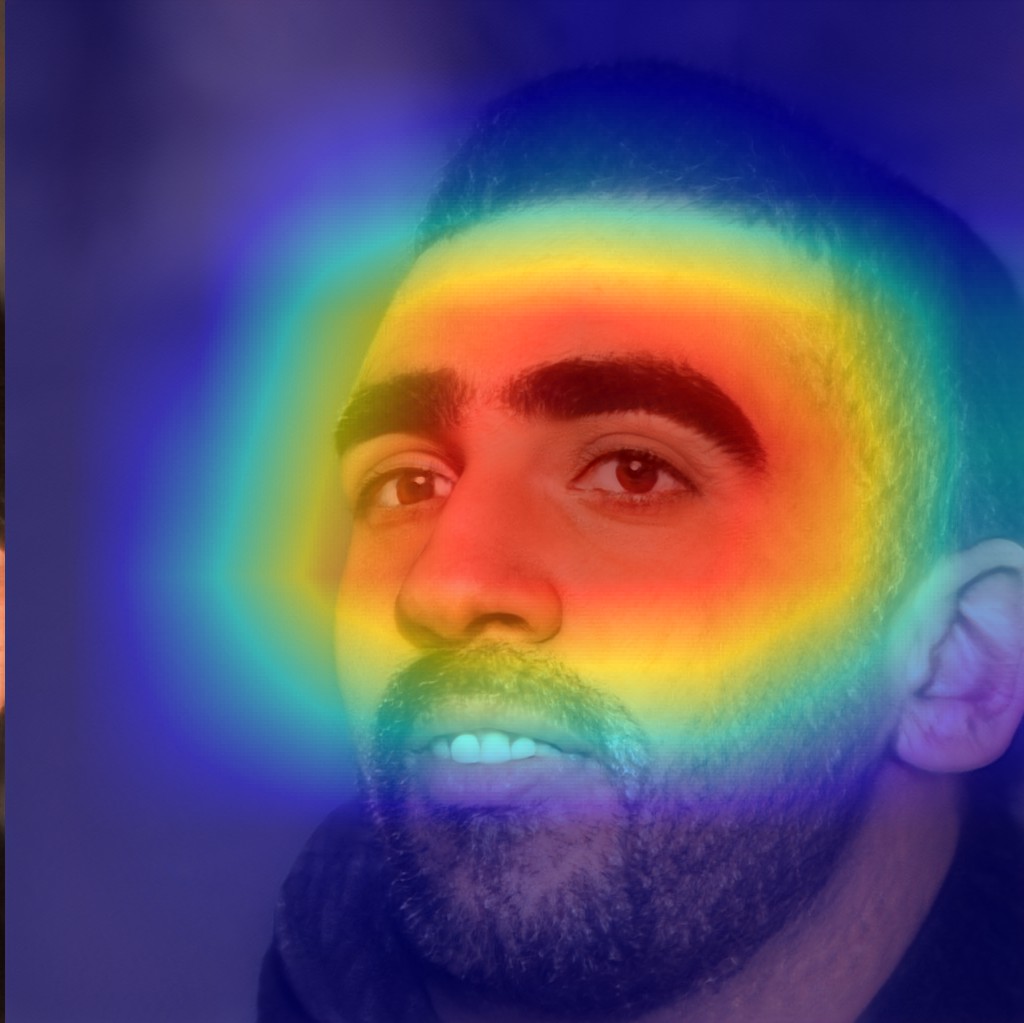}\\
		\vspace{.7mm}
		\includegraphics[width=\linewidth]{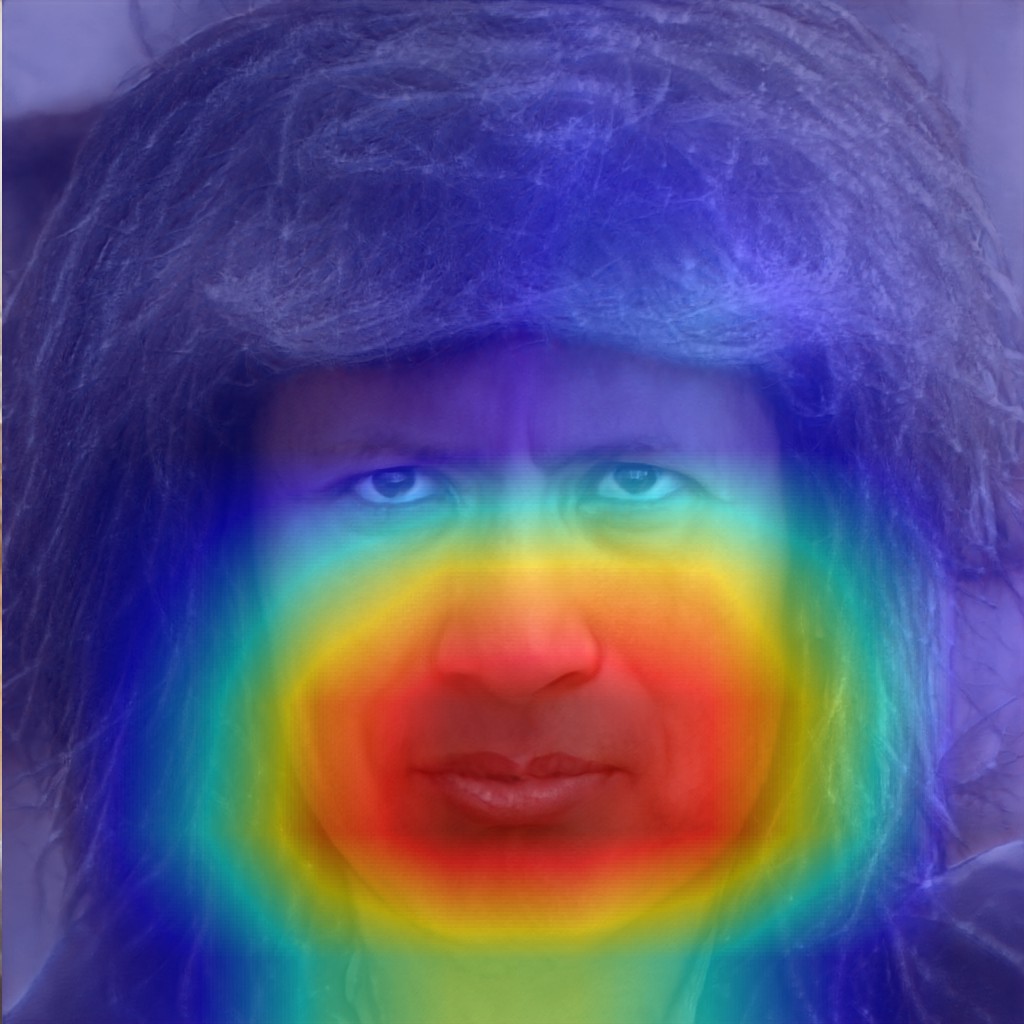}\\
		\caption{Diff-CAM \protect\\ Mask}\label{fig:teaser_d}
	\end{subfigure}
	\begin{subfigure}{.18\linewidth}
		\centering
		\captionsetup{justification=centering}
		\includegraphics[width=\linewidth]{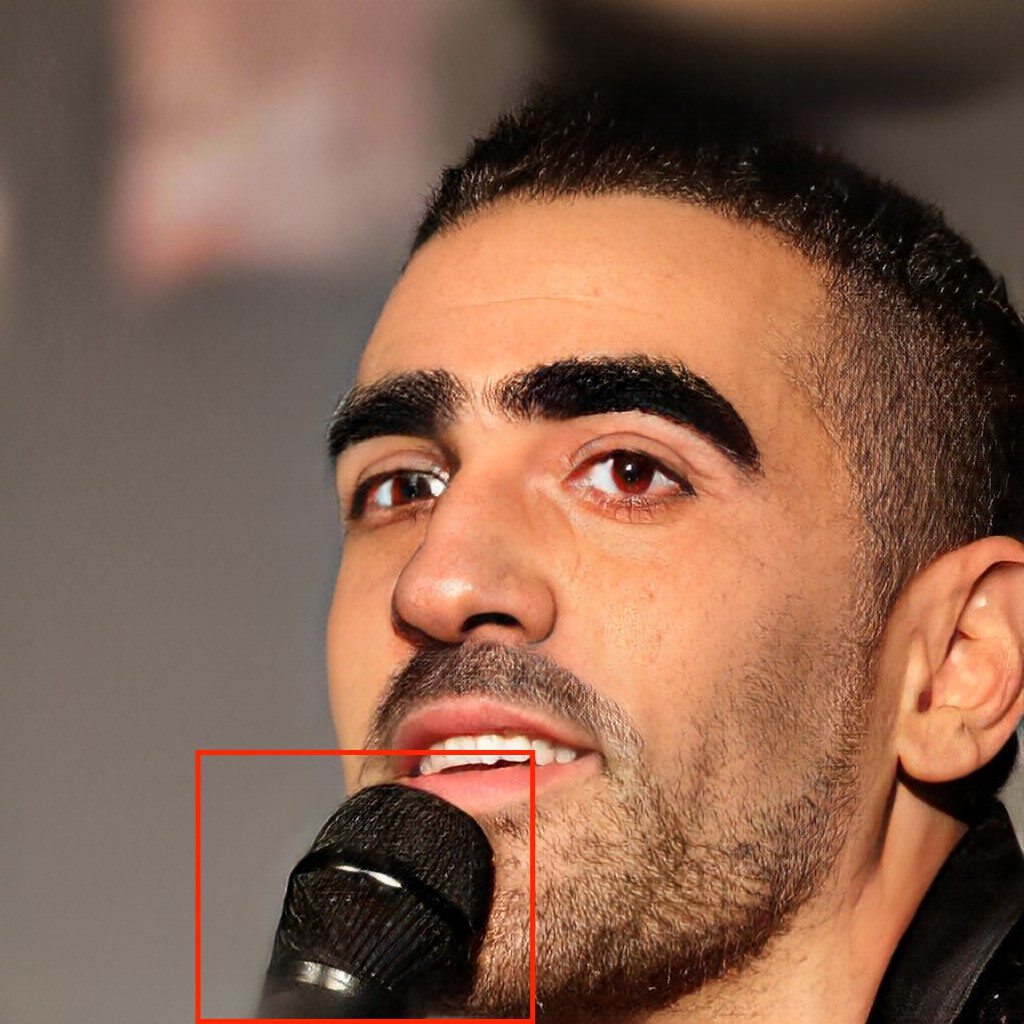}\\
		\vspace{.7mm}
		\includegraphics[width=\linewidth]{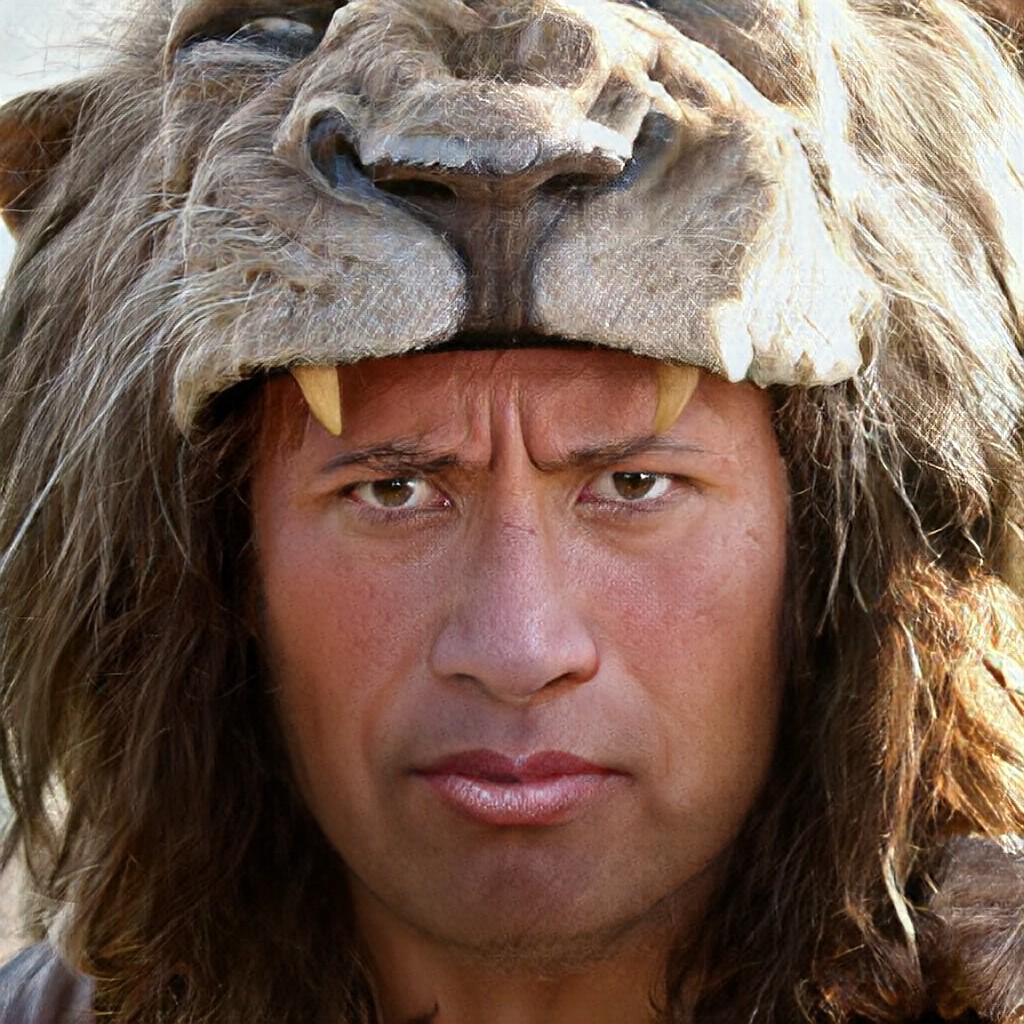}\\
		\caption{Our \protect\\ Result}\label{fig:teaser_e}
	\end{subfigure}
	\caption{We delve deep into the editing problem of out-of-domain GAN inversion. (b) shows that for out-of-domain real images, GAN inversion cannot obtain a faithful reconstruction and therefore produce unacceptable editing (c). Our framework localizes semantic changes with differential activations (d), enabling the preservation of out-of-domain image content (like the lion hat and microphone) while activating the editing ability of GAN priors.}\vspace{-4mm}
	\label{fig:teaser}
\end{figure}

Generative Adversarial Networks (GANs)~\cite{goodfellow2014generative,isola2017image,arjovsky2017wasserstein} have demonstrated impressive image editing capability. From a random noise input, GAN models can encode abundant semantic information and spontaneously excavate interpretable directions in a latent space (\eg, $\cal{W}$ space~\cite{karras2019style}, $\cal{W}^+$ space~\cite{karras2020analyzing} and etc.). By varying the latent codes along the controllable directions, highly realistic images with diverse attributes can be synthesized using GANs. However, such manipulations are applicable only in the latent space. For real images, a mapping function is required to transform the RGB input to a latent code.

GAN inversion~\cite{zhu2020domain,richardson2020encoding} which aims at inverting a given image back into the latent space of a pretrained GAN model such as StyleGAN~\cite{karras2019style}, can enable the corresponding semantic directions to be applicable for real image editing. As a consequence, numerous GAN inversion based image processing frameworks~\cite{Yang_2021_CVPR,abdal2020image2stylegan++,gu2020image, zhou2022pro, zhong2022faithful, xu2022high} have emerged. However, existing inversion methods are stuck in a dilemma that they cannot faithfully invert those images that are not from the distribution of training data. For example, as shown in Fig.~\ref{fig:teaser_b}, both real images are fed into the pSp encoder~\cite{richardson2020encoding} for inversion, and then the codes are sent to a pretrained StyleGAN2~\cite{karras2020analyzing} for generation, but it turns out that the microphone and the lion hat are vanished or distorted. This is due to the misaligned data domains. Such an out-of-domain issue can undoubtedly lead to unstable editing performance and thus severely hinder the practicality of GAN inversion. On the other hand, the powerful attribute-aware manipulation capability of pretrained GANs is indispensable for image editing. These facts motivate us to raise a natural idea: it would be feasible if we can properly integrate the edited region from the corresponding inversion with its unedited counterpart from the original input.

\begin{figure}[t]
	\centering
	\begin{subfigure}[t]{0.23\linewidth}
		\centering
    \includegraphics[width=\linewidth]{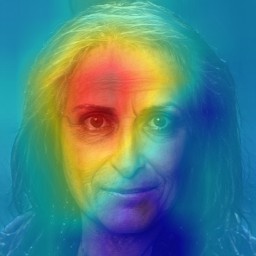}
    \caption{CAM~\cite{zhou2016learning}}\label{fig:mask_a}
  \end{subfigure}
	\begin{subfigure}[t]{0.23\linewidth}
		\centering
    \includegraphics[width=\linewidth]{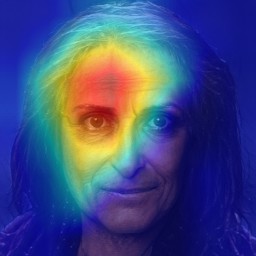}
    \caption{Grad-CAM~\cite{selvaraju2017grad}}\label{fig:mask_b}
  \end{subfigure}
	\begin{subfigure}[t]{0.23\linewidth}
		\centering
    \includegraphics[width=\linewidth]{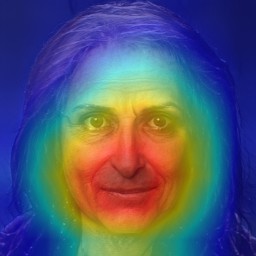}
    \caption{Ours}\label{fig:mask_c}
  \end{subfigure}
	\label{fig:masks}
	\vspace{-3mm}\caption{Activation maps generated by CAM, Grad-CAM and our Diff-CAM models. Our activation map has a broader and more comprehensive coverage.}\vspace{-4mm}
\end{figure}

To achieve this goal, we turn to consider how to detect the edited region in the inversion. This reminds us of class activation mapping (CAM)~\cite{zhou2016learning,selvaraju2017grad}. Such techniques focus on producing an attention map that highlights the regions that contribute to the classification decision, and have been widely used in visual explanations, such as weakly-supervised localization~\cite{bae2020rethinking} and visual question answering~\cite{patro2019u}. Also for image manipulation, Kim \etal~\cite{kim2021not} utilize Grad-CAM~\cite{selvaraju2017grad} to generate a mask for localizing the attribute-relevant regions. A critical problem of CAM-based methods is that, its principle is to locate the activation regions that make the final decision, but making such a decision does not require a comprehensive activation on the attribute-relevant regions (\eg, locating the wrinkle instead of the whole face can classify the ``old'' attribute). As a result, they tend to produce localized activations (see Fig.~\ref{fig:mask_a} and ~\ref{fig:mask_b}). Relying on CAM for editing is apparently not flexible, as the editing of some attributes like ``sex'' may change the entire face, but binarily classifying male or female typically concentrate on facial components. This contradicts with our objective to combine edited region and its unedited counterpart.

In this paper, we propose a novel GAN prior based editing framework to resolve the above problems. Specifically, our editing method is executed in a composition-decomposition manner. In the composition stage, our aim is to generate a coarse reconstruction via combining the edited inversion with the original input, weighting by a Diff-CAM mask which is used for indicating the edited region. In particular, we present a simple yet effective \textit{differential activation} mechanism to track the semantic changes rather than locating classification-relevant regions. It is performed by capturing the variational features between the edited and the original inversions, and we shed light on the differential features that reveal the editing attributes. In this way, the produced mask can specify the range of the edited region more accurately (see in Fig.~\ref{fig:mask_c}), as the semantical differences are explicitly embedded in the hidden responses. While in the decomposition stage, we need to remove the ghosting effect occurred in the coarse result. To deal with it, we further design a deghosting network that reuses the GAN prior, which is used for separating the final fine edited inversion from the coarse reconstruction in a multi-scale aggregation manner. Extensive experiments show that our method is the first feasible real-image editing method that built upon GAN inversion.

In summary, our key contributions are as follows:
\begin{itemize}
	\item We delve deep into the out-of-domain problem existed in GAN inversion, and propose a novel GAN prior based editing framework in a composition-decomposition manner. Our method can use the original input to generate the unedited region, as well as maintaining a high quality of editing.

	\item We tailor a differential activation strategy to track semantic changes before and after editing. This design allows to embed more accurate range of the edited region with neglectable additional computational cost.

	\item We present a deghosting network with hierarchical GAN priors, for effectively alleviating the ghosting effect in the coarse reconstruction.

	\item We outperform state-of-the-art methods in terms of qualitative and quantitative evaluations, and we demonstrate the flexibility and robustness in both scenarios of single attribute and multi-attribute manipulations.
\end{itemize}

\section{Related Work}\label{sec:related}
\vspace{-4mm}\textbf{Non-GAN prior based image manipulation.} Non-GAN prior based methods~\cite{kim2017learning, isola2017image, zhu2017unpaired} usually manipulate attributes of images via an adversarial training process. Kim \etal~\cite{kim2017learning} propose a GAN based framework to discover cross-domain relations. Isola \etal~\cite{isola2017image} propose to use conditional GANs for image-to-image translation. And Zhu \etal~\cite{zhu2017unpaired} propose to translate images across different domains without paired training data. In general, these methods are designed to learn a model that corresponds to a specific translation, which leads to inflexibility in practical applications. To address this problem, StarGAN~\cite{choi2018stargan} is proposed to learn the mapping among multiple domains, using only a single generator and a discriminator. CMP~\cite{kim2021not} proposes to refine image-to-image translation results by introducing a cam-consistency loss to force the network to focus on attribute-relevant regions. Note that all these methods need to train models from scratch, and thus cannot capture GAN priors which is proven to be extremely effective for image manipulation~\cite{karras2019style, karras2020analyzing}. Also, they are limited in synthesizing images at high resolution.

\textbf{GAN prior based real image editing.} GAN prior based methods, \ie, GAN inversion, are proposed to inference a latent code of a given image based on a pretrained GAN model such as StyleGAN~\cite{karras2019style, karras2020analyzing}.
These methods can be roughly divided into two categories, optimization-based~\cite{abdal2019image2stylegan, raj2019gan, abdal2020image2stylegan++, gu2020image, collins2020editing, xu2021continuity}
and learning-based~\cite{zhu2016generative, bau2019inverting, gu2020image, guan2020collaborative, tov2021designing, chai2021using, yang2021discovering}. The main advantage of the former techniques is that they can ensure superior image reconstruction, while the corresponding cost is a higher computational complexity.
In contrast, learning-based methods have a fast inference speed. Richardson \etal~\cite{richardson2020encoding} propose a pSp encoder that can embed real images into an extended ${\cal W}^+$ space. Xu \etal~\cite{xu2021generative} propose to train a hierarchical encoder based on a feature pyramid network. Alaluf \etal~\cite{alaluf2021restyle} introduce an
iterative refinement mechanism for learning the inversion of real images. However, these methods cannot faithfully reconstruct the image content, mainly due to the misalignment between training and test data. We aim to solve this problem in a novel composition-decomposition paradigm via differential activations.

\textbf{Interpreting CNN.}
Recent interpreting CNN models~\cite{zhou2016learning, selvaraju2017grad, fukui2019attention} attempt to understand the behaviour of the networks. Zhou \etal~\cite{zhou2016learning} propose CAM which aims to highlight the model's attention regarding a specific class. Selvaraju \etal~\cite{selvaraju2017grad} propose Grad-CAM without relying on the global average pooling layer. Lee \etal~\cite{lee2021lfi} propose LFI-CAM which treats the feature maps as masks and learns the feature importance for generating the attention maps. Note that these methods are usually performed on responses themselves, while our strategy, with a clear aim of real image editing, explores to capture the variation between edit and unedited image features.

\section{Approach}\label{sec:method}
\subsection{Overview}
\begin{figure*}[t]
	\centering
	\includegraphics[width=\linewidth]{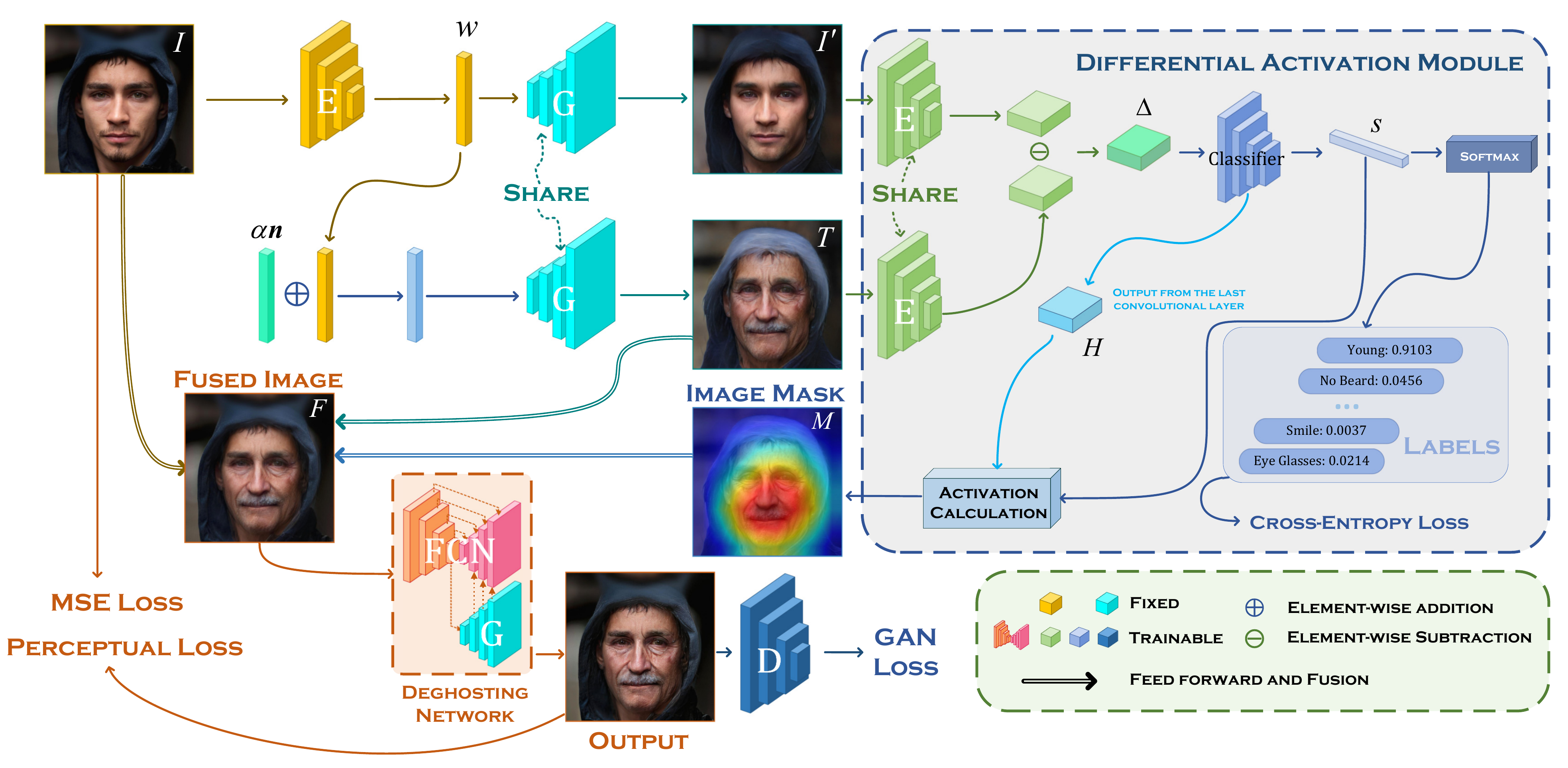}
	\vspace{-4mm}\caption{Overall pipeline of our model. Given an input image, we first invert it to the latent space and perform user-desired editing. Then a differential activation module is applied to track the semantic changes of the manipulation. Edited region and unedited one are further combined using a Diff-CAM mask to produce a coarse reconstruction, on which the ghosting artifacts can be mitigated by the deghosting network with the aid of GAN priors.}\vspace{-4mm}
	\label{fig:overall}
\end{figure*}

Due to the misaligned distributions between training and test data, existing GAN inversion methods cannot guarantee the fidelity of the reconstruction. And the quality of the subsequent edited image is therefore severely limited by such an out-of-domain problem. To remedy this, we propose a composition-decomposition paradigm for image editing and illustrate its overall pipeline in Fig.~\ref{fig:overall}.

Specifically, given an image as input, our method firstly inverts it into the latent space. Semantic manipulation can then be produced by feeding and varying the latent code into a pretrained and fixed generator. Consequently, an initial result can be obtained by fusing the original input and the edited inversion with a Diff-CAM mask as weight. In particular, the procedure of generating the Diff-CAM mask is encapsulated in a self-contained differential activation module, which exploits the differential information between two reconstructions to promote the accuracy for determining the range of the editing-relevant region. The final output is further generated by a deghosting network, which resorts to the diverse facial prior for mitigating the ghosting effect and enhancing the realism of the initial result. Note that we use the StyleGAN2 generator as the pretrained one in our model.

\subsection{GAN Inversion and Single Attribute Editing}
To achieve GAN inversion of a given image $\bm{I}$, we need to map it into a latent space in which rich semantic information is embedded. This can be implemented via many existing methods, for example, a pretrained pSp encoder $E_{\textrm{fixed}}(\cdot)$, and the latent code $\bm{w}$ can thus be formulated by $\bm{w}=E_{\textrm{fixed}}(I)$. Then we can obtain the inverted image $\bm{I}'$ by a pretrained StyleGAN2 generator $G(\cdot)$, which is formulated by $\bm{I}'=G(\bm{w})$. Note that there most likely exists a bias between $\bm{I}'$ and $\bm{I}$, due to the out-of-domain problem.


And to manipulate the corresponding attributes, the latent code would be varied along various intepretable directions that are discovered in the latent space. The edited inversion $\bm{T}$ can then be produced based on the altered code by the same generator $G(\cdot)$. Given a specific direction $\bm{n}$ for single attribute editing, this process can be formulated as $\bm{T}=G(\bm{w}+\alpha\bm{n})$, where $\alpha$ is a scaling factor. Note that our method can also support multi-attribute editing, and we will discuss this in Sec.~\ref{method:multi}.\vspace{-4mm}


\subsection{Differential Activation Module}
Once we have the paired images $\{\bm{I}', \bm{T}\}$, we respectively feed them into a plain trainable encoder $E_{\textrm{trainable}}(\cdot)$, and can easily obtain the differential features $\bm{\Delta}$ via a simple subtraction operation:
\begin{equation}
	\bm{\Delta}=E_{\textrm{trainable}}(\bm{I}')-E_{\textrm{trainable}}(\bm{T}).\\
	\label{equ:diff}
\end{equation}
Subsequently, these features are sent to a lightweight network which serves as a classifier and consists of convolutional and fully connected layers. We use the cross-entropy loss ${\cal L}_{\textrm{ce}}$ to train the encoder and the classifier together, which can be formulated as follows:
\begin{equation}
	{\cal L}_{\textrm{ce}}=-\sum_{c=1}^{N}{y_clog\frac{e^{s_c}}{\sum_{i=1}^{N}e^{s_i}}},\\
\end{equation}
where $\bm{y}=\{y_1,y_2,\cdots,y_N\}$ is a one-hot vector that indicates which attribute has been edited, $\bm{s}=\{s_1,s_2,\cdots,s_N\}$ denotes the output vector of the classifier before softmax operation, $e$ denotes the natural constant, and $N$ is the total number of attributes.

Now we are ready for performing activation calculation. The first step is to define the weight $\beta_c^k$ that corresponded to the $k$th channel of $\bm{H}$ and the $c$th attribute, which is formulated as follows:
\begin{equation}
	\beta_c^k=\overset{\textrm{global~average~pooling}}{\overbrace{\frac{1}{Z}\sum_i\sum_j}}\frac{\partial s_c}{\partial \bm{H}^k_{ij}},
\end{equation}
where $\bm{H}$ is the features generated by the last convolutional layer in the classifier, $i$ and $j$ respectively denotes the height and the width of the features.

Then our Diff-CAM mask $\bm{M}_{\textrm{Diff-CAM}}$ can be represented as a piecewise linear transformation of weighted differential features, that is
\begin{equation}
	\bm{M}_{\textrm{Diff-CAM}}=\textrm{ReLU}(\sum_k\beta_c^k\bm{H}^k).\\
	\label{equ:mask}
\end{equation}
Finally, we normalize the above mask into the interval of $[0,1]$ via $\bm{M}_{\textrm{Diff-CAM}}=\bm{M}_{\textrm{Diff-CAM}}/\textrm{max}(\bm{M}_{\textrm{Diff-CAM}})$. Since the Diff-CAM mask is generated based on the differential features that describe semantically changes, the range of the editing-relevant region can be detected more accurately via a comprehensive activation. It is thus more suitable than other CAM-based masks for image editing.

\subsection{Composition}
After obtaining the Diff-CAM mask, it is time to composite the edited image with the original input for resolving the out-of-domain issue. We have the fused image $\bm{F_{\textrm{fused}}}$ by the following weighted average formula:
\begin{equation}
	\bm{F_{\textrm{fused}}}=\bm{T}\odot \bm{M}_{\textrm{Diff-CAM}}+\bm{I}\odot(1-\bm{M}_{\textrm{Diff-CAM}}), \\
\end{equation}
where $\odot$ denotes the hadamard product. However, the quality of such an initial blending result is unsatisfactory due to an inevitable ghosting effect.\vspace{-4mm}

\subsection{Deghosting Network}
To cope with ghosting artifacts, we treat the coarse reconstruction $\bm{F_{\textrm{fused}}}$ as a combination of a target image and a ghost image. In order to decompose the target image out, we further perform a deghosting process on the coarse result via a deghosting network. As shown in Fig.~\ref{fig:overall}, the architecture of the network includes a fully convolutional network which consists of an encoder (the orange part), a decoder (the pink part), a pretrained StyleGAN2 generator, and a discriminator $D(\cdot)$. Note that we denote the aggregation of the first three modules as $\phi(\cdot)$.

Our goal is to utilize the ghosting-free nature of the inherent facial prior in the pretrained GAN model, such that ghosting artifacts can be removed without destroying the original facial details. In particular, we first feed $\bm{F_{\textrm{fused}}}$ to an FCN-like~\cite{long2015fully} encoder-decoder architecture for two purposes: the encoder generates the latent code of $\bm{F_{\textrm{fused}}}$, and the decoder is trained to produce ghosting-free results. Meanwhile, with the predicted latent code, $\bm{F_{\textrm{fused}}}$ is inverted in the latent space and reconstructed by the StyleGAN2 generator without ghosting artifacts. We aggregate the corresponding features of the generator with the decoder hierarchically, yielding the final deghosting result.

Since the fused image $\bm{F_{\textrm{fused}}}$ has no ground-truth counterpart, we synthesize a set of paired data $\{\bm{F}_{\textrm{train}},\bm{I}\}$ to train the deghosting network. The training image $\bm{F}_{\textrm{train}}$ is given by
\begin{equation}
	\bm{F}_{\textrm{train}}=\bm{T}\odot \bm{M}_{\textrm{train}}+\bm{I}\odot(1-\bm{M}_{\textrm{train}}), \\
\end{equation}
and the corresponding Diff-CAM mask $\bm{M}_{\textrm{train}}$ is defined as follows:
\begin{equation}
	\small
	\bm{M}_{\textrm{train}}(i,j)= \left\{\begin{matrix}
		\bm{M}_{\textrm{Diff-CAM}}(i,j), & \textrm{if}~~~\bm{M}_{\textrm{Diff-CAM}}(i,j)\le0.5, \\
		1 - \bm{M}_{\textrm{Diff-CAM}}(i,j), & \textrm{if}~~~\bm{M}_{\textrm{Diff-CAM}}(i,j)>0.5.
	\end{matrix}\right.
\end{equation}

The rationale behind this setting is that: 1) the mask $\bm{M}_{\textrm{train}}$ is thus regularized into an interval of $[0,0.5]$ so that the content of $\bm{I}$ dominates that of $\bm{F}_{\textrm{train}}$. We can then treat image $\bm{I}$ as the required ground truth. And 2) meanwhile, the ghosting effect still exists in $\bm{F}_{\textrm{train}}$. Note that the corresponding attribute in regard to generate the mask $\bm{M}_{\textrm{train}}$ is consistent with $\bm{T}$ and randomly selected. The total objective ${\cal L}_{\textrm{deghost}}$ for optimizing the deghosting network is defined as follows:
\begin{equation}
	{\cal L}_{\textrm{deghost}}=\lambda_m{\cal L}_{\textrm{mse}}+\lambda_p{\cal L}_{\textrm{percep}}+\lambda_a{\cal L}_{\textrm{adv}},
	\label{equ:obj}
\end{equation}
where ${\cal L}_{\textrm{mse}}$, ${\cal L}_{\textrm{percep}}$ and ${\cal L}_{\textrm{adv}}$ respectively denotes MSE loss, perceptual loss and adversarial loss, $\lambda_m$, $\lambda_p$, $\lambda_a$ are the balance factors. And the involved three losses are respectively defined as follows:
\begin{equation}
	{\cal L}_{\textrm{mse}}=\frac{1}{Q}\|\bm{I}-\phi(\bm{F}_{\textrm{train}})\|_2,
\end{equation}
\begin{equation}
	{\cal L}_{\textrm{percep}}=\frac{1}{Q}\|V(\bm{I})-V(\phi(\bm{F}_{\textrm{train}}))\|_2,
\end{equation}
\begin{equation}
	{\cal L}_{\textrm{adv}}=\underset{\bm{I}\sim P_r}{\mathbb{E}}log(D(\bm{I})) + \underset{\bm{F}_{\textrm{train}}\sim P_g}{\mathbb{E}}log(1-D(\phi(\bm{F}_{\textrm{train}}))),
\end{equation}
where $Q$ indicates the number of pixels, $P_r$ and $P_g$ respectively denotes the distribution of real data and generated data, $V(\cdot)$ denotes a pretrained VGG-16 network, and we select the features produced by the conv4$\_$3 layer for modeling the loss.\vspace{-4mm}

\subsection{Multi-attribute editing}\label{method:multi}
Our method also has a flexibility to handle multi-attribute editing. In fact, it can be decomposed into a sequence of single attribute editing tasks. Suppose the number of the attributes needed to be edited is $r$, three special points should be noted: 1) In the $i$th ($i\neq1$) single attribute editing, the paired images $\{\bm{T}_i,\bm{T}_{i-1}\}$ are used for calculating the Diff-CAM mask. At last we will have a set of $r$ masks. 2) The final Diff-CAM mask is the result of performing element-wise maximization operation on the mask set. And 3) The final fused image is the composition of $\bm{T}_r$ and $\bm{I}$ with the final mask as weight.\vspace{-4mm}

\section{Experiments}\label{sec:experiment}
\subsection{Implementation Details}
We implement our method in Pytorch on a PC with an Nvidia GeForce RTX 3090. During training, we use Adam as the optimizer with a learning rate of 0.0001, $\beta_1=0.9$, $\beta_2=0.99$. The hyperparameters in Eq.~(\ref{equ:obj}) are empirically set to $\lambda_m=1$, $\lambda_p=0.8$ and $\lambda_a=0.01$. Before being sent to $E_{\textrm{train}}$, $\bm{I}'$ and $\bm{T}$ are downsampled from a resolution of $1024\times1024$ to that of $256\times256$. And the Diff-CAM mask computed by Eq.~(\ref{equ:mask}) will be upsampled to a resolution of $1024\times1024$ before being used in the composition process.\vspace{-2mm}

\subsection{Experimental Data}
FFHQ dataset~\cite{karras2019style} and Celeba-HQ dataset~\cite{liu2015deep} both contain human face images of high quality and resolution, with 70000 and 30000 images respectively. We employ the FFHQ dataset for training the differential activation module and the deghosting network, while we utilize the Celeba-HQ dataset for testing. All the quantitative metrics are calculated on the Celeba-HQ dataset.\vspace{-2mm}

\subsection{Component Analysis}
\textbf{Effectiveness of DA module.} First, in order to prove the effectiveness of our design of the DA module structure, we replace the DA module with the commonly used Grad-CAM~\cite{selvaraju2017grad} and check out how the masks differ and influence the editing.

\begin{figure*}[t]
	\centering
	\captionsetup[subfigure]{font=scriptsize,labelfont=scriptsize}
	\begin{subfigure}[t]{0.155\textwidth}
		\centering
		\captionsetup{justification=centering}
		\includegraphics[width=\textwidth]{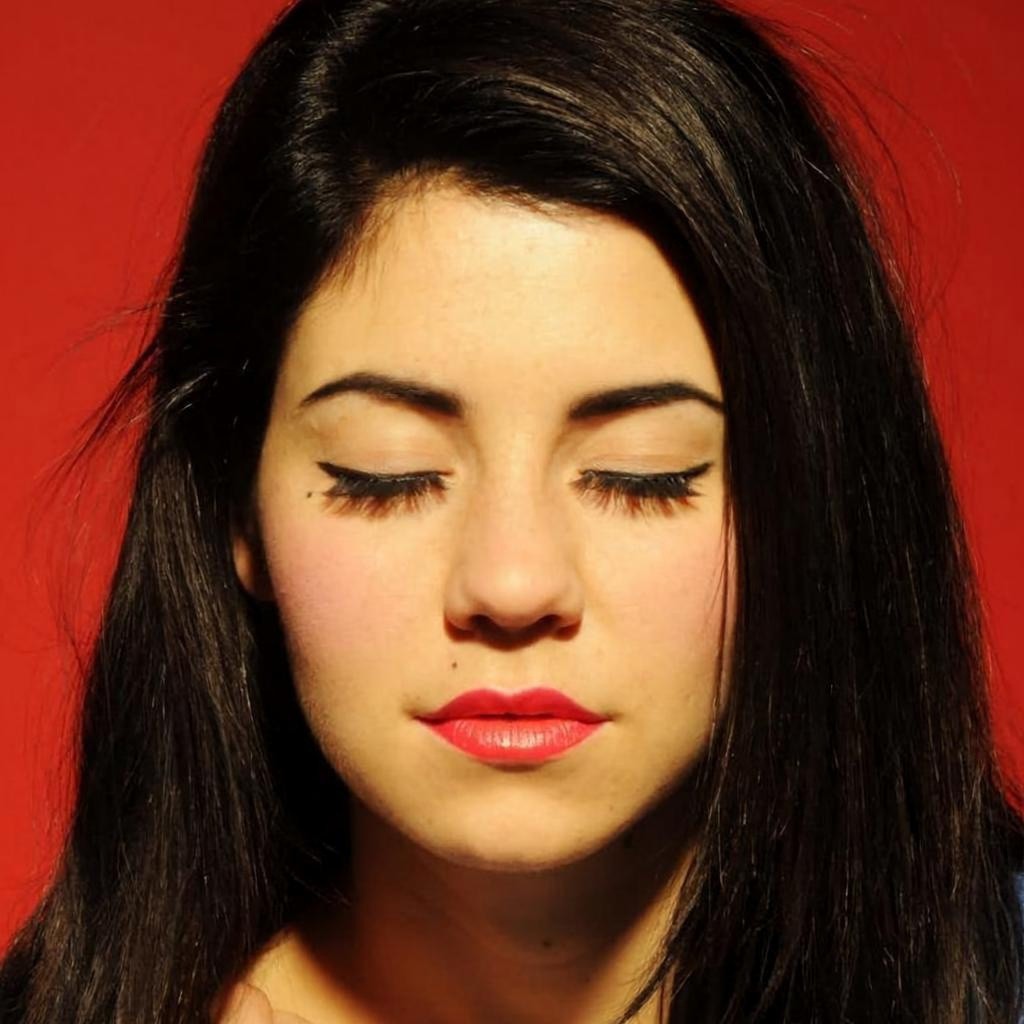} \includegraphics[width=\textwidth]{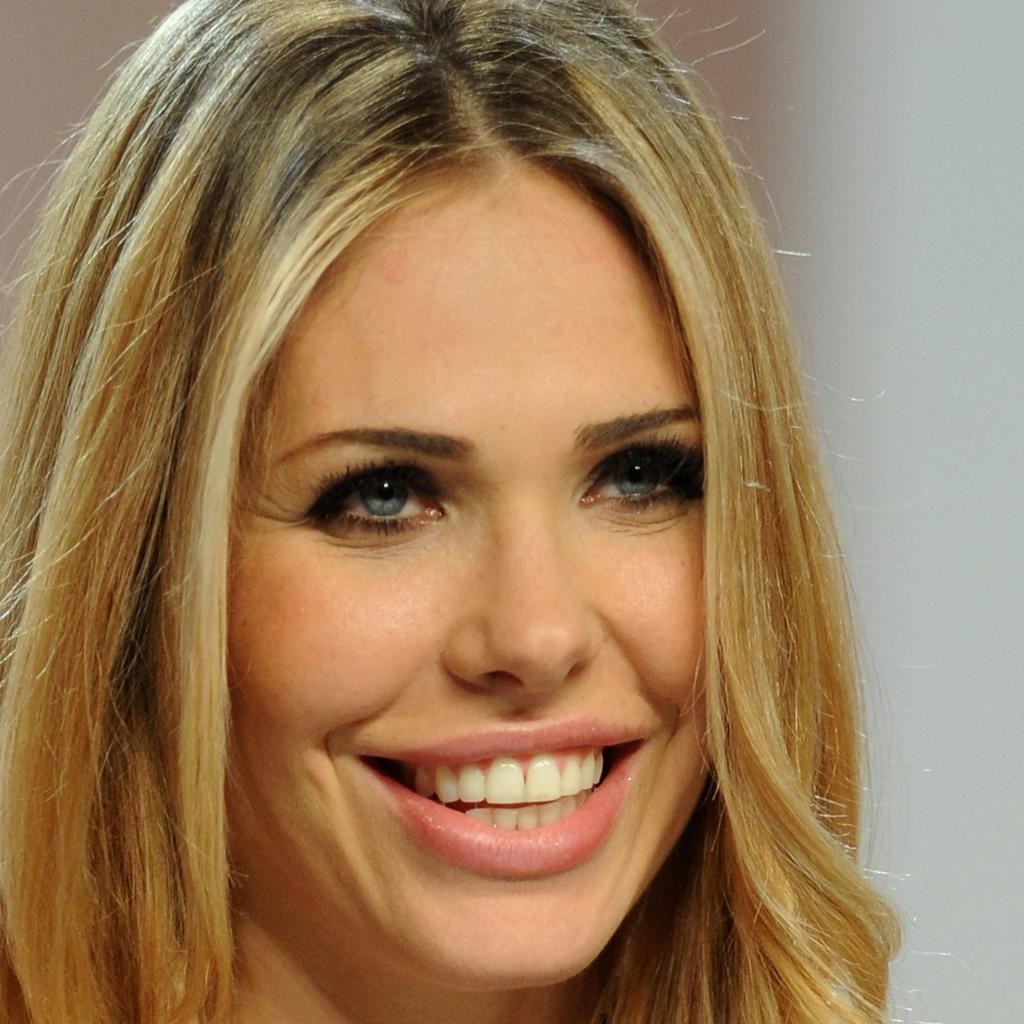}
		\caption{Input}
	\end{subfigure}
	\begin{subfigure}[t]{0.155\textwidth}
		\centering
		\captionsetup{justification=centering}
		\includegraphics[width=\textwidth]{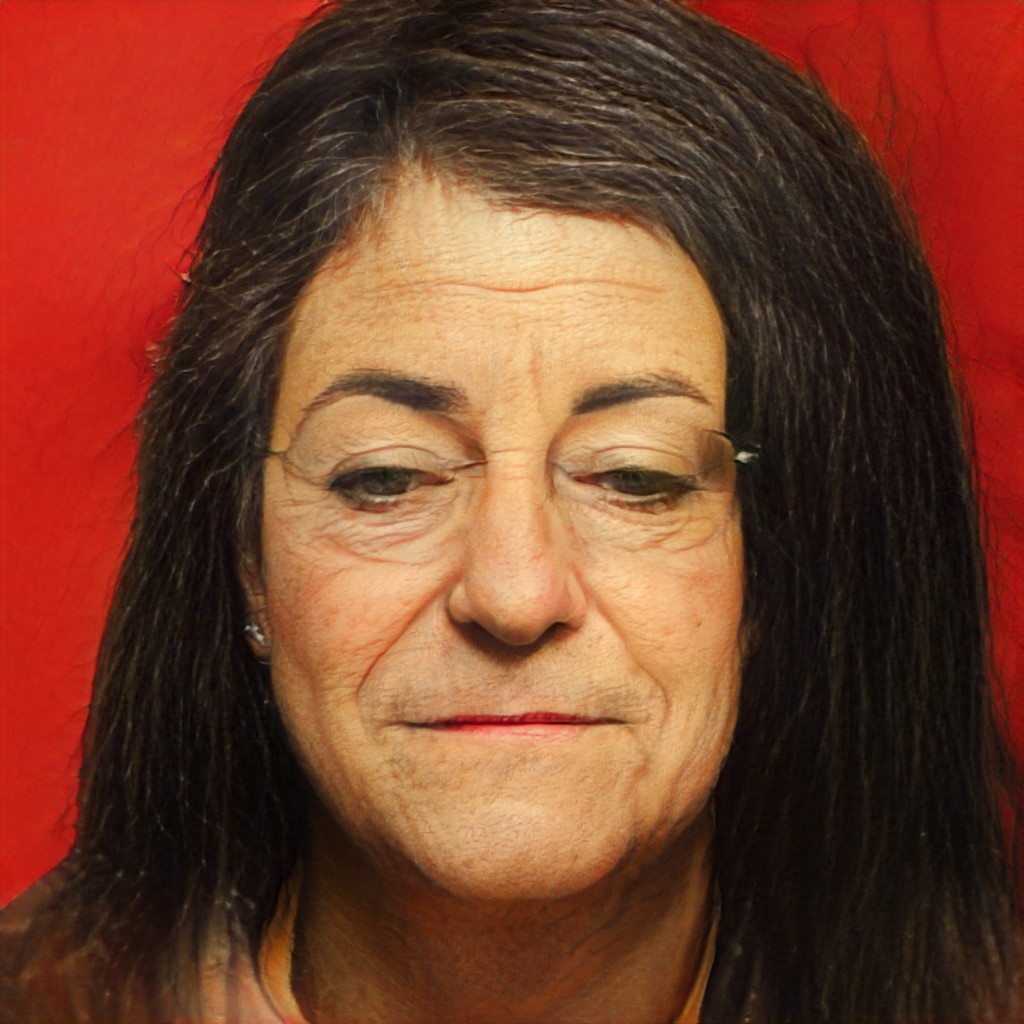} \includegraphics[width=\textwidth]{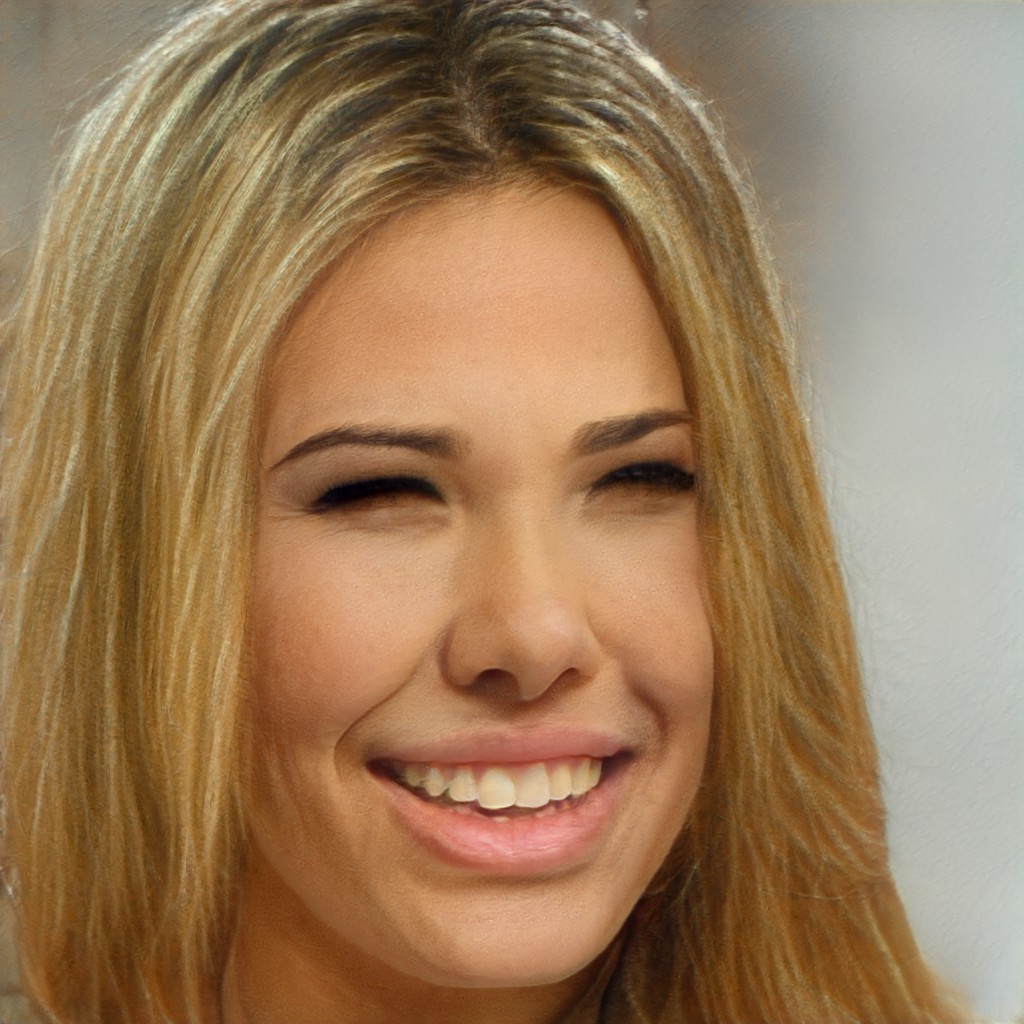}
		\caption{``Age'' \protect\\ \& ``Eyes open''}
	\end{subfigure}
	\begin{subfigure}[t]{0.155\textwidth}
		\centering
		\captionsetup{justification=centering}
		\includegraphics[width=\textwidth]{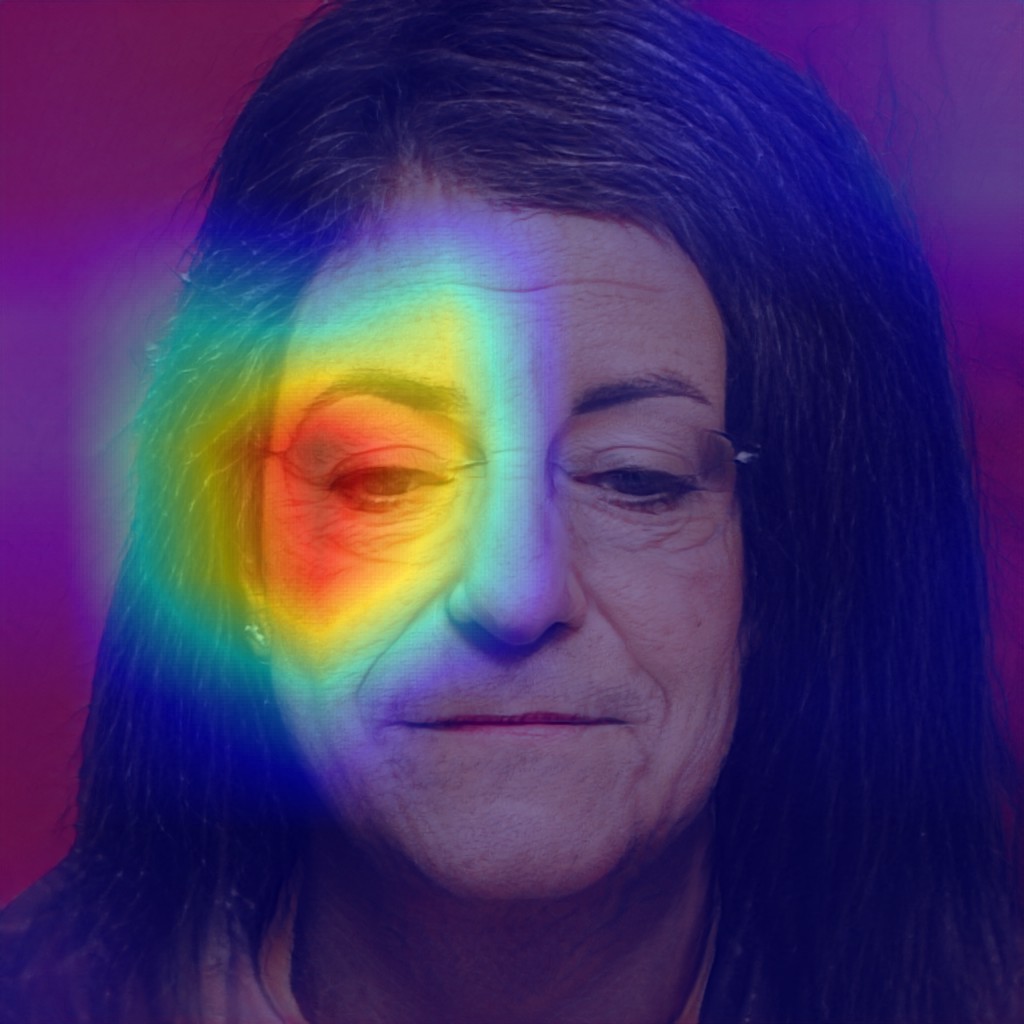} \includegraphics[width=\textwidth]{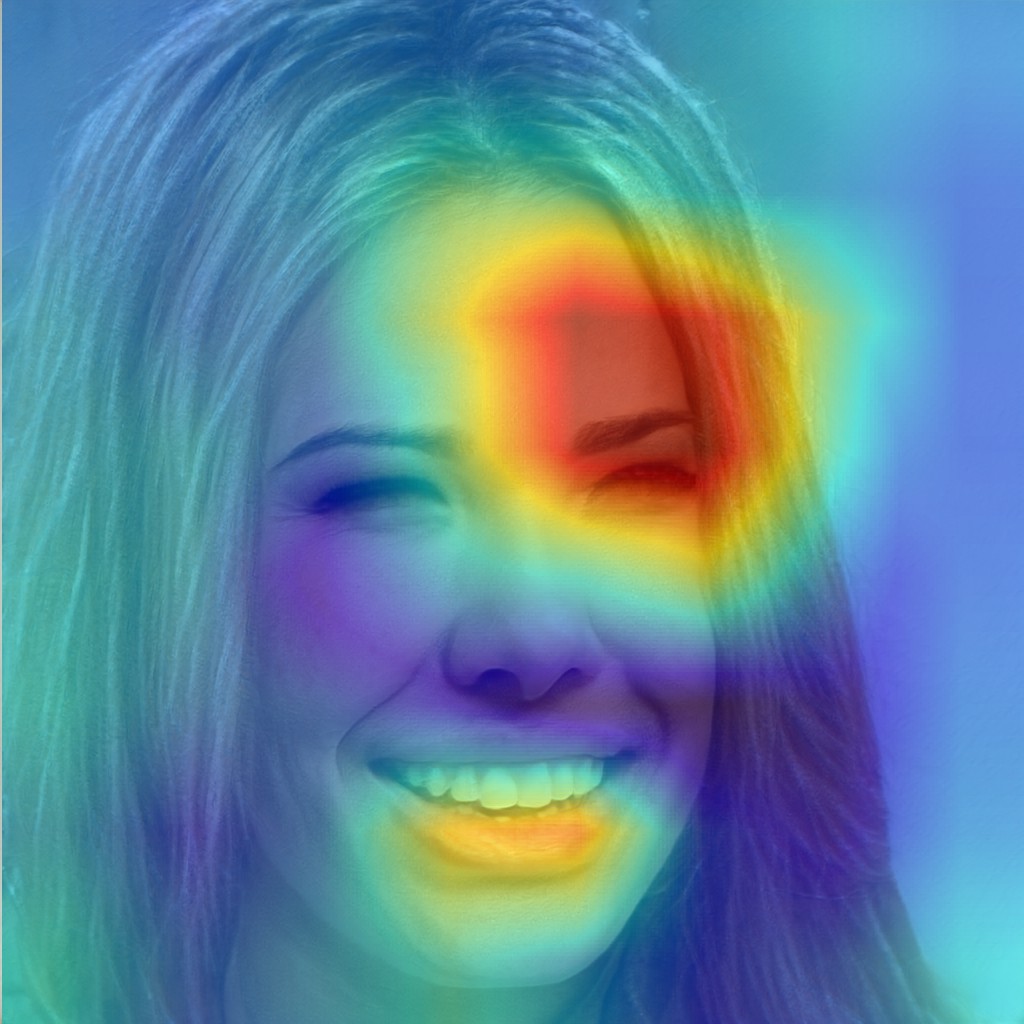}
		\caption{Grad-CAM\protect\\\cite{selvaraju2017grad}}
	\end{subfigure}
	\begin{subfigure}[t]{0.155\textwidth}
		\centering
		\captionsetup{justification=centering}
		\includegraphics[width=\textwidth]{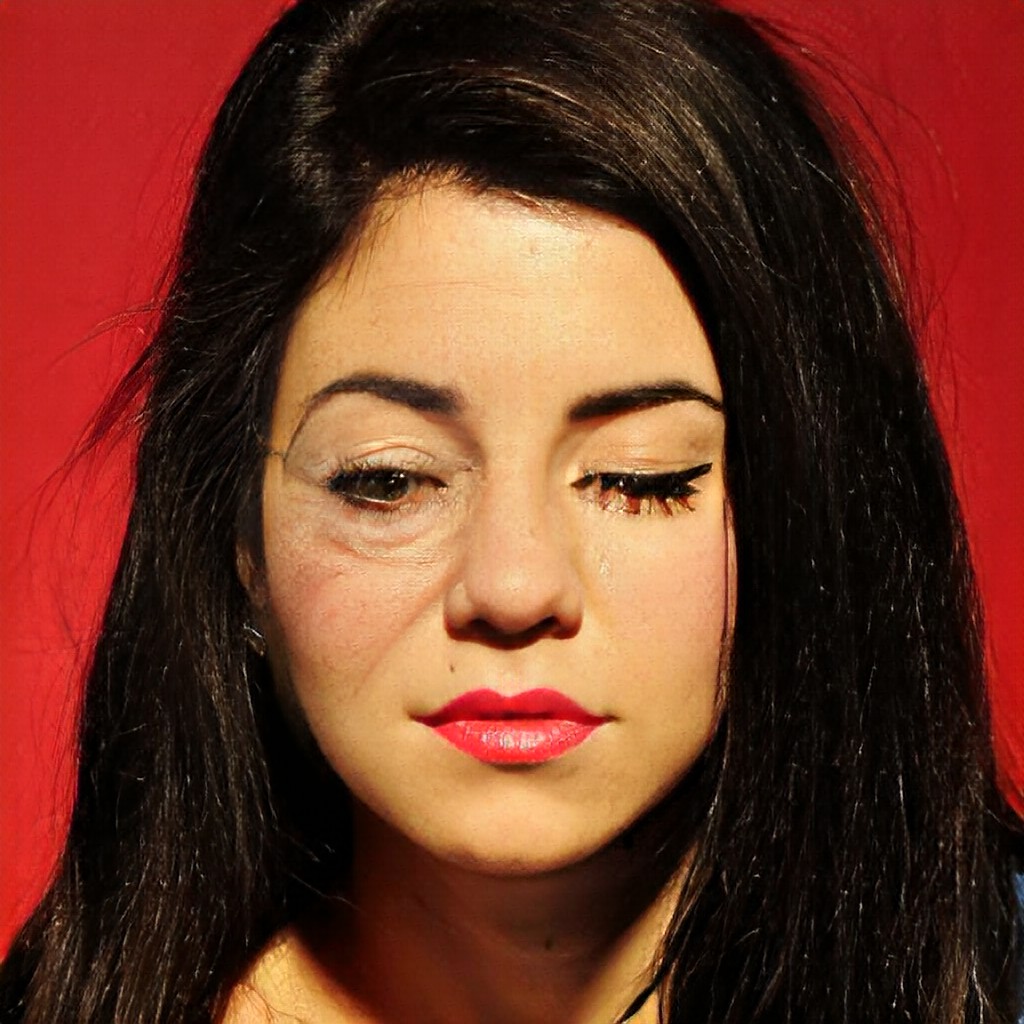} \includegraphics[width=\textwidth]{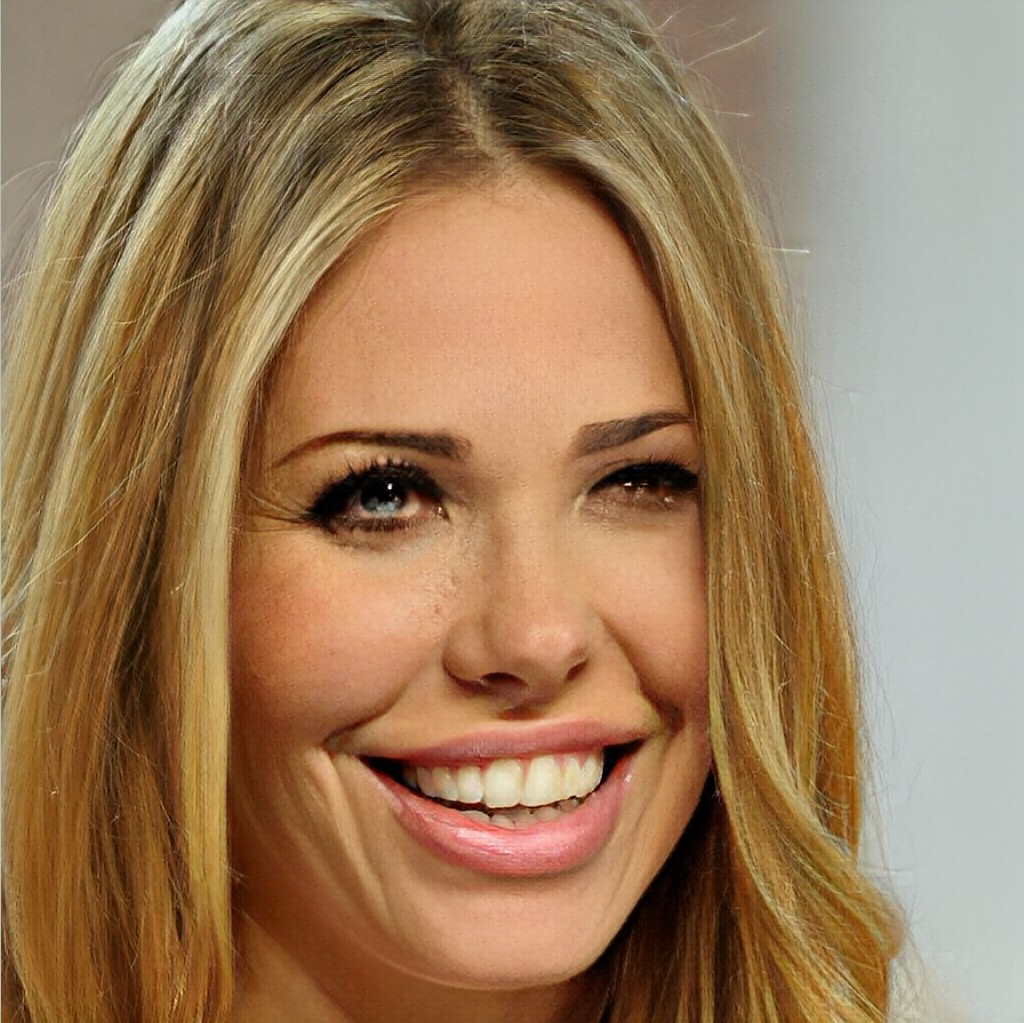}
		\caption{Editing \protect\\based on (c)}
	\end{subfigure}
	\begin{subfigure}[t]{0.155\textwidth}
		\centering
		\captionsetup{justification=centering}
		\includegraphics[width=\textwidth]{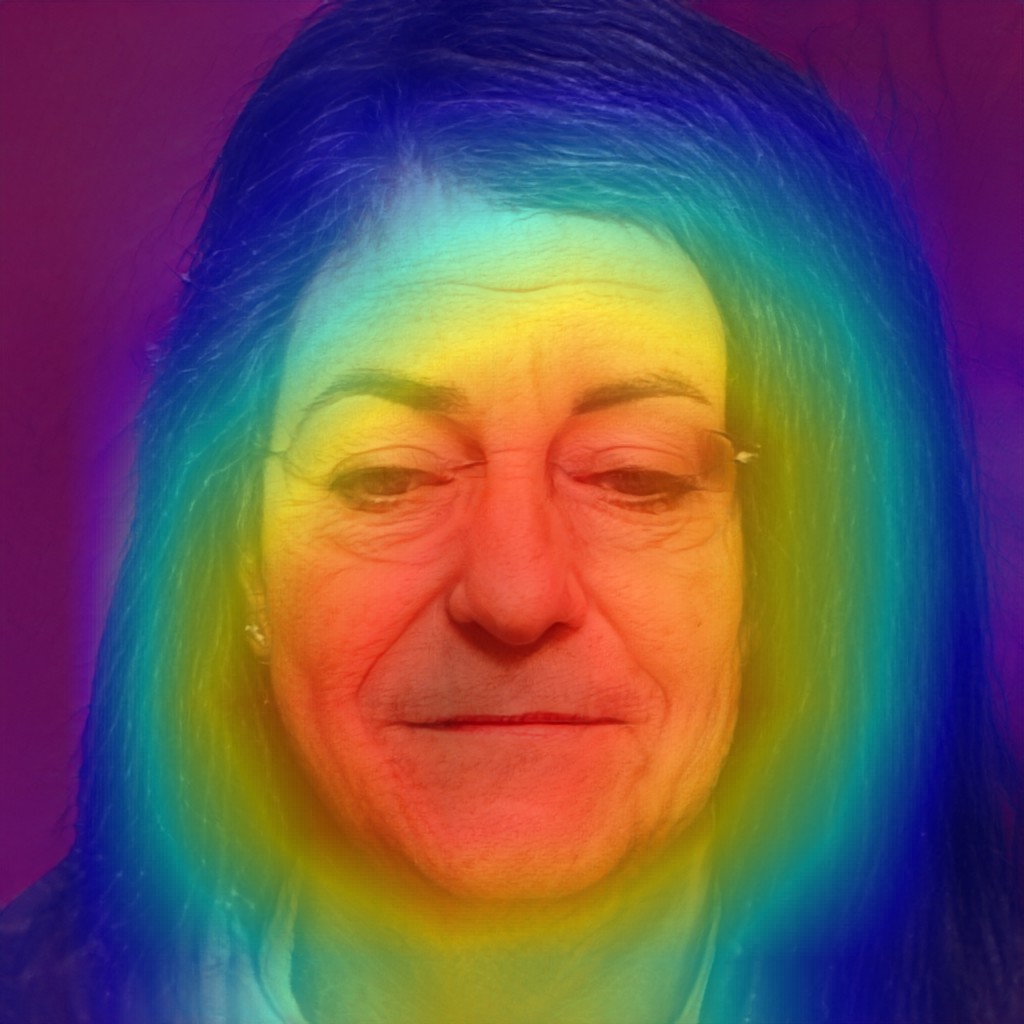} \includegraphics[width=\textwidth]{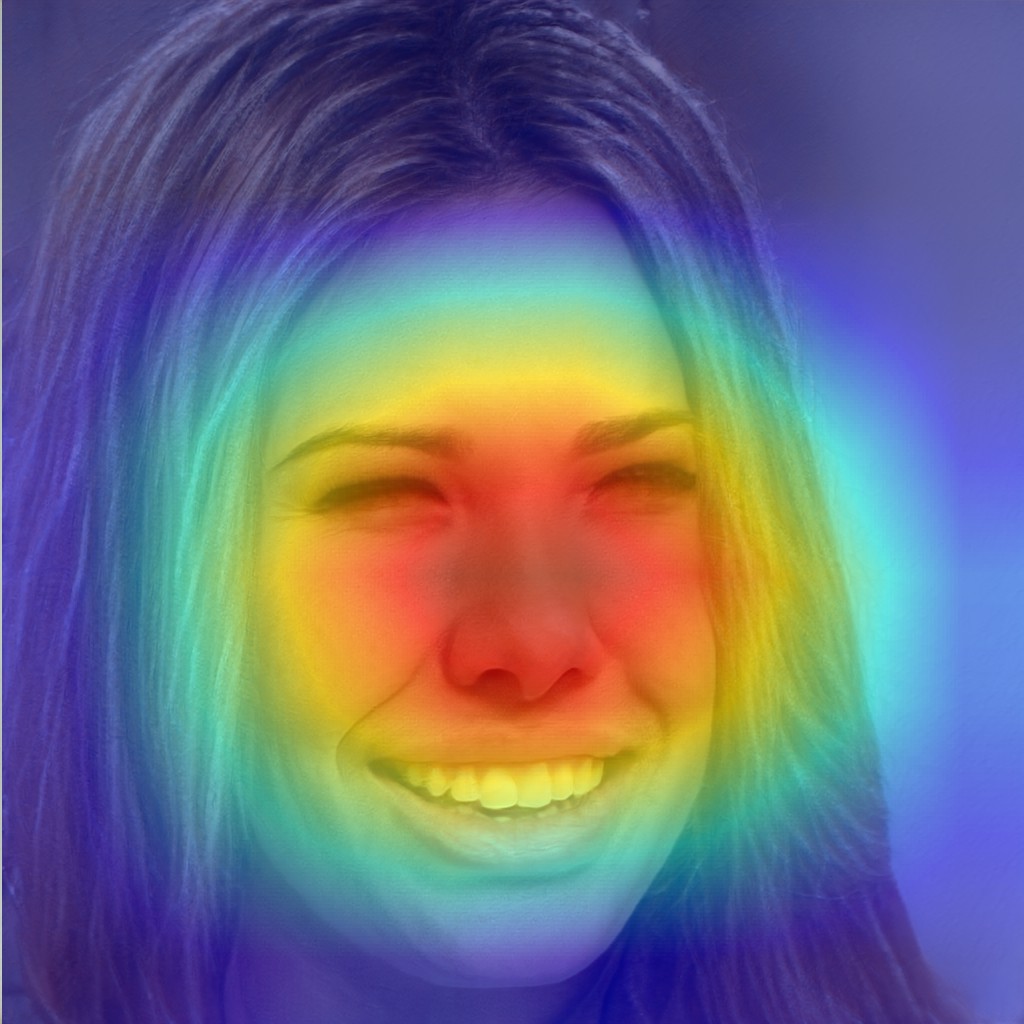}
		\caption{Our \protect\\Diff-CAM}
	\end{subfigure}
	\begin{subfigure}[t]{0.155\textwidth}
		\centering
		\captionsetup{justification=centering}
		\includegraphics[width=\textwidth]{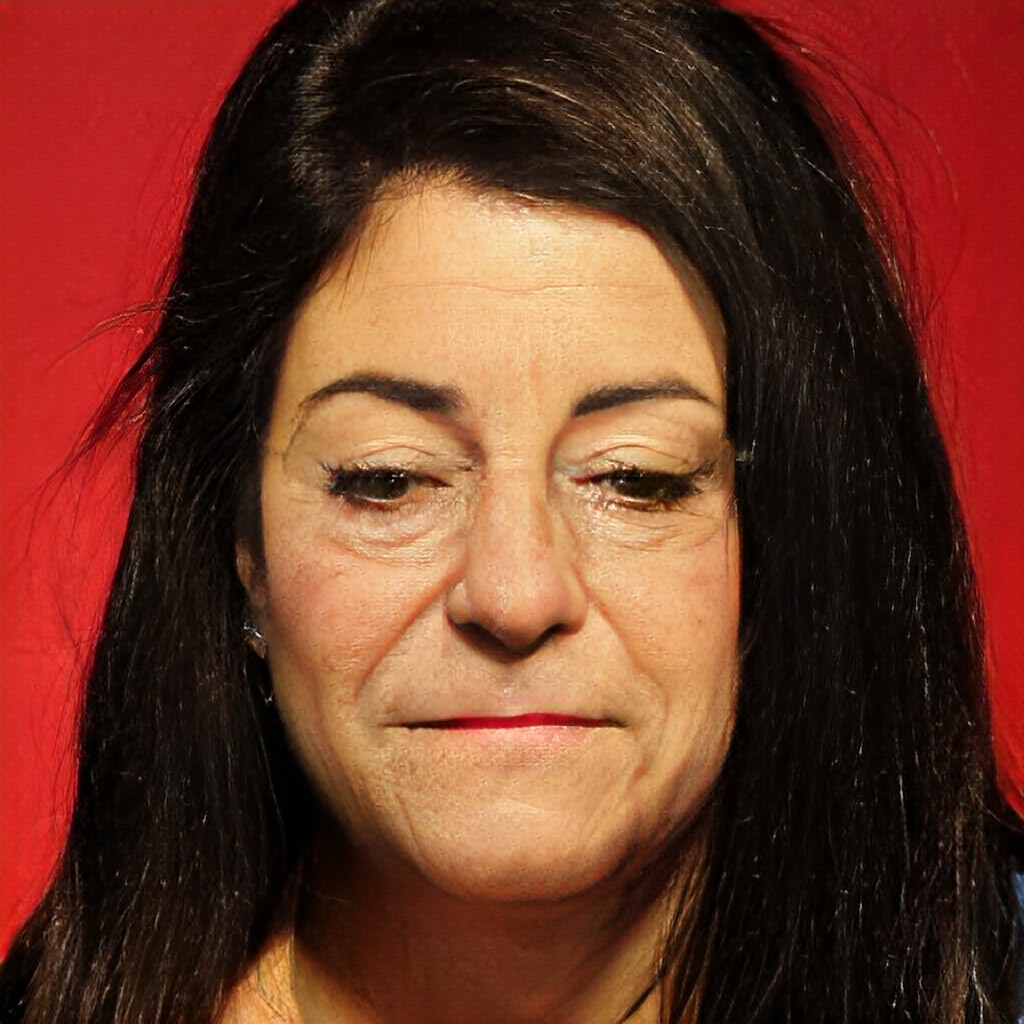} \includegraphics[width=\textwidth]{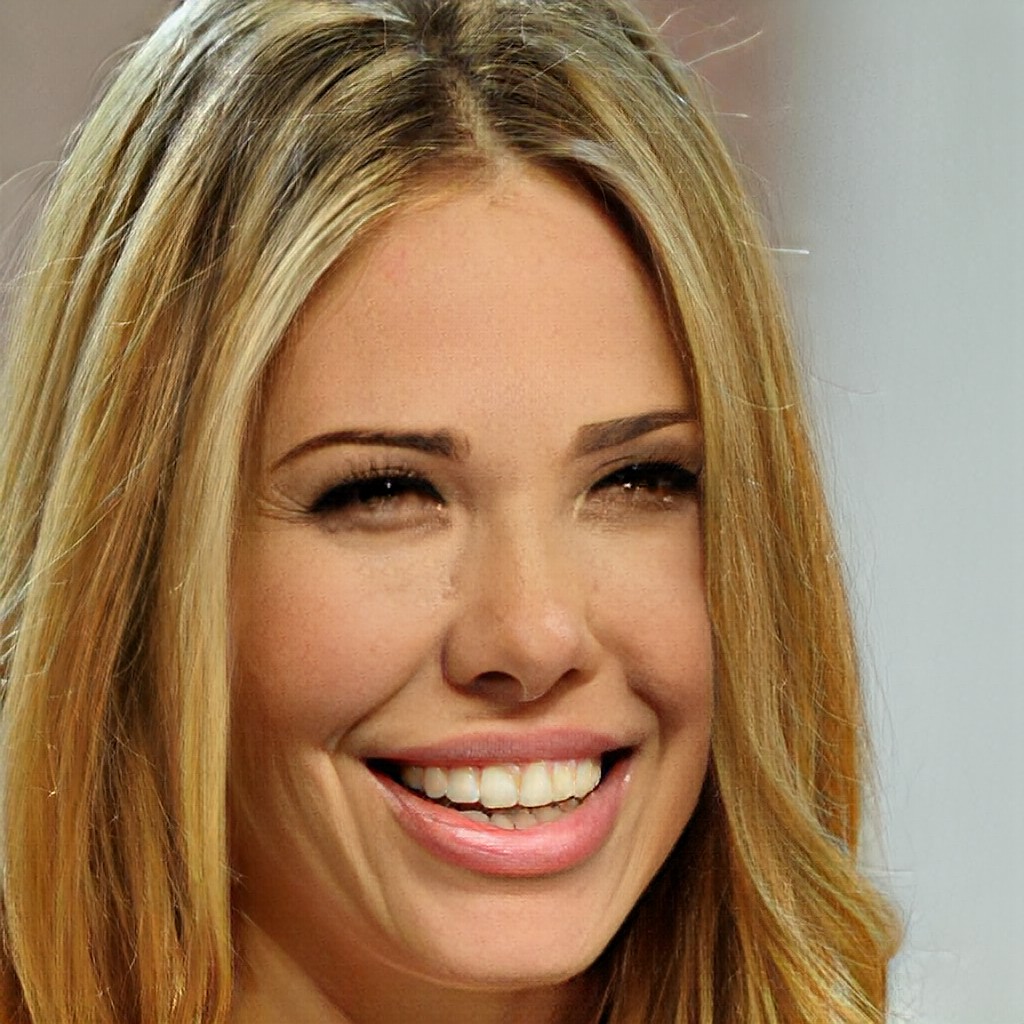}
		\caption{Editing \protect\\ based on (e)}
	\end{subfigure}
	\vspace{-2mm}\caption{Comparison with Grad-CAM in our editing framework. Grad-CAM exhibits localized attentions, while ours can correctly locate semantic changes during editing.}\vspace{-6mm}
	\label{fig:component-ordinary classifier}
\end{figure*}

The results are shown in Fig.~\ref{fig:component-ordinary classifier}. The results show the limitation of Grad-CAM that activates only in local areas. The resulted masks cannot suit for discovering semantic differences and therefore not suit for blending GAN inversion with the source image. On the contrary, the masks generated by our DA module successfully cover the editing-relevant regions, producing a global coverage for ``age'' attribute while a local attention for ``eyes open'' attribute. This largely aids the blending of edited and unedited regions.

\begin{figure}[t]
	\centering
	\captionsetup[subfigure]{font=scriptsize,labelfont=scriptsize}
	\begin{subfigure}[t]{0.23\linewidth}
		\centering
		\includegraphics[width=\textwidth]{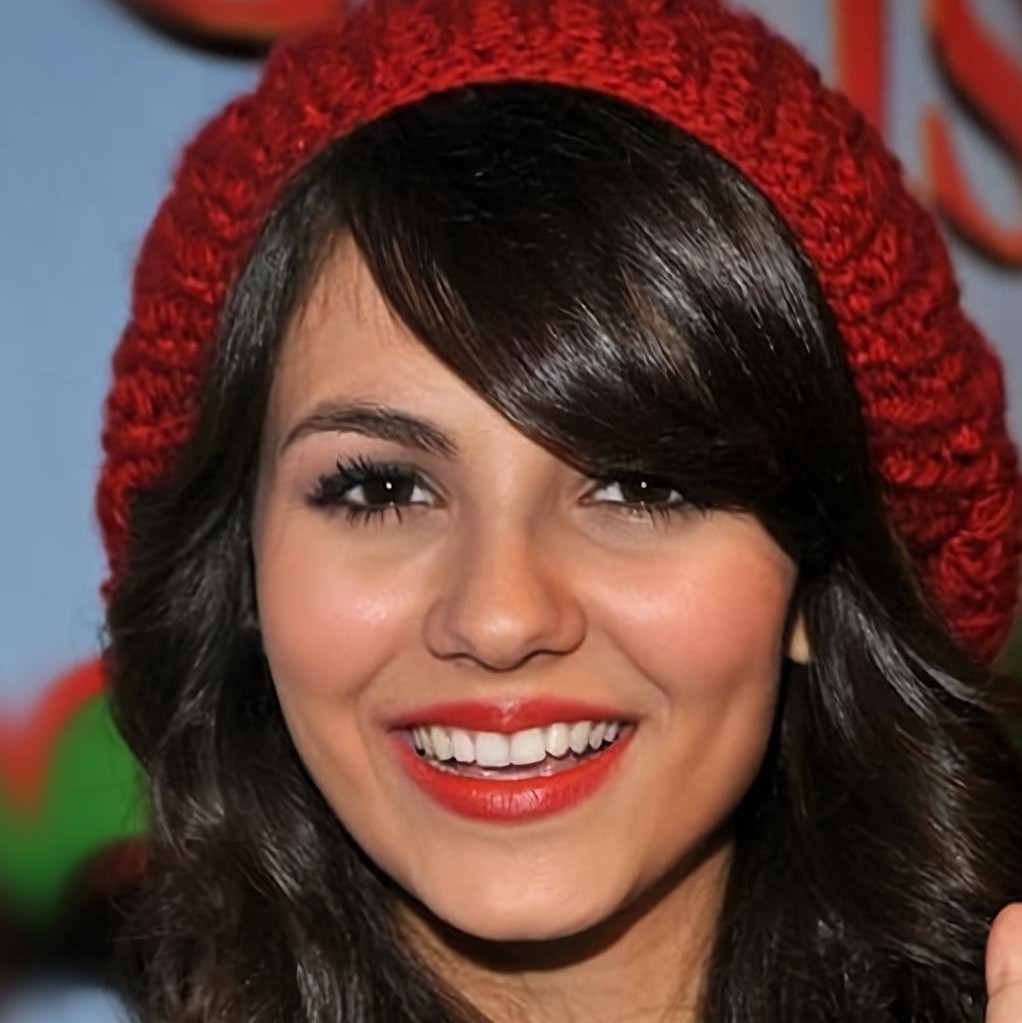}
		\includegraphics[width=\textwidth]{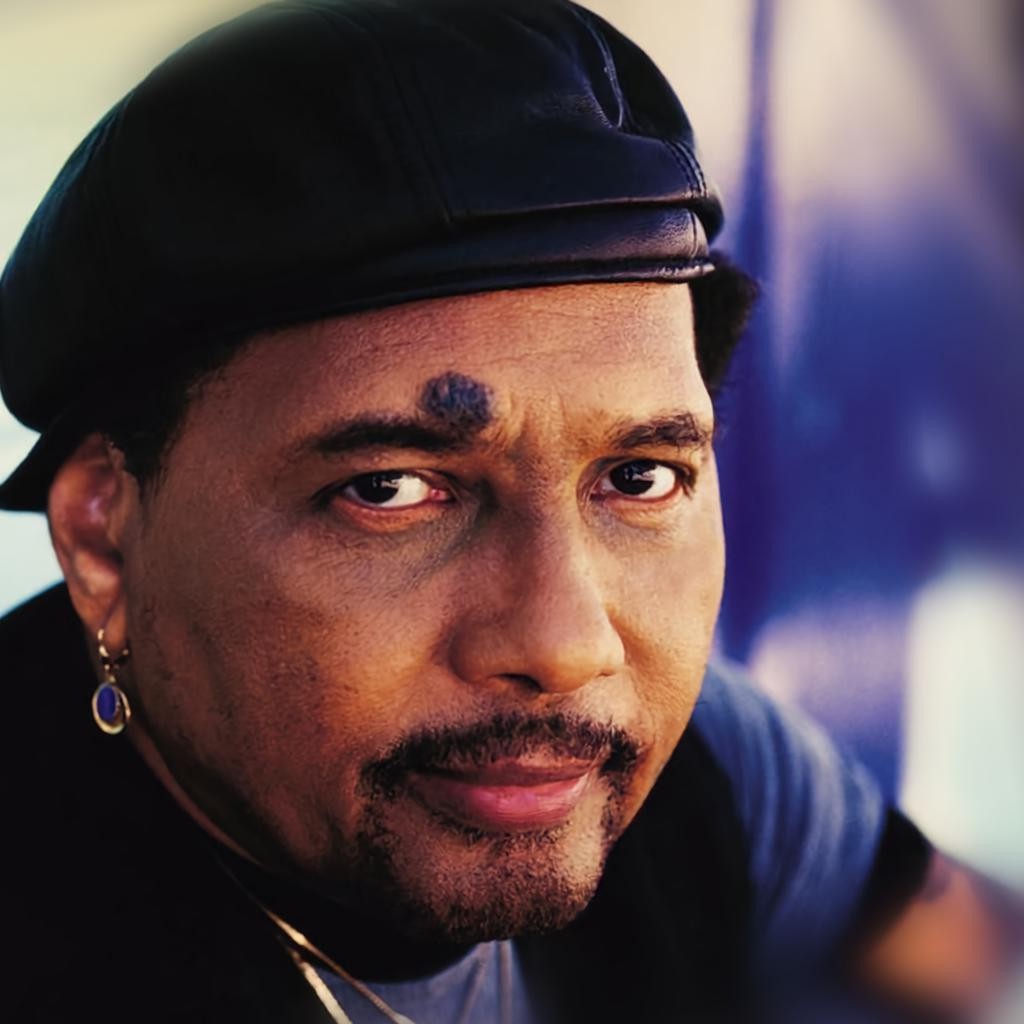}
		\caption{Input}
		\label{fig:a}
	\end{subfigure}
	\begin{subfigure}[t]{0.23\linewidth}
		\centering
		\includegraphics[width=\textwidth]{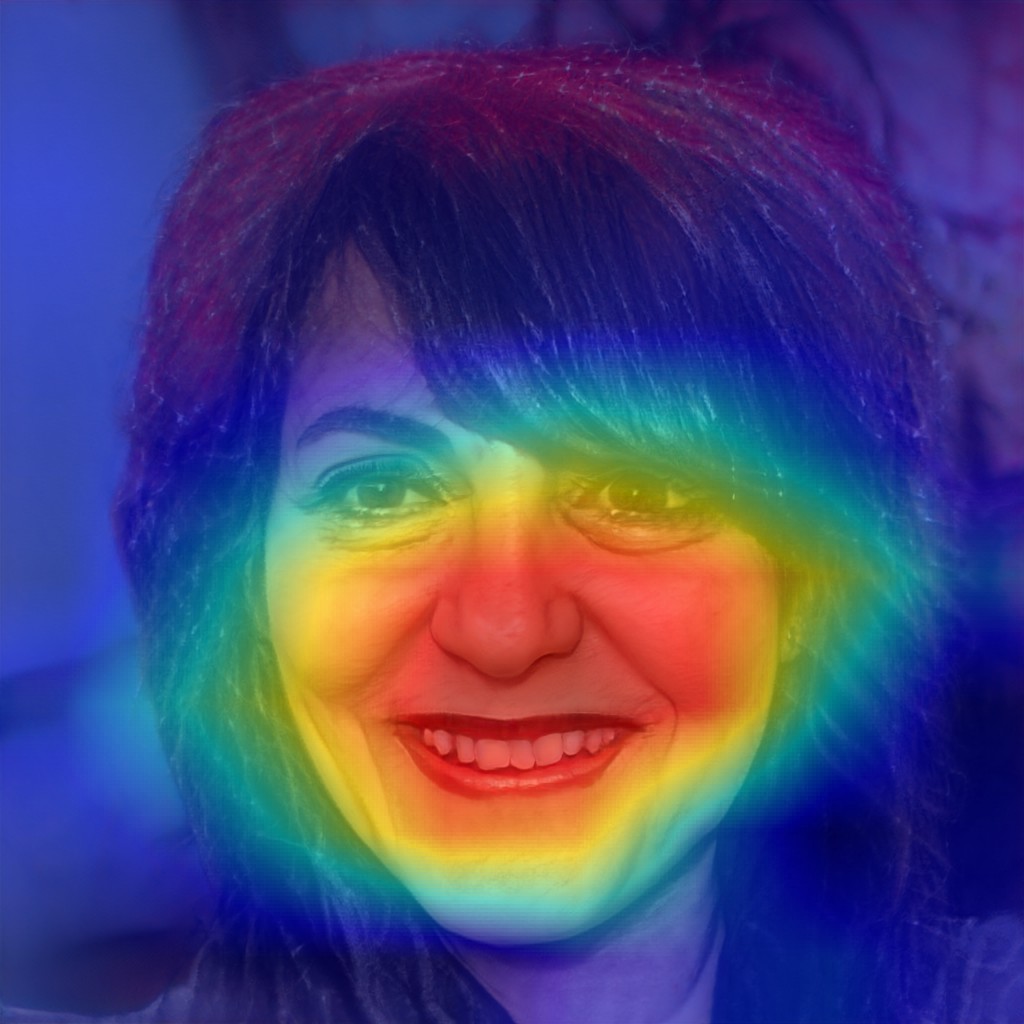}
		\includegraphics[width=\textwidth]{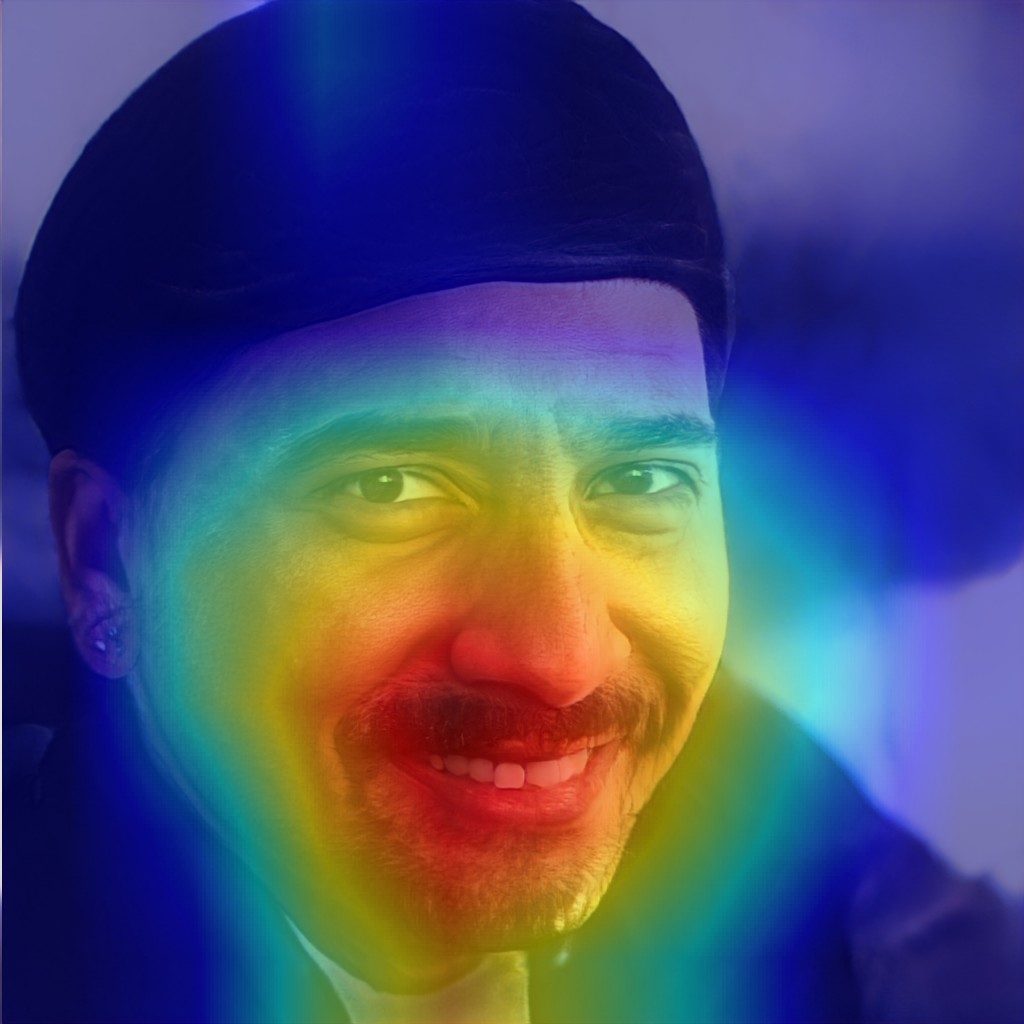}
		\caption{Editing \& Mask}
		\label{fig:d}
	\end{subfigure}
	\begin{subfigure}[t]{0.23\linewidth}
		\centering
		\includegraphics[width=\textwidth]{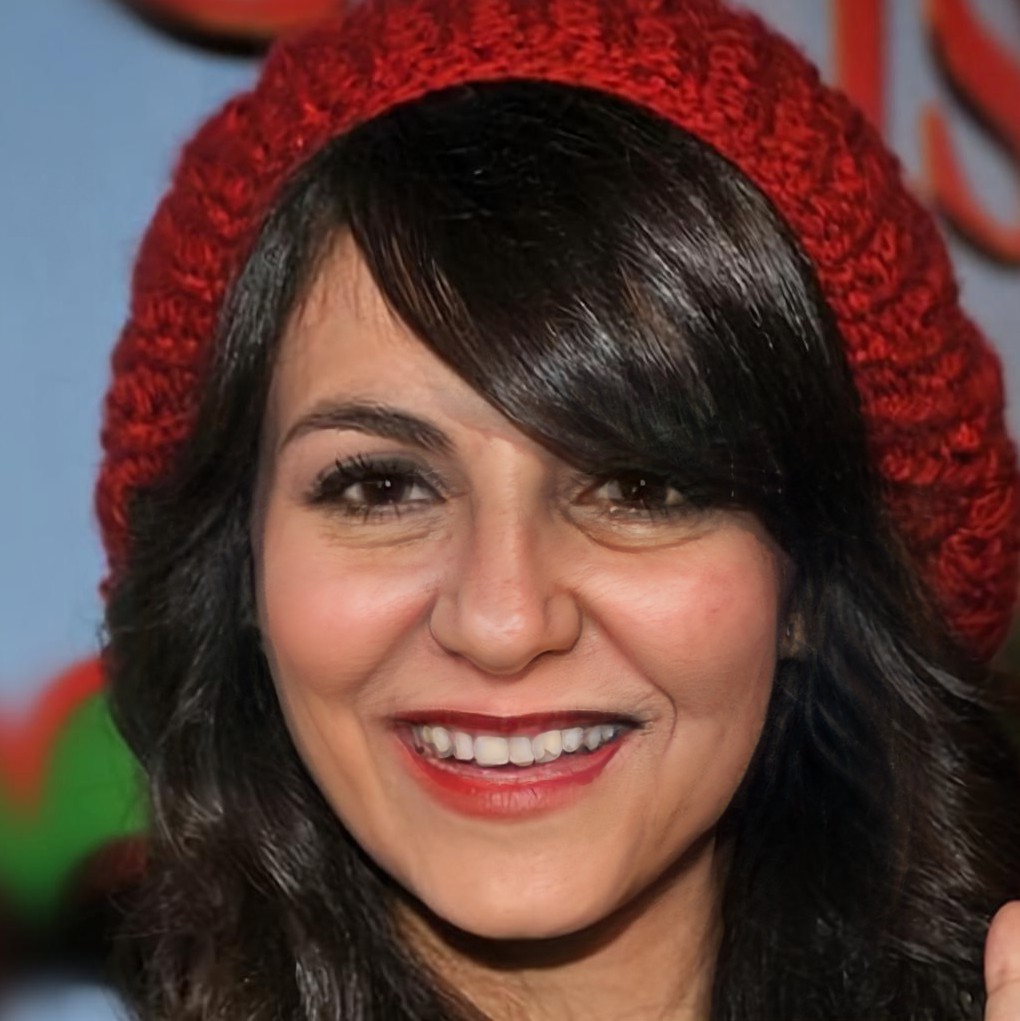}
		\includegraphics[width=\textwidth]{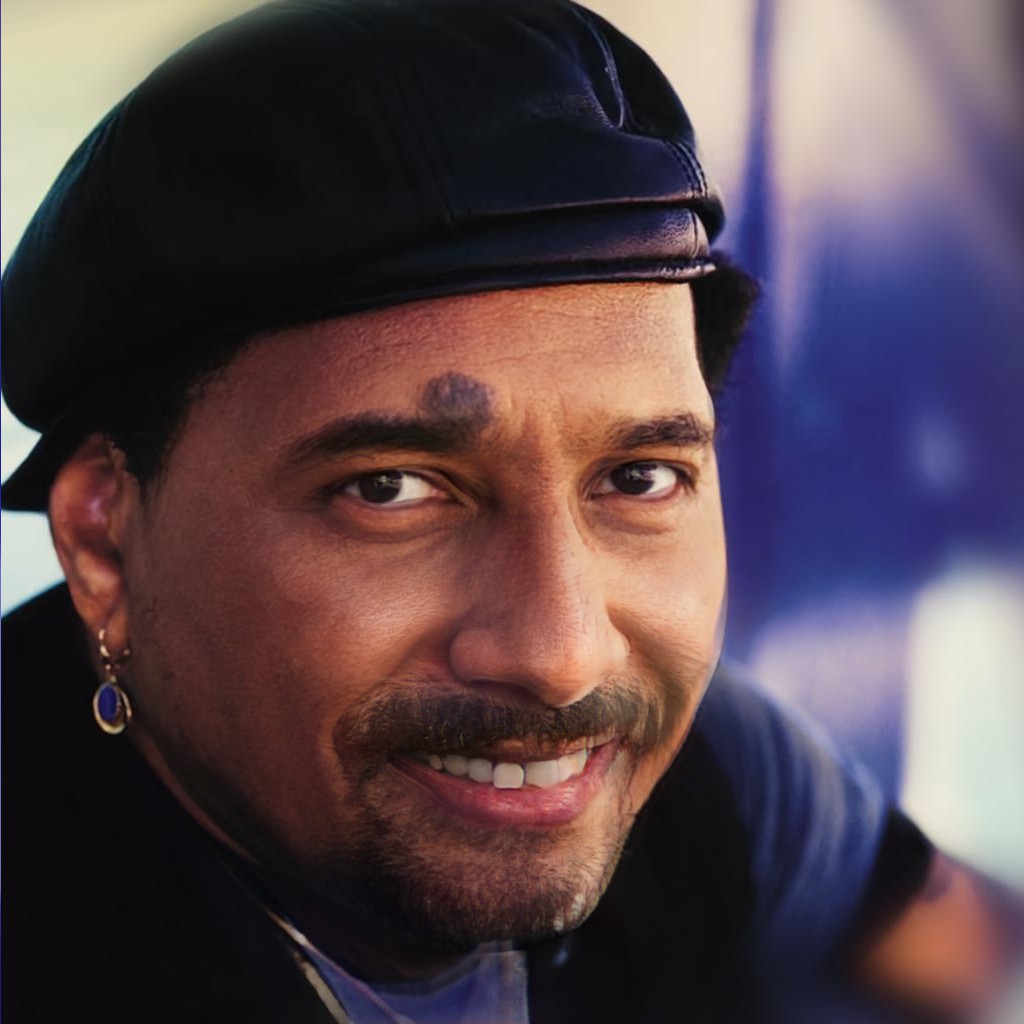}
		\caption{Composition}
		\label{fig:e}
	\end{subfigure}
	\begin{subfigure}[t]{0.23\linewidth}
		\centering
		\includegraphics[width=\textwidth]{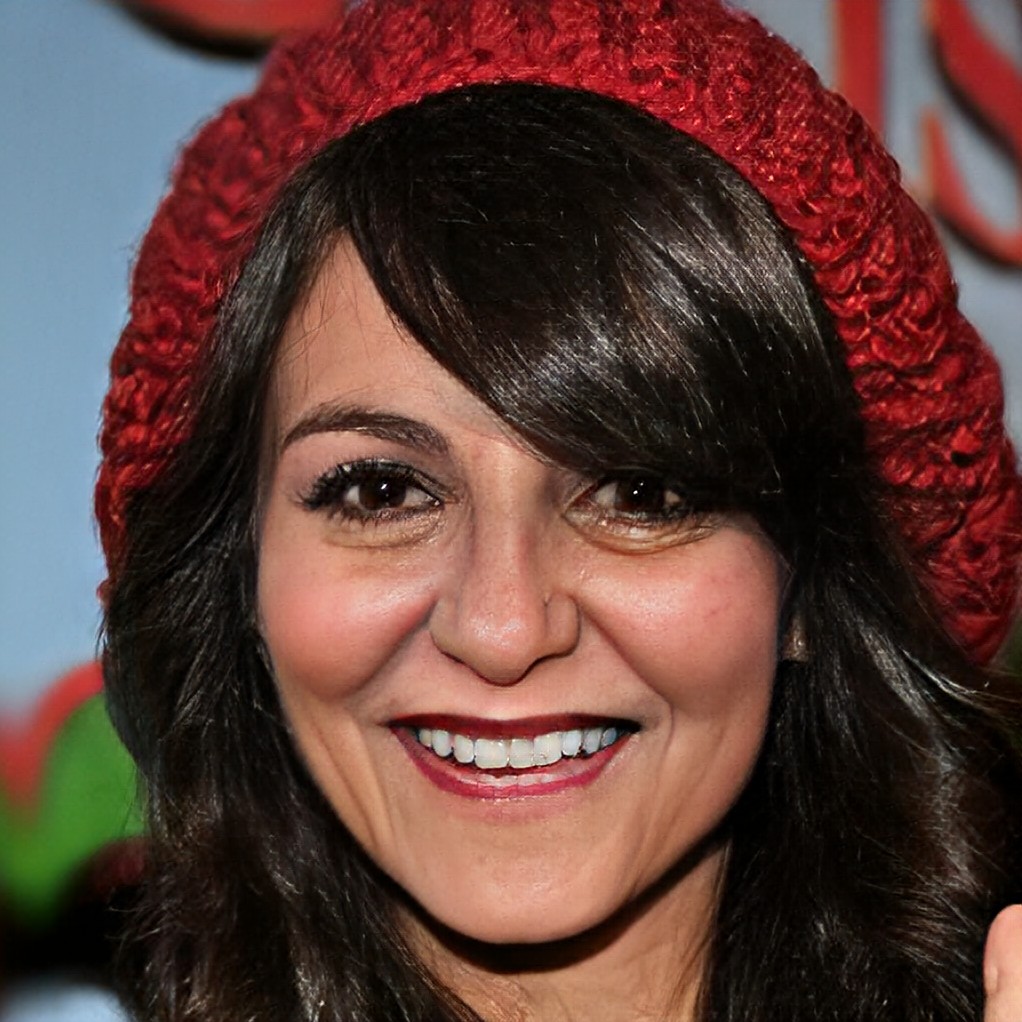}
		\includegraphics[width=\textwidth]{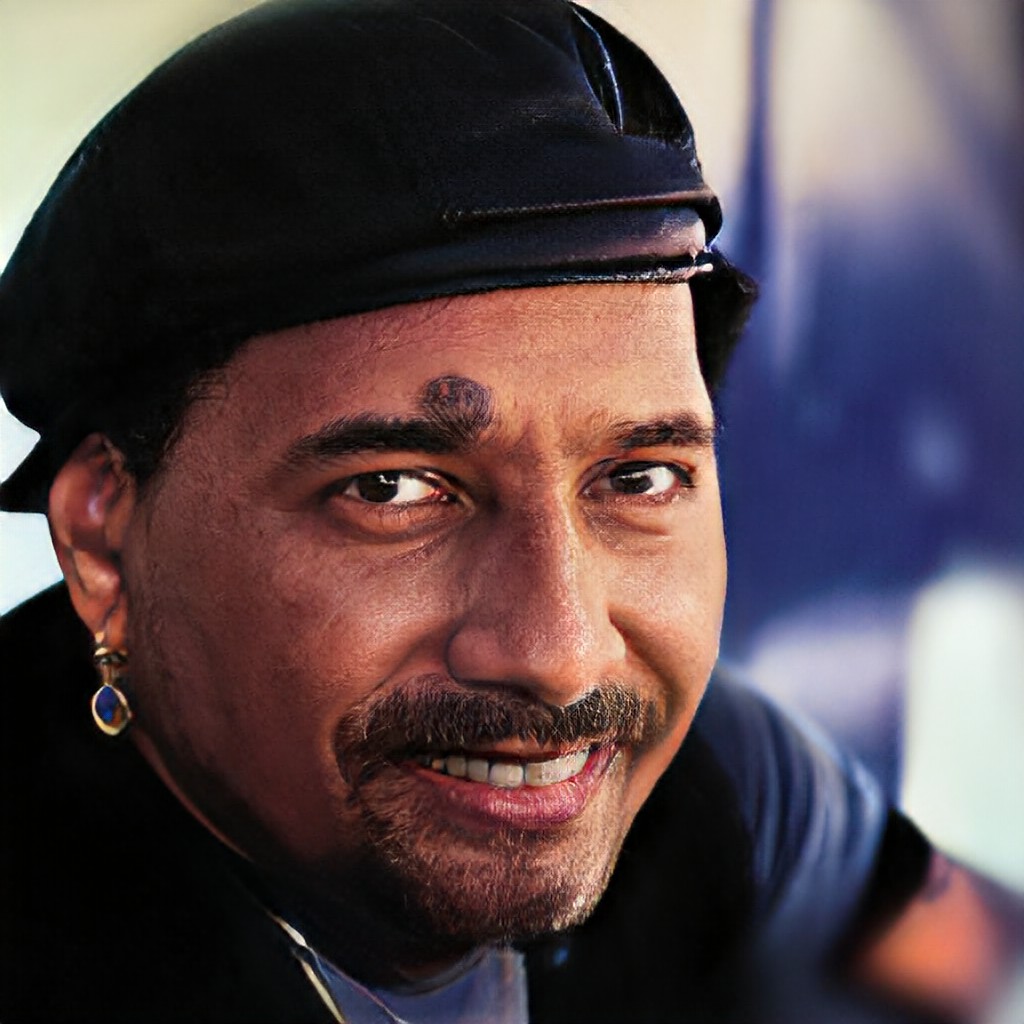}
		\caption{Deghosting}
		\label{fig:f}
	\end{subfigure}
	\vspace{-2mm}\caption{Effect of our deghosting network. Directly blending two images with a mask inevitably presents ghosting artifacts (teeth and face shape, zoom in for better view). Our deghosting network utilizes GAN priors to faithfully remove artifacts while retaining facial details.}\vspace{-6mm}
	\label{fig:component-intermediate}
\end{figure}

\textbf{Effectiveness of Deghosting Network.} Here we show the results before and after deghosting in Fig.~\ref{fig:component-intermediate}. Even with a correctly detected mask, blending two images inevitably produces blurry and ghosting artifacts. Our deghosting network takes advantage of the rich facial priors from the pretrained GAN model, and effectively removes non-face artifacts as well as generating a clear blending of faces.

\begin{figure}[t]
	\centering
	\footnotesize
	\setlength{\abovecaptionskip}{0cm}
	\centering
	\setlength{\tabcolsep}{0.05em}
	\begin{tabular}{cccc:ccc}
		\rotatebox[origin=lc]{90}{\hspace{3mm} Input}&
		\includegraphics[width=.16\linewidth]{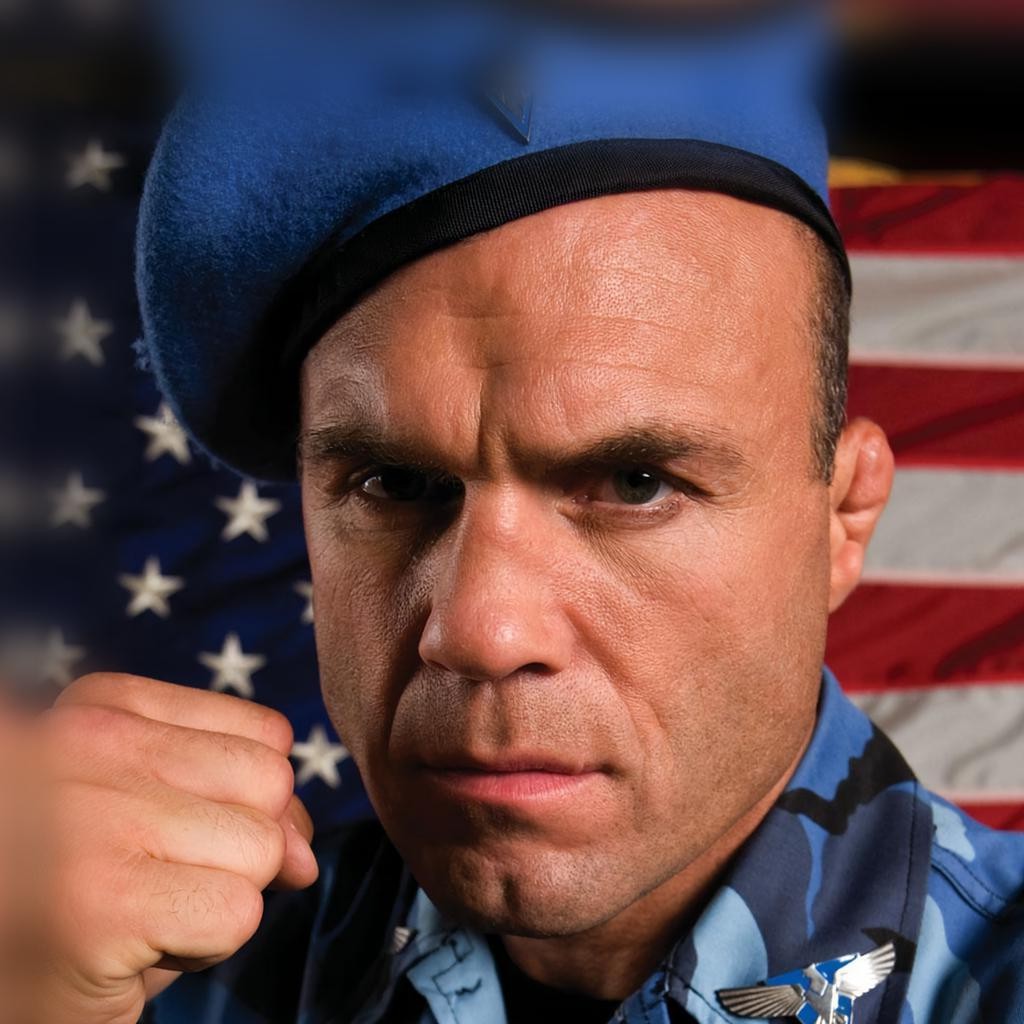}&&&
		\includegraphics[width=.16\linewidth]{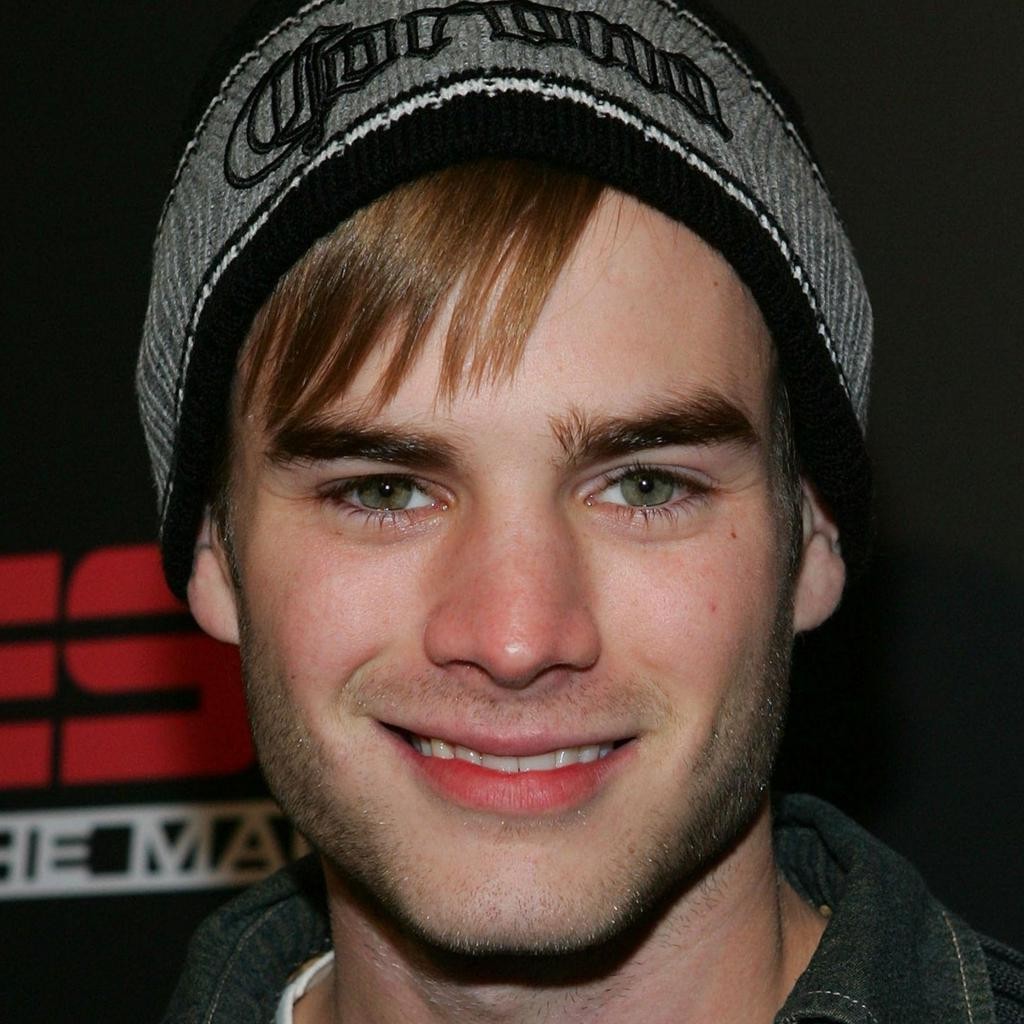}&&\\		
		\rotatebox[origin=lc]{90}{\hspace{1mm}Inversion}&
		\includegraphics[width=.16\linewidth]{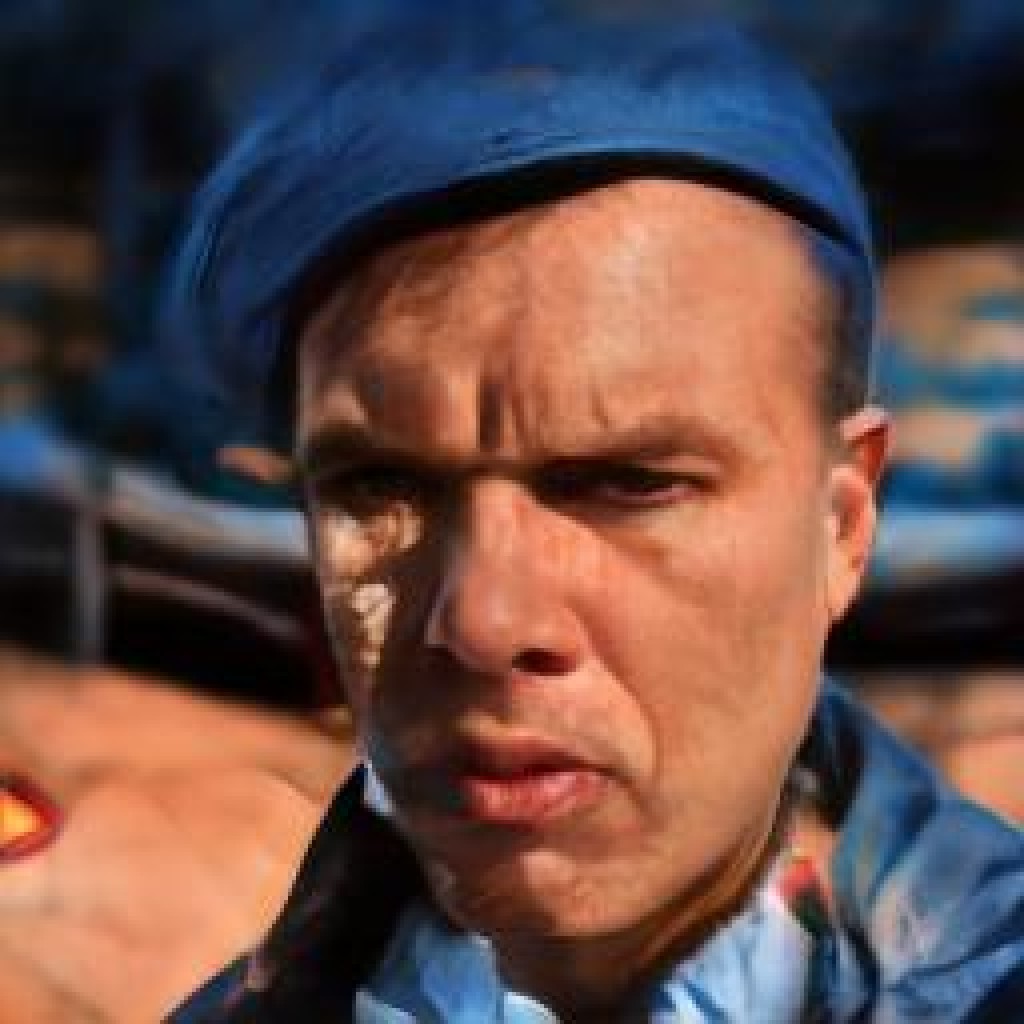}&
	    \includegraphics[width=.16\linewidth]{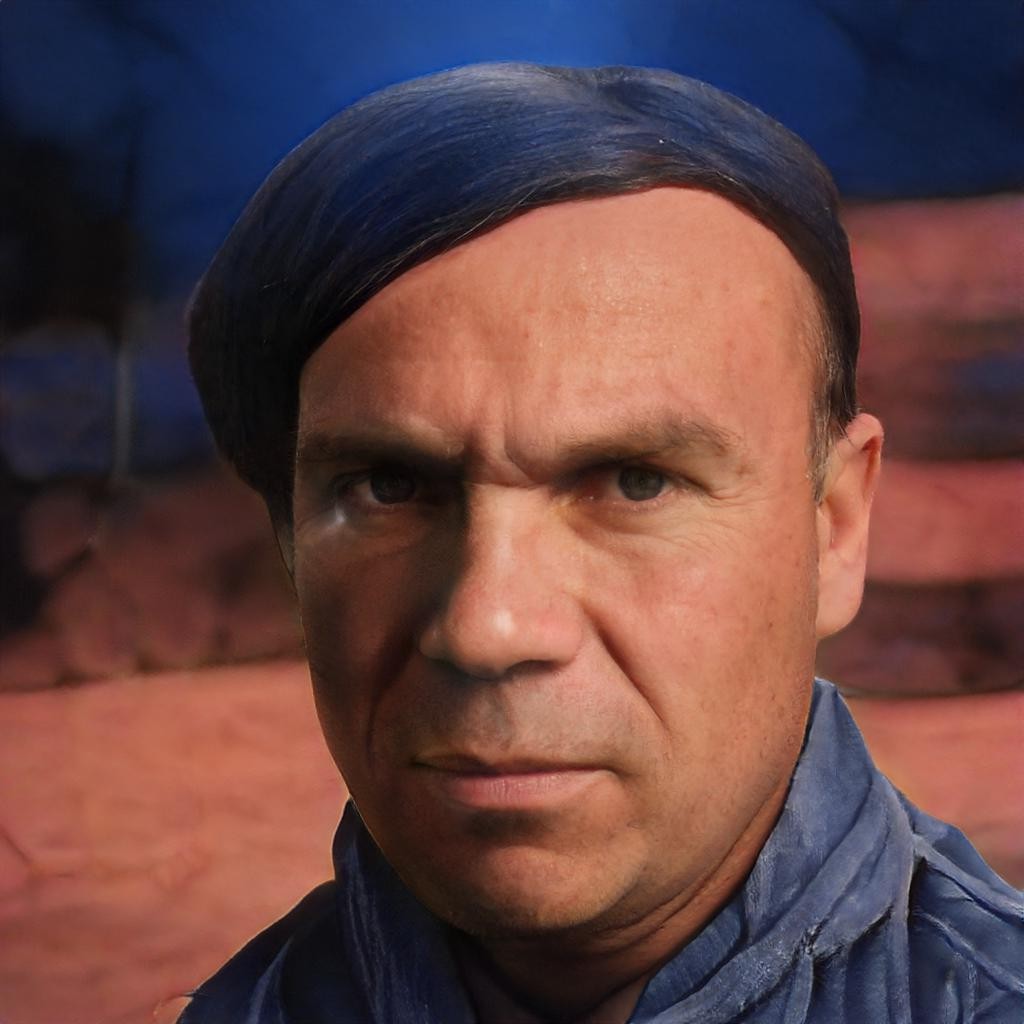}&
		\includegraphics[width=.16\linewidth]{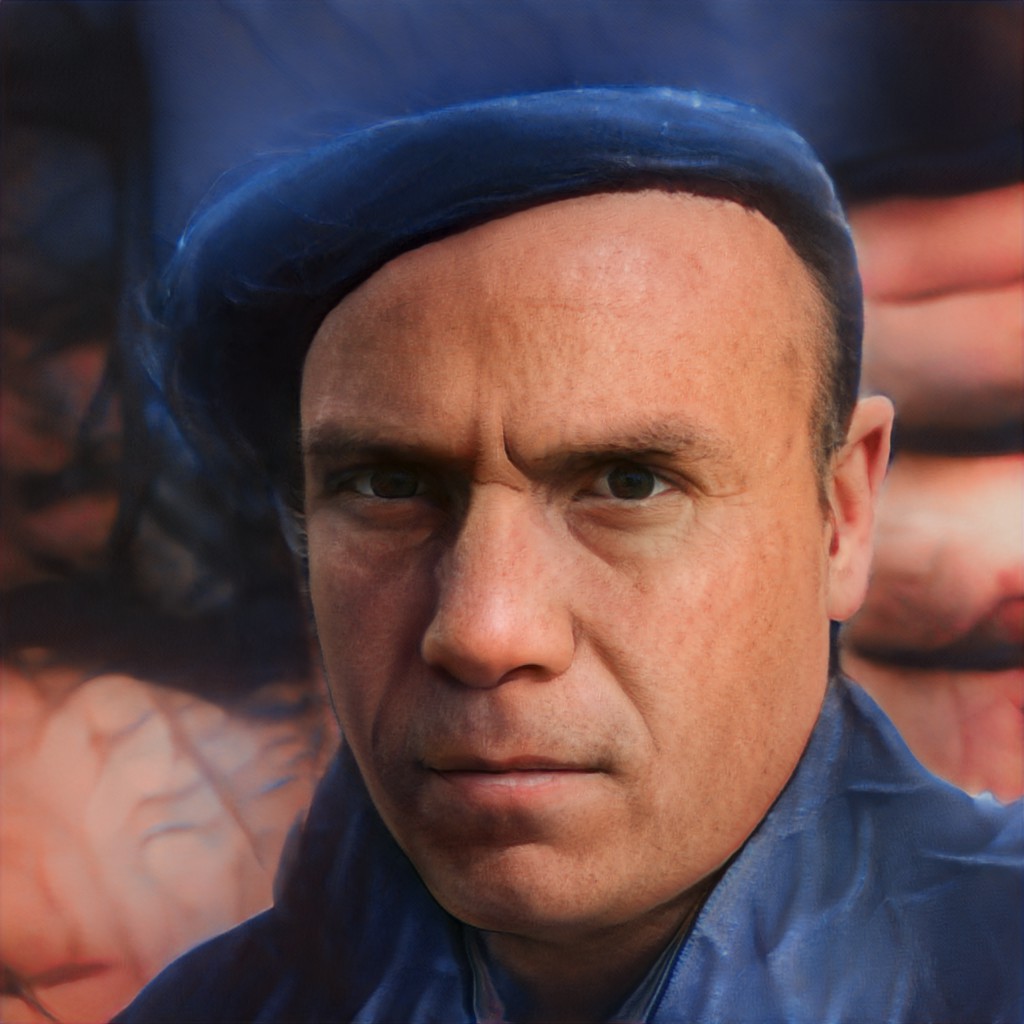}&
		\includegraphics[width=.16\linewidth]{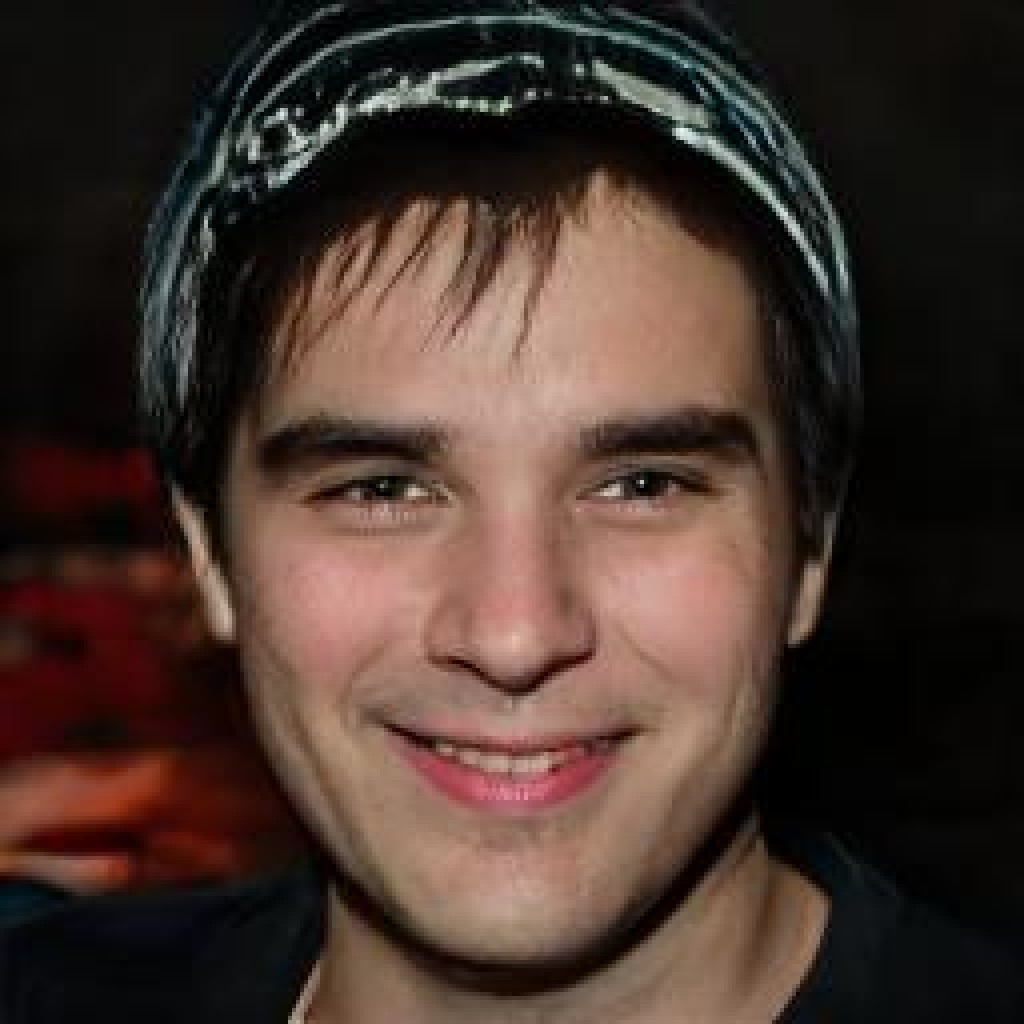}&
		\includegraphics[width=.16\linewidth]{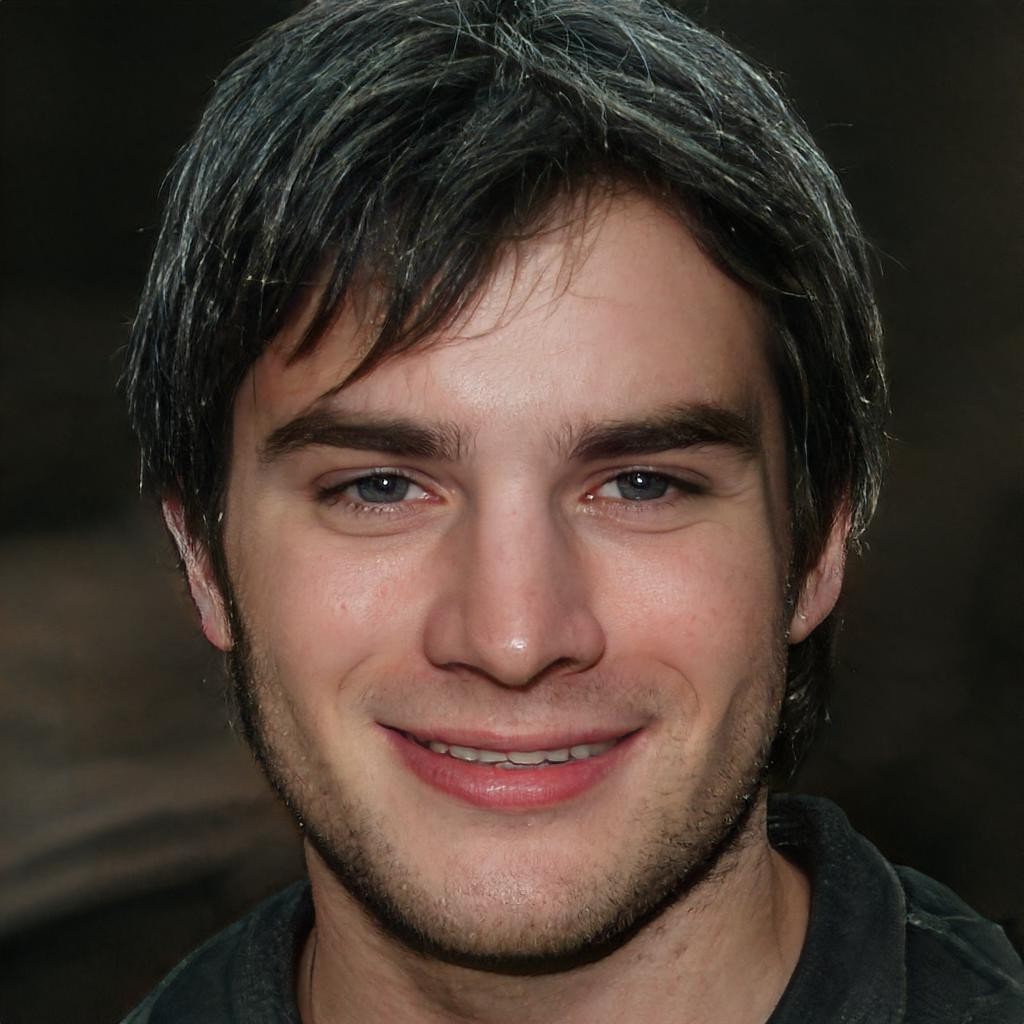}&
		\includegraphics[width=.16\linewidth]{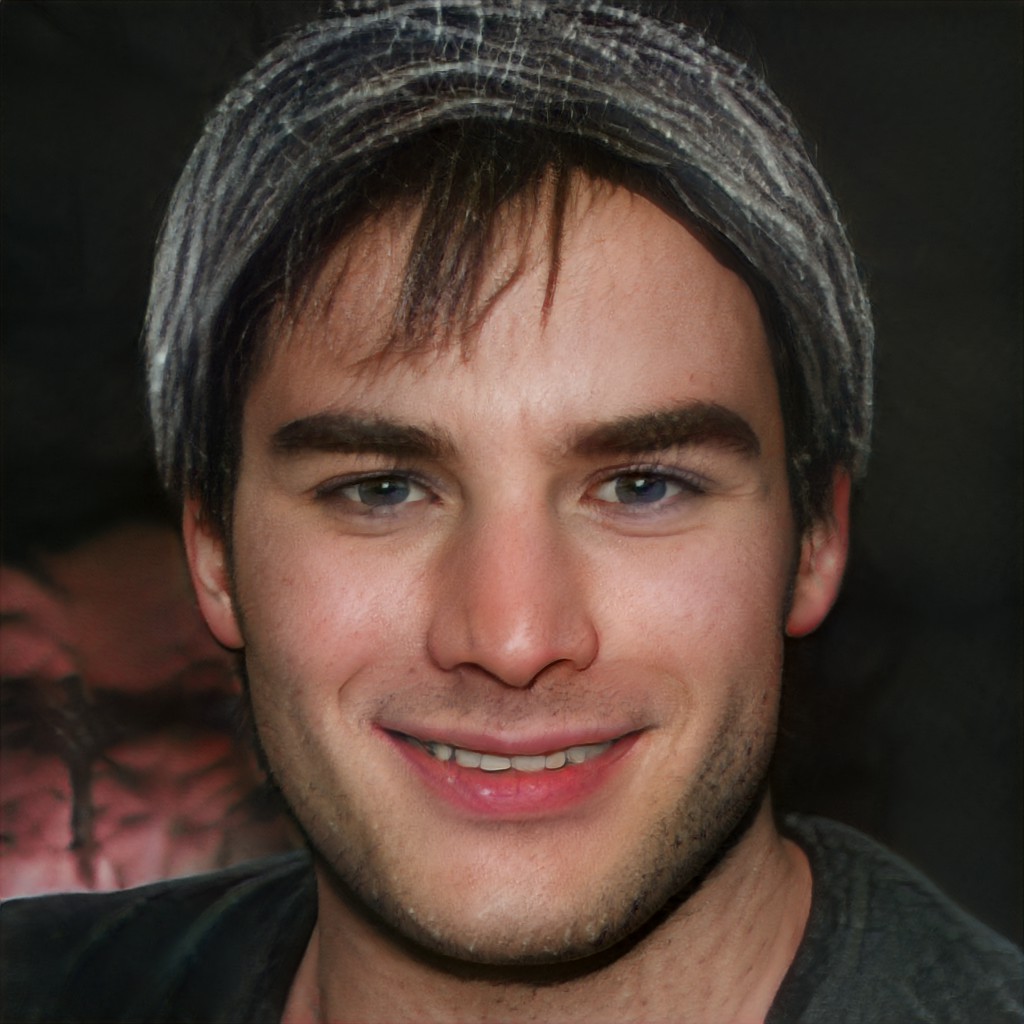}\\
		\rotatebox[origin=lc]{90}{\hspace{1mm} Editing}&
		\includegraphics[width=.16\linewidth]{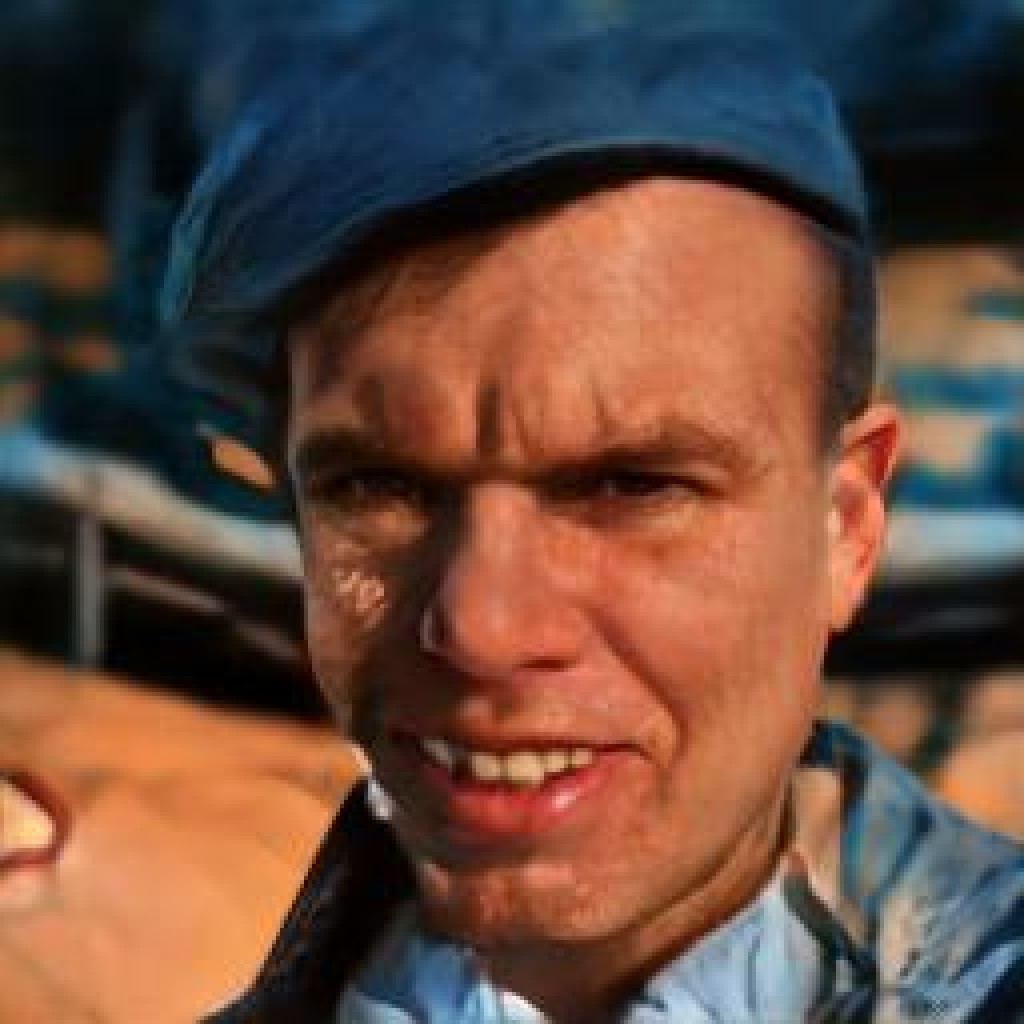}&
		\includegraphics[width=.16\linewidth]{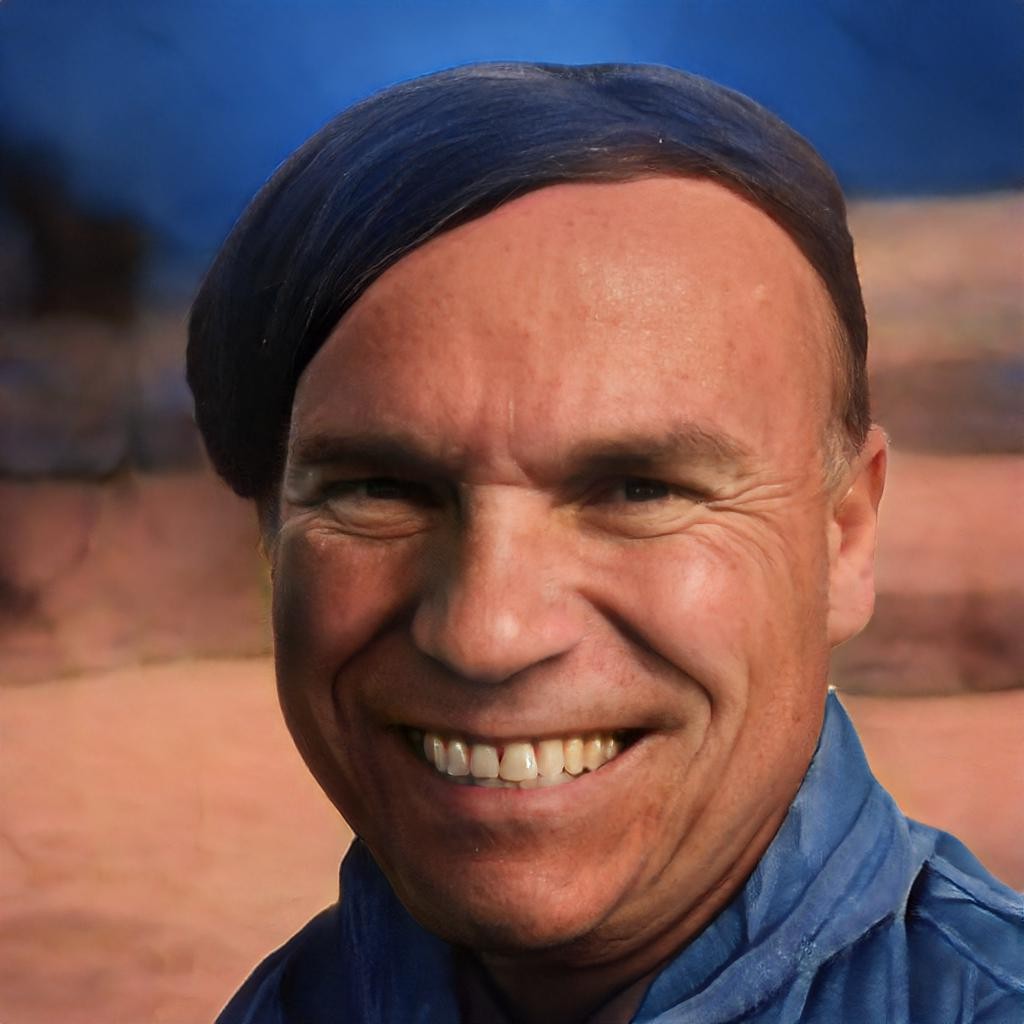}&
		\includegraphics[width=.16\linewidth]{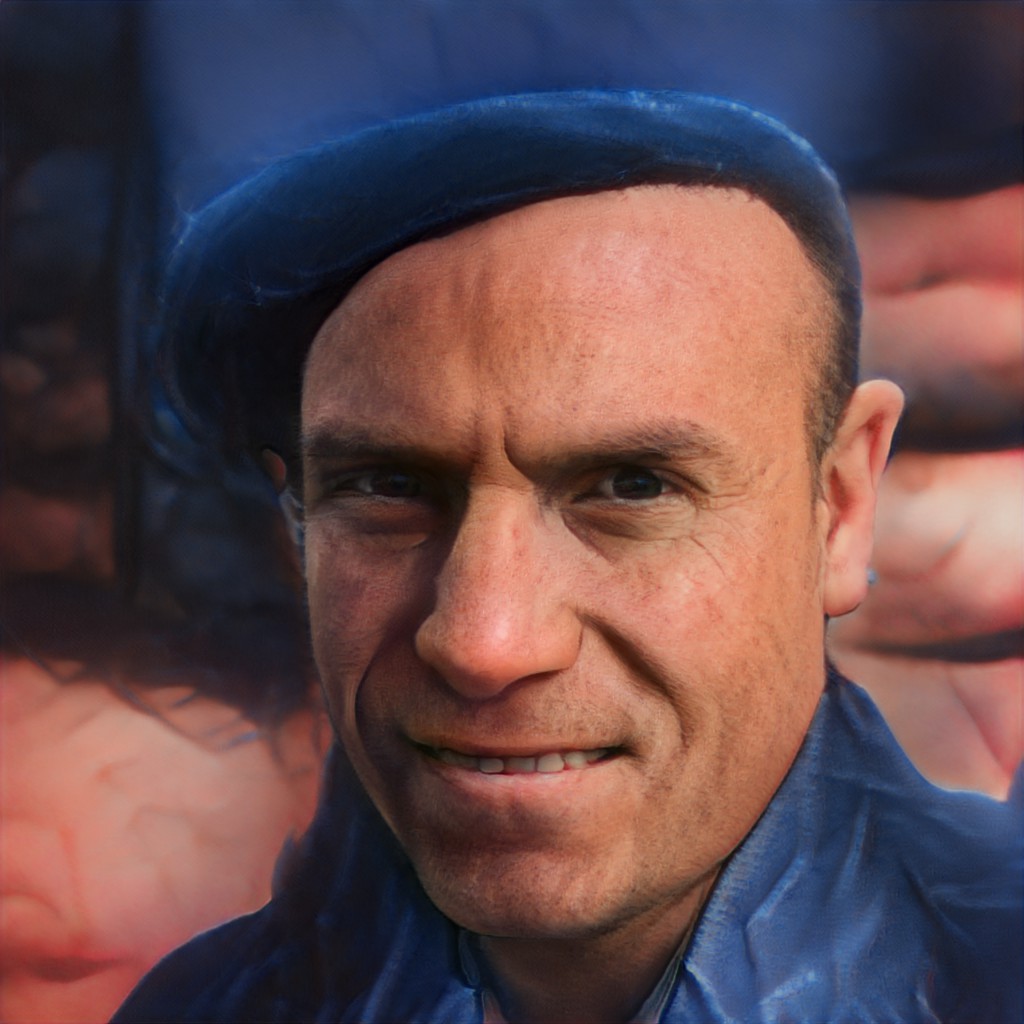}&
		\includegraphics[width=.16\linewidth]{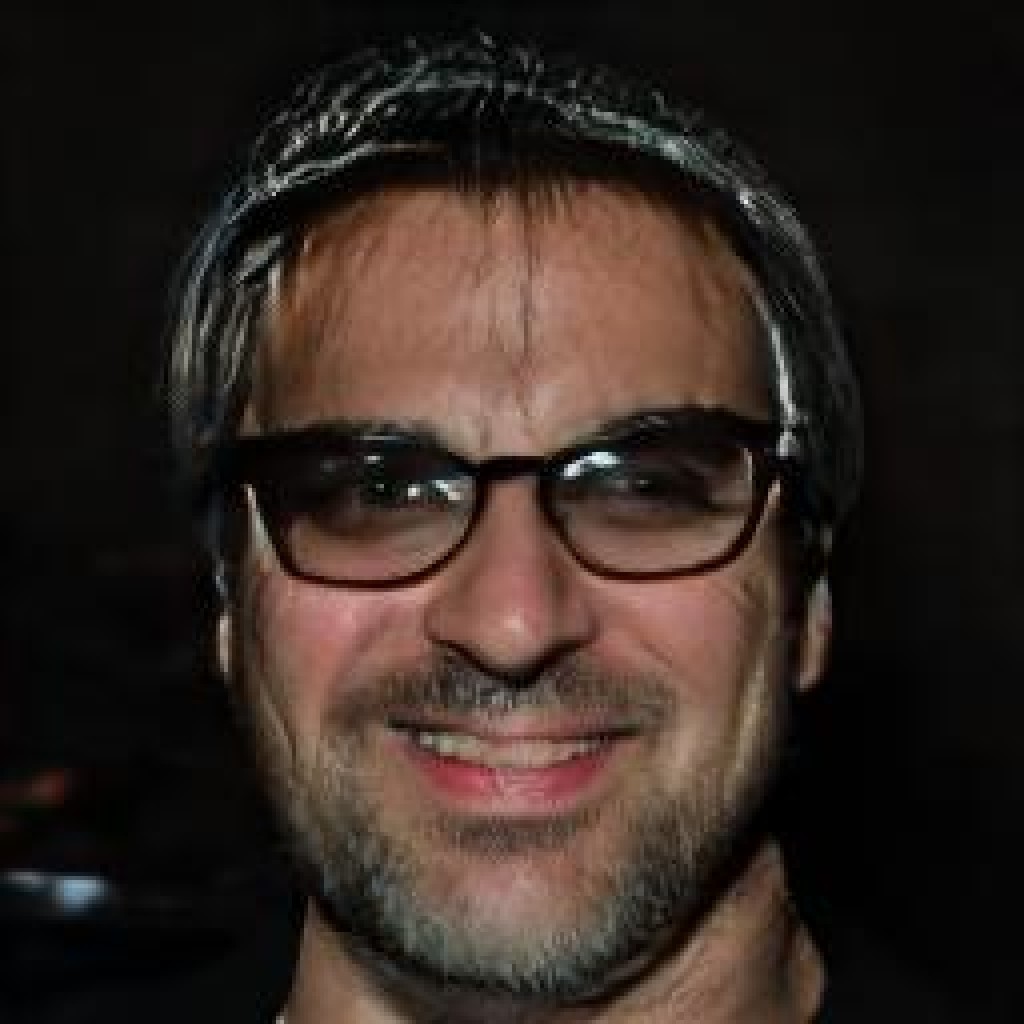}&
		\includegraphics[width=.16\linewidth]{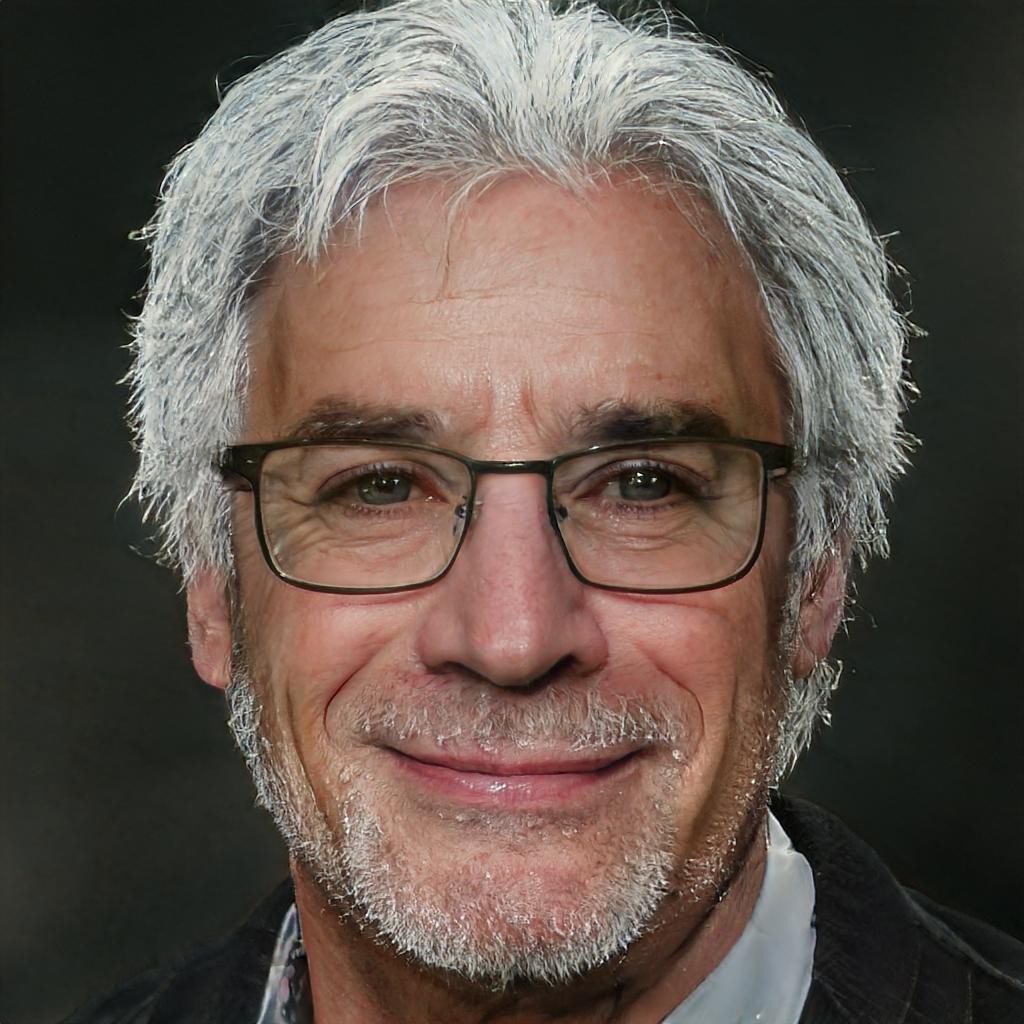}&
		\includegraphics[width=.16\linewidth]{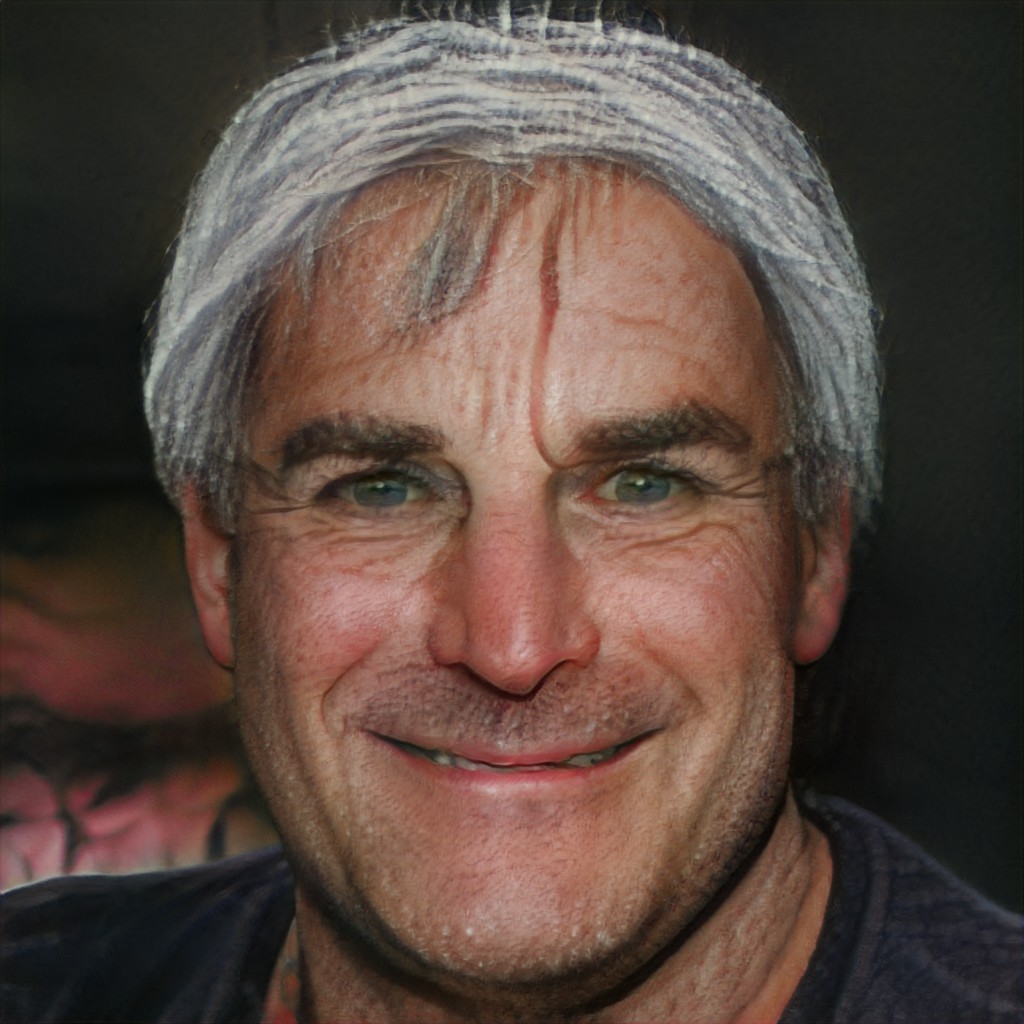}\\
		\rotatebox[origin=lc]{90}{\hspace{3mm} Ours}&
		\includegraphics[width=.16\linewidth]{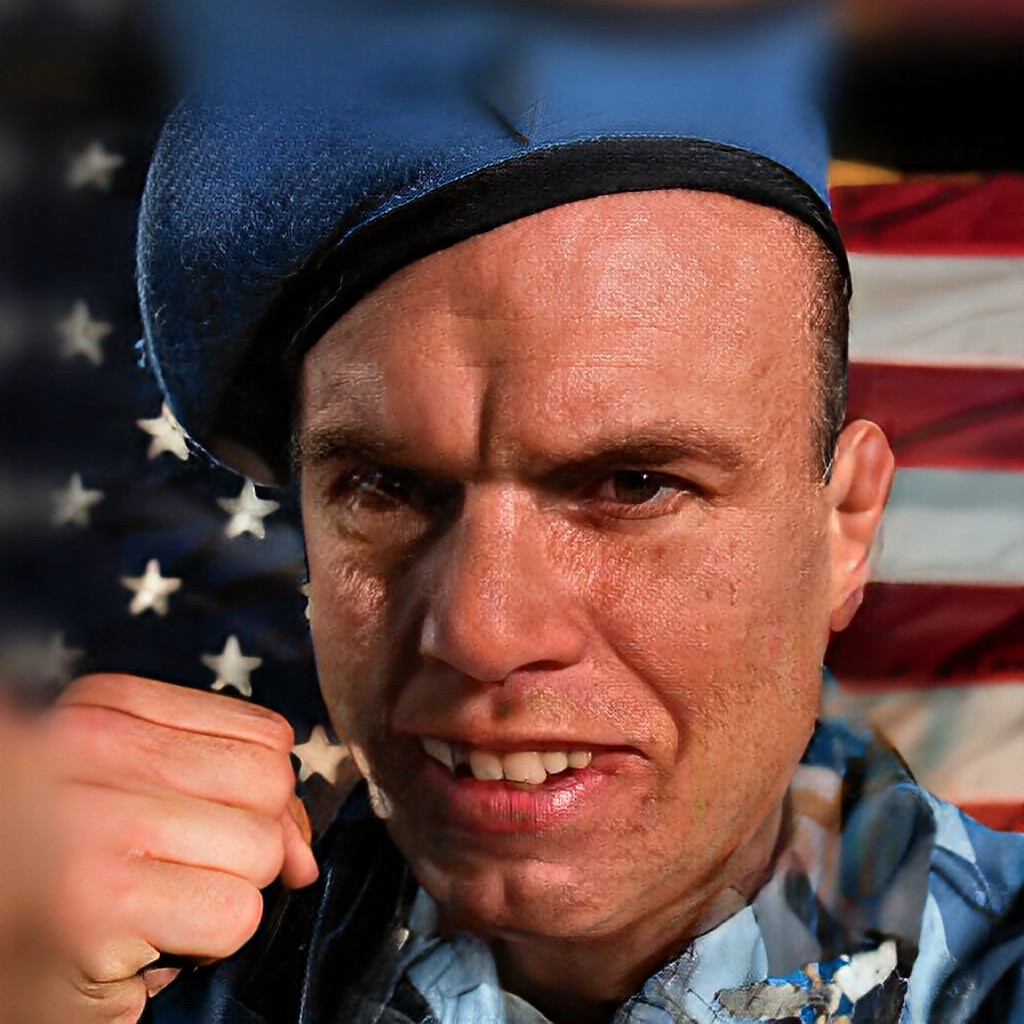}&
		\includegraphics[width=.16\linewidth]{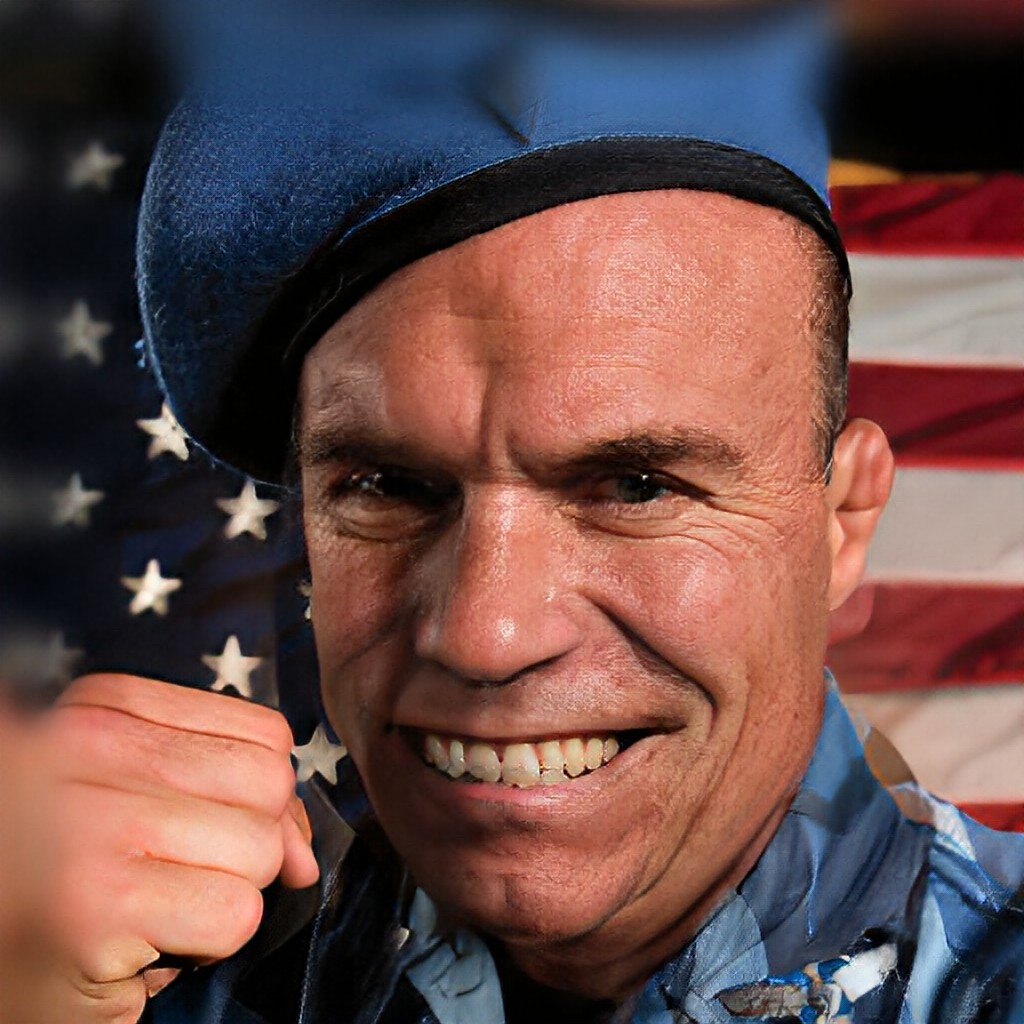}&
		\includegraphics[width=.16\linewidth]{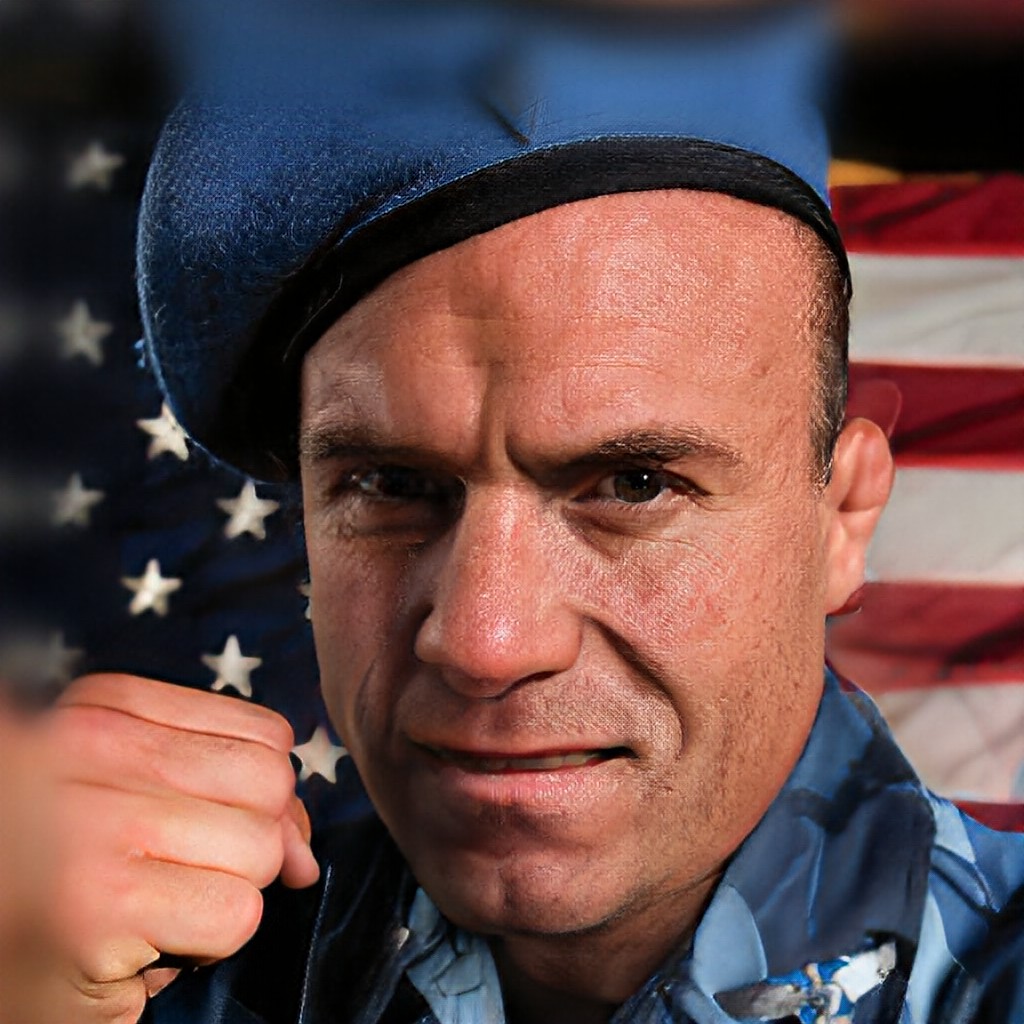}&
		\includegraphics[width=.16\linewidth]{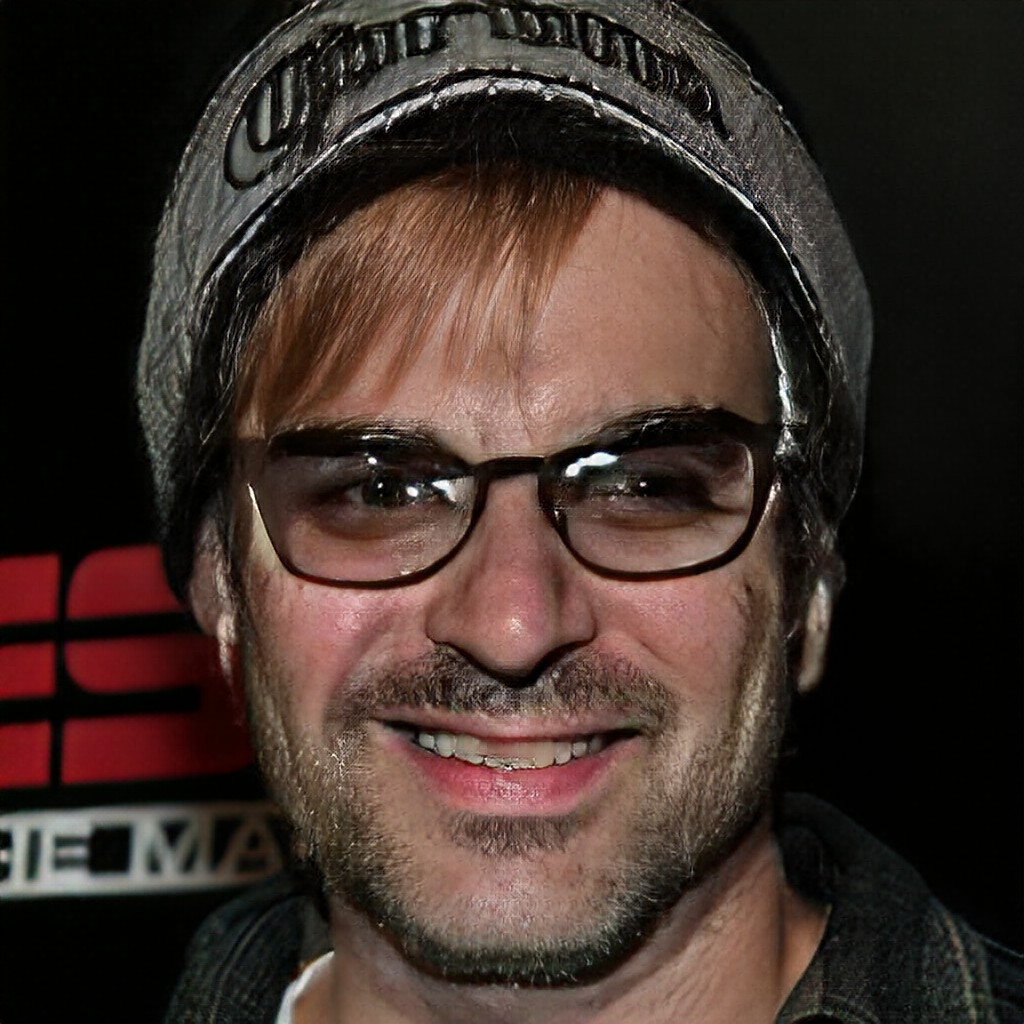}&
		\includegraphics[width=.16\linewidth]{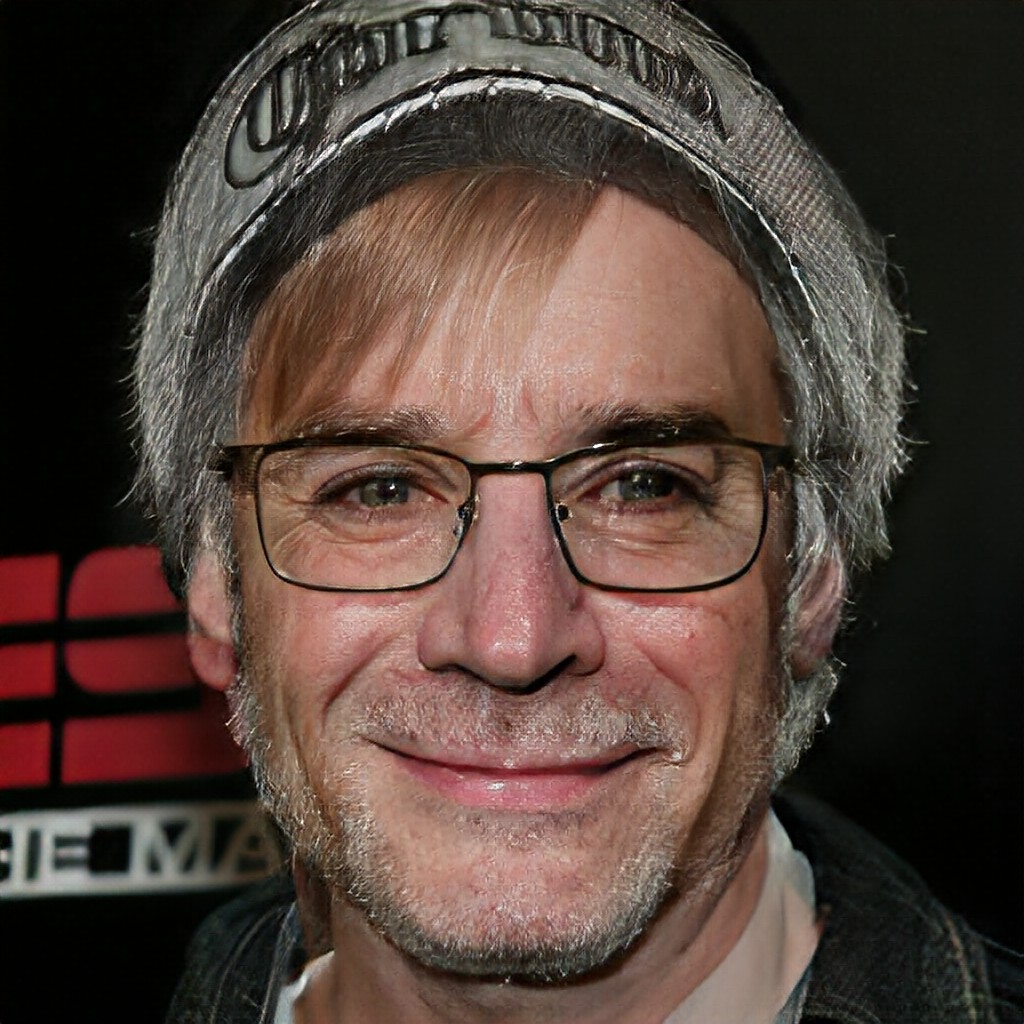}&
		\includegraphics[width=.16\linewidth]{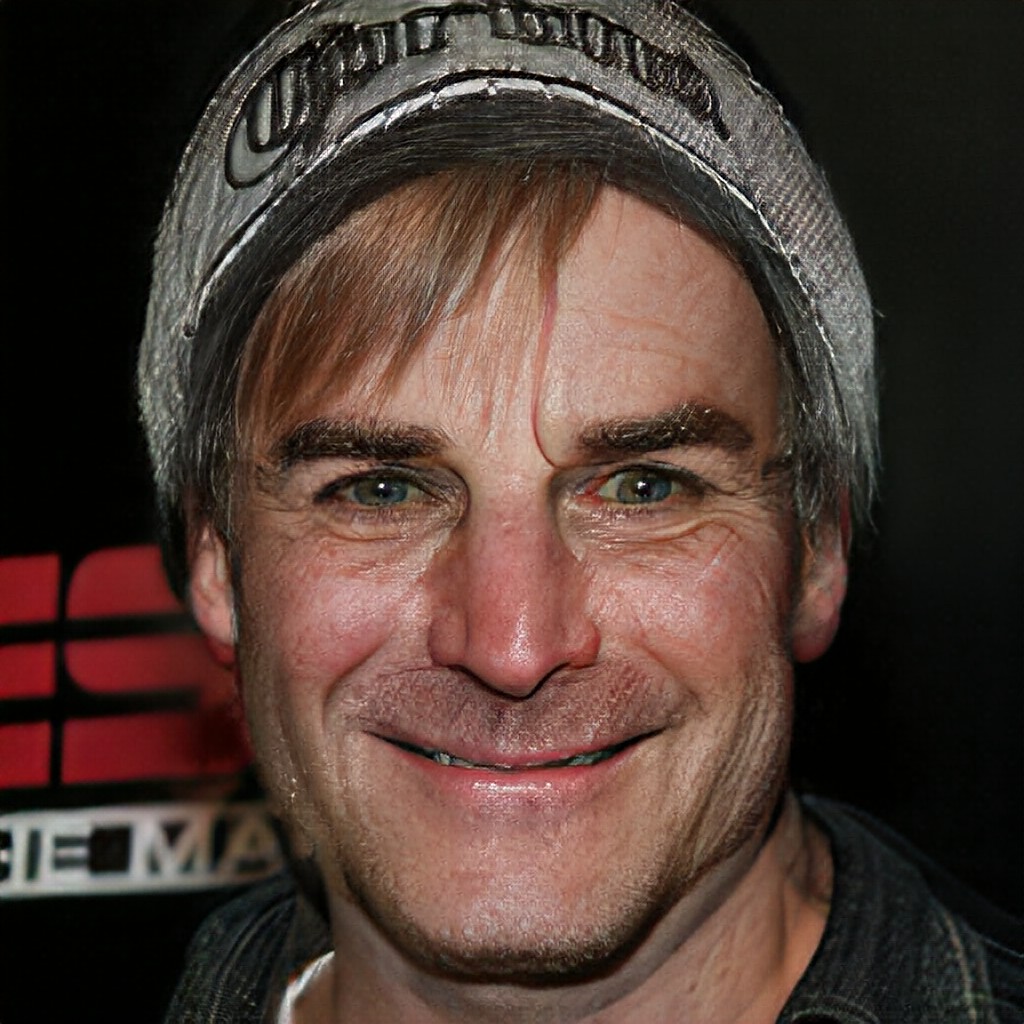}\\
		&\scriptsize IdInv. + IFG&\scriptsize e4e + SFlow&\scriptsize pSp + Distill&\scriptsize IdInv. + IFG&\scriptsize e4e + SFlow&\scriptsize pSp + Distill\\	
	\end{tabular}
	\vspace{2mm}\caption{Our framework is independent to GAN inversion and interpretable editing directions. We can work with arbitrary combinations of encoders and directions. (Zoom in for better view.)}\vspace{-6mm}
	\label{fig:compare with different encoders}
\end{figure}

\textbf{Flexibility Analysis.} Our method is flexible and independent of the applied GAN inversion and interpretable directions. We choose several different inversion and interpretable direction models to work with our framework. Three combinations of inversion and direction methods are used, \ie, the pSp encoder~\cite{richardson2020encoding} together with directions found by StyleGAN2 distillation~\cite{viazovetskyi2020stylegan2}, the IdInvert encoder~\cite{zhu2020domain} with InterfaceGAN~\cite{shen2020interpreting, shen2020interfacegan}, and the e4e encoder~\cite{tov2021designing} together with the directions obtained by StyleFlow~\cite{abdal2021styleflow}.

\begin{figure*}[t]
	\centering
	\footnotesize
	\setlength{\abovecaptionskip}{0cm}
	\centering
	\setlength{\tabcolsep}{0.05em}
	\begin{tabular}{cccccccc}
		& Input & StarGAN & AttGAN & STGAN & StarGAN* & AttGAN* & Ours \\
		
		\rotatebox[origin=lc]{90}{Eyeglasses} &
		\includegraphics[width=.135\linewidth]{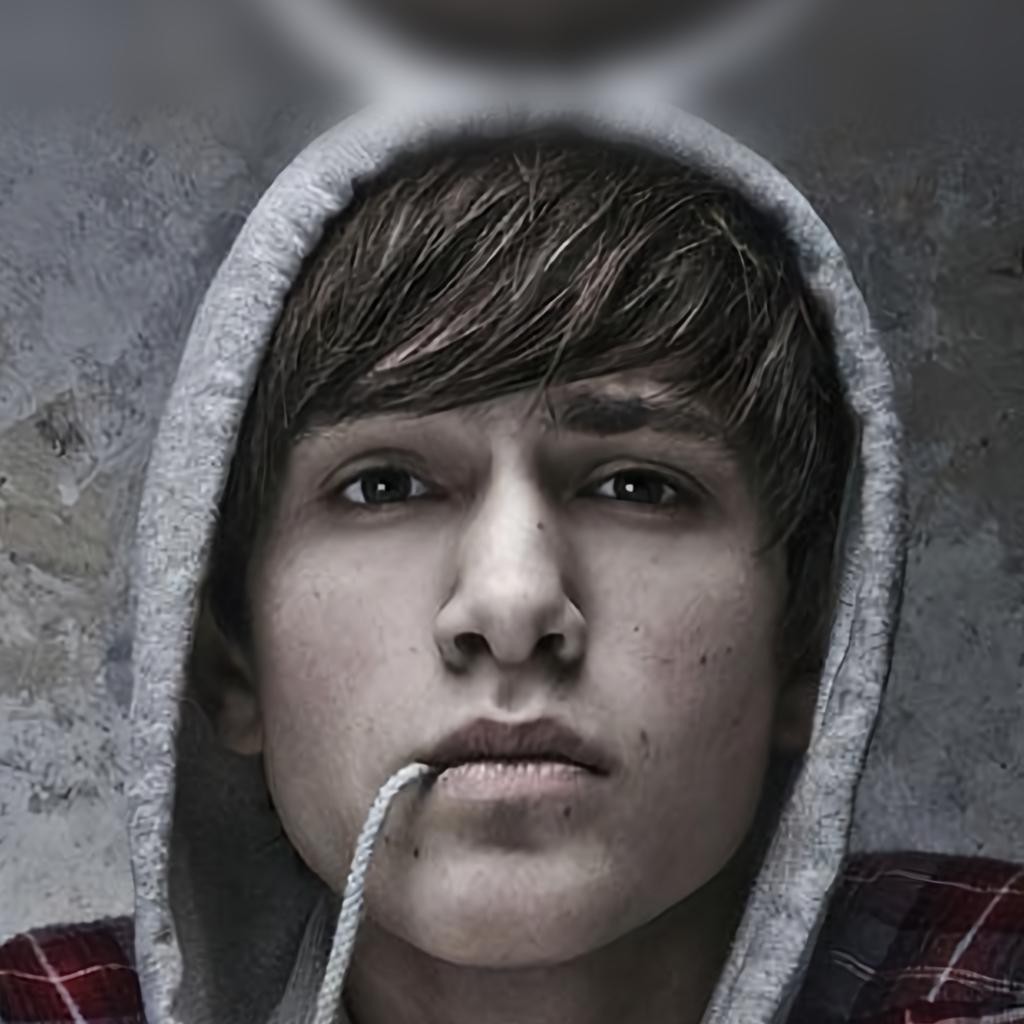} &
		\includegraphics[width=.135\linewidth]{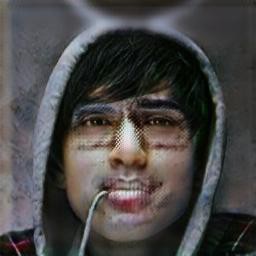} &
		\includegraphics[width=.135\linewidth]{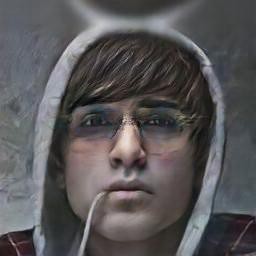} &
		\includegraphics[width=.135\linewidth]{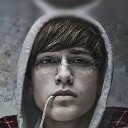} &
		\includegraphics[width=.135\linewidth]{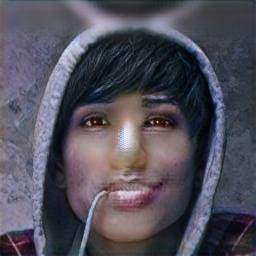} &
		\includegraphics[width=.135\linewidth]{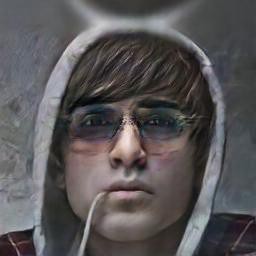} &
		\includegraphics[width=.135\linewidth]{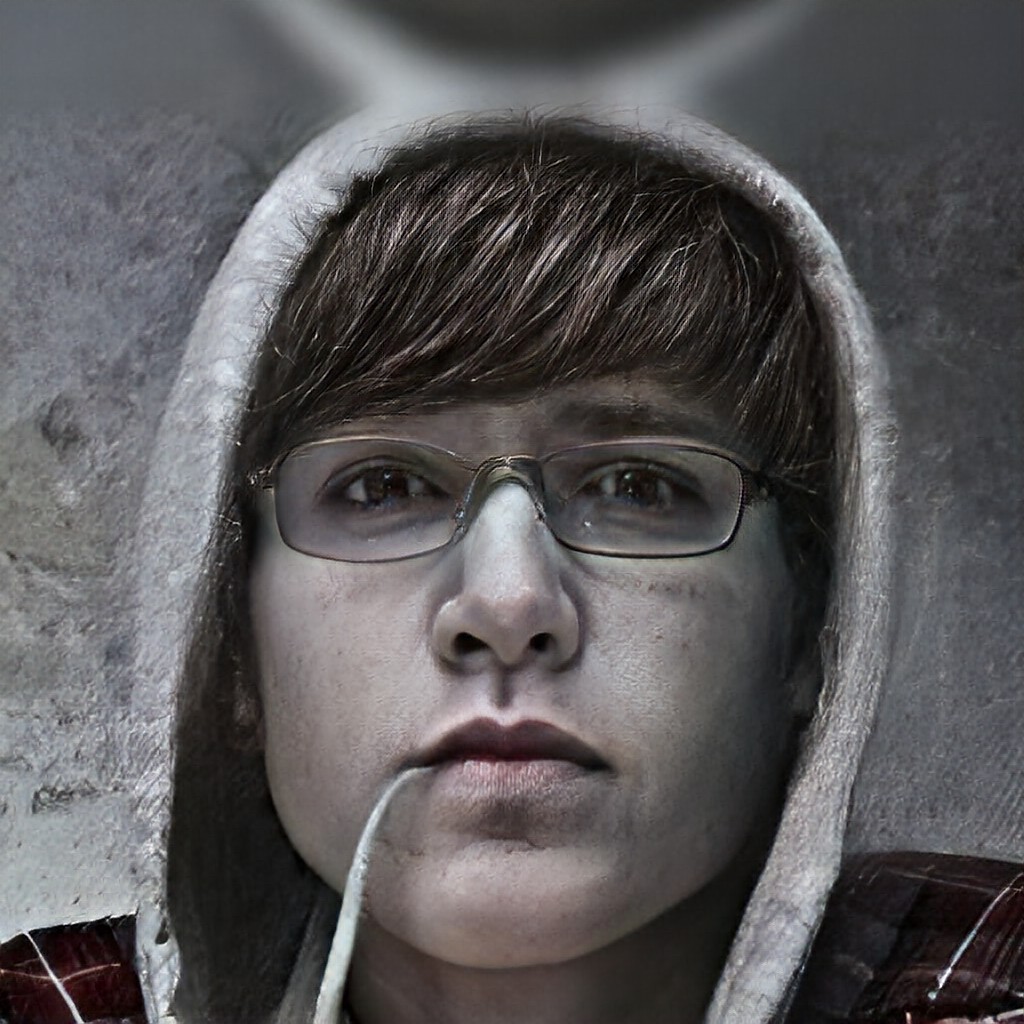}\\
		
		\rotatebox[origin=lc]{90}{\hspace{1mm}Eyebrows} &
		\includegraphics[width=.135\linewidth]{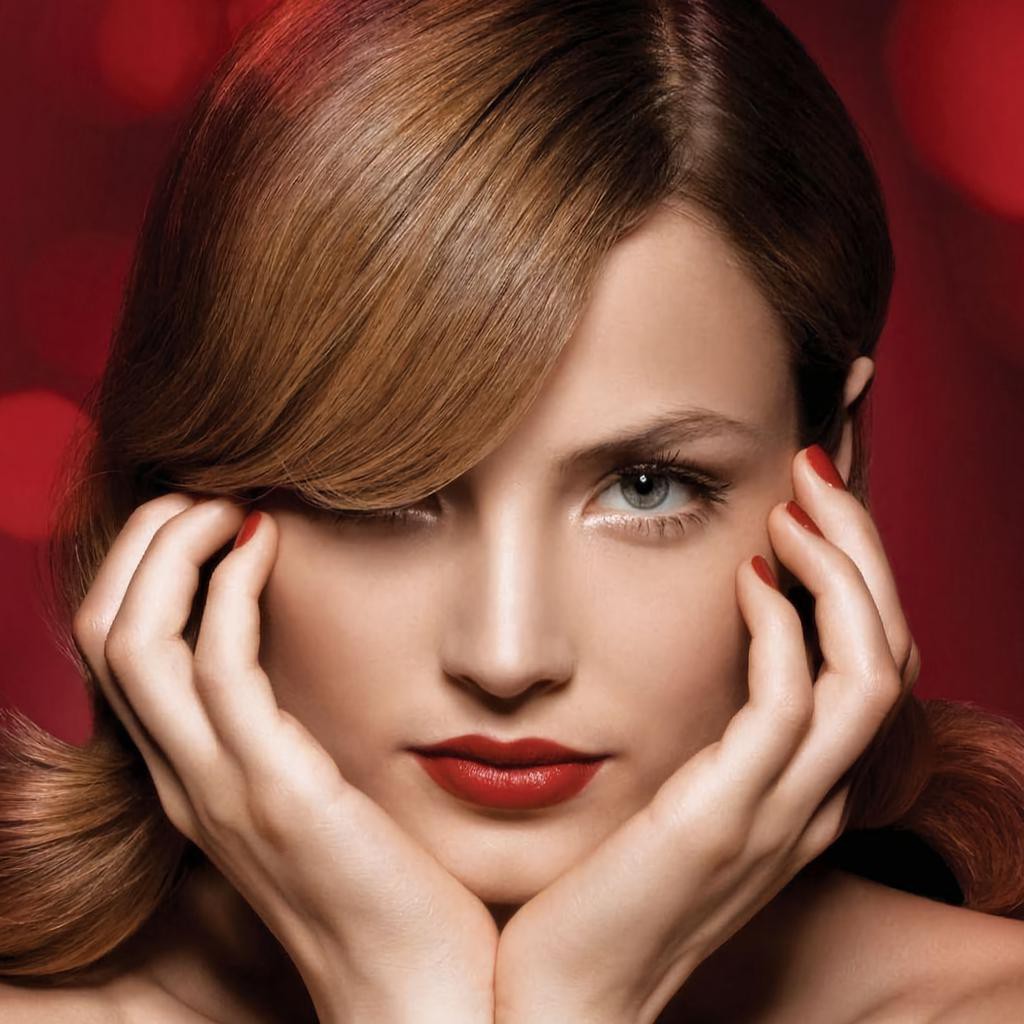} &
		\includegraphics[width=.135\linewidth]{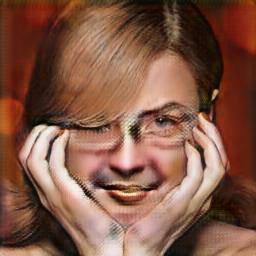} &
		\includegraphics[width=.135\linewidth]{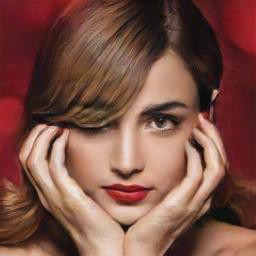} &
		\includegraphics[width=.135\linewidth]{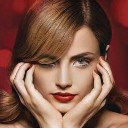} &
		\includegraphics[width=.135\linewidth]{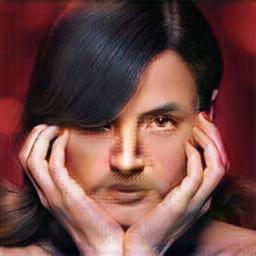} &
		\includegraphics[width=.135\linewidth]{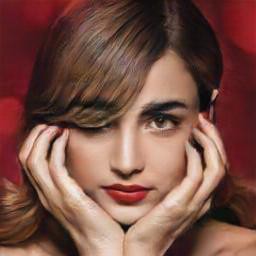} &
		\includegraphics[width=.135\linewidth]{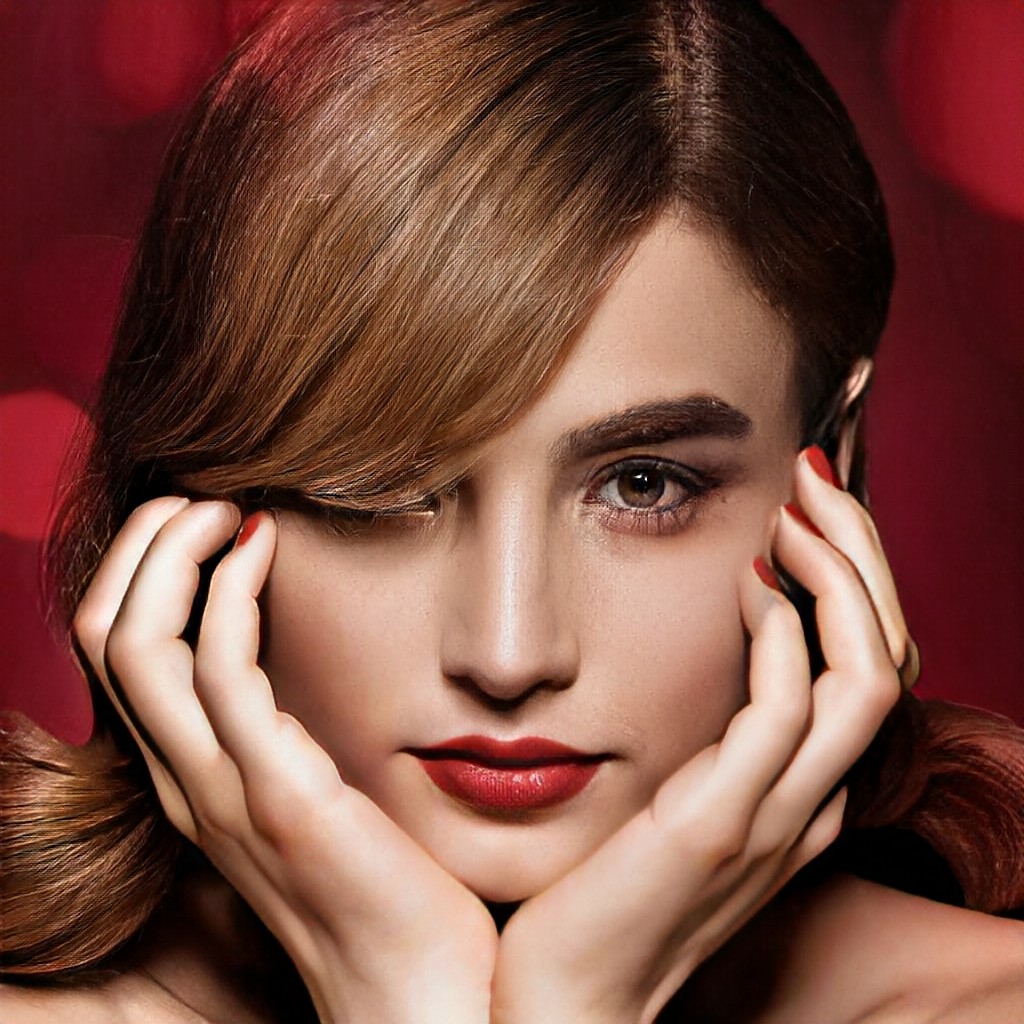}\\
		
		\rotatebox[origin=lc]{90}{\hspace{1mm} Beard} &
		\includegraphics[width=.135\linewidth]{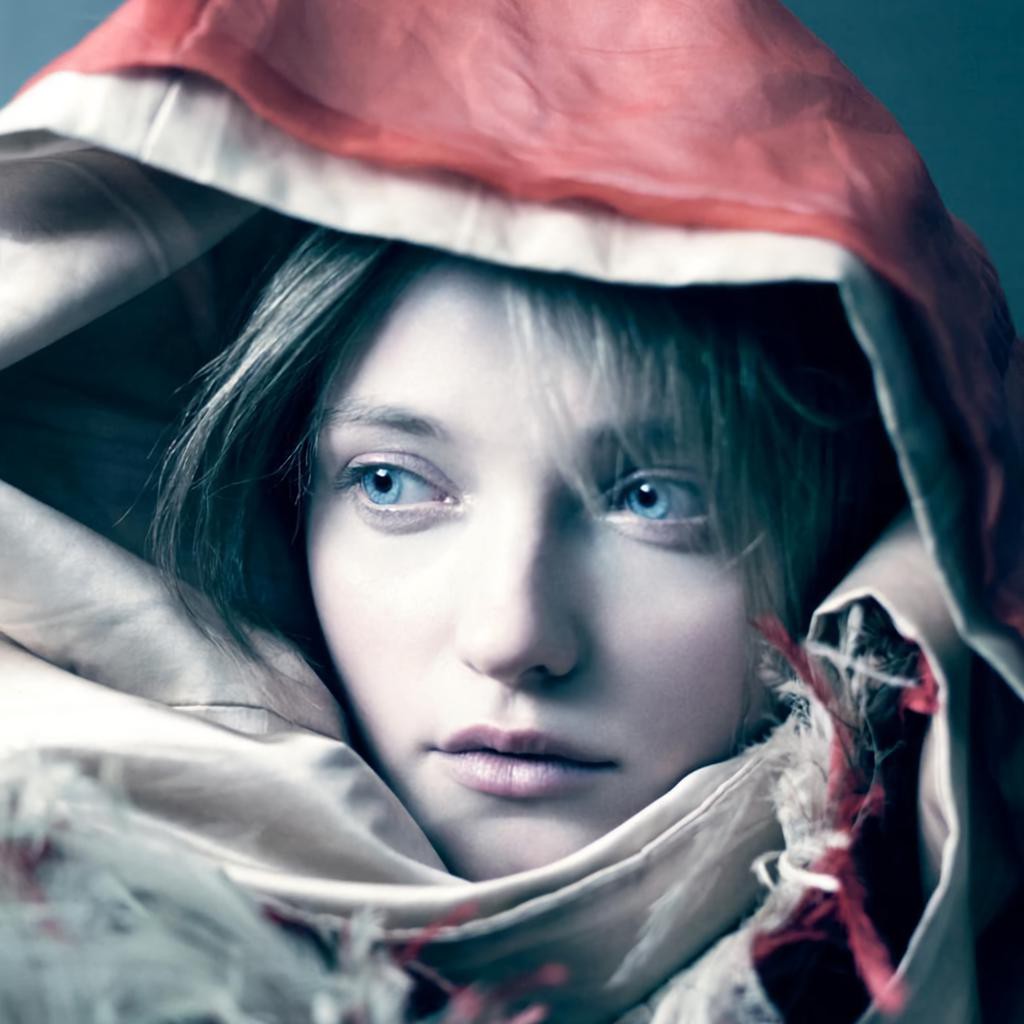} &
		\includegraphics[width=.135\linewidth]{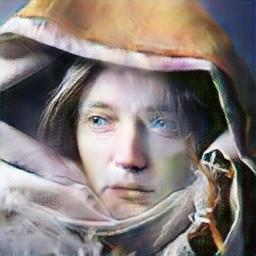} &
		\includegraphics[width=.135\linewidth]{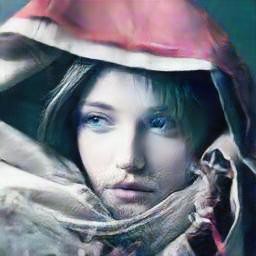} &
		\includegraphics[width=.135\linewidth]{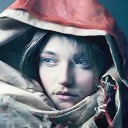} &
		\includegraphics[width=.135\linewidth]{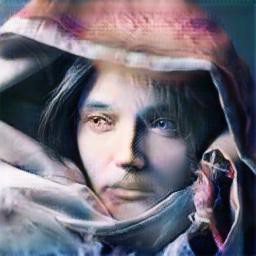} &
		\includegraphics[width=.135\linewidth]{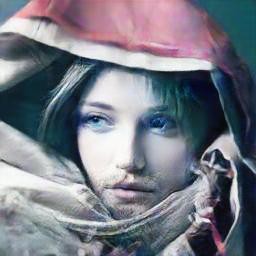} &
		\includegraphics[width=.135\linewidth]{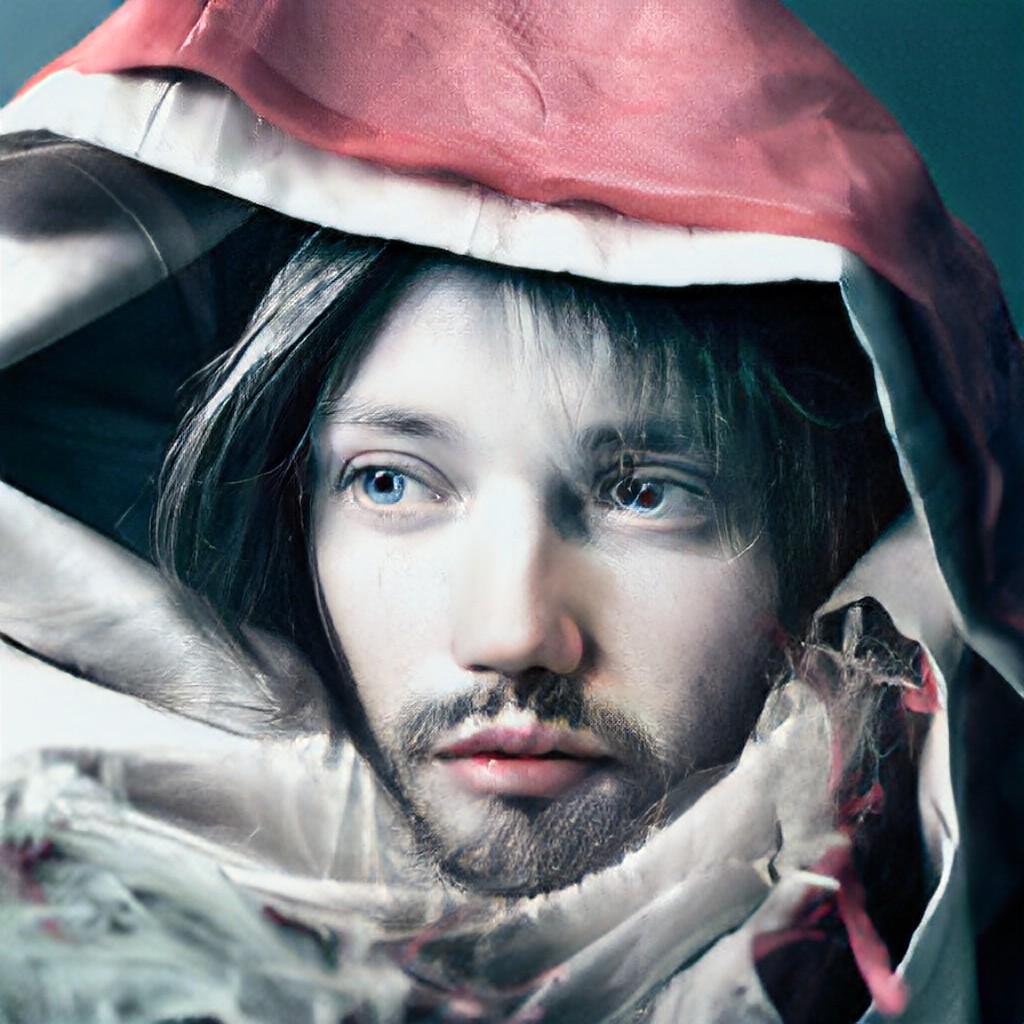} \\
		
		\rotatebox[origin=lc]{90}{\hspace{3mm} Age} &
		\includegraphics[width=.135\linewidth]{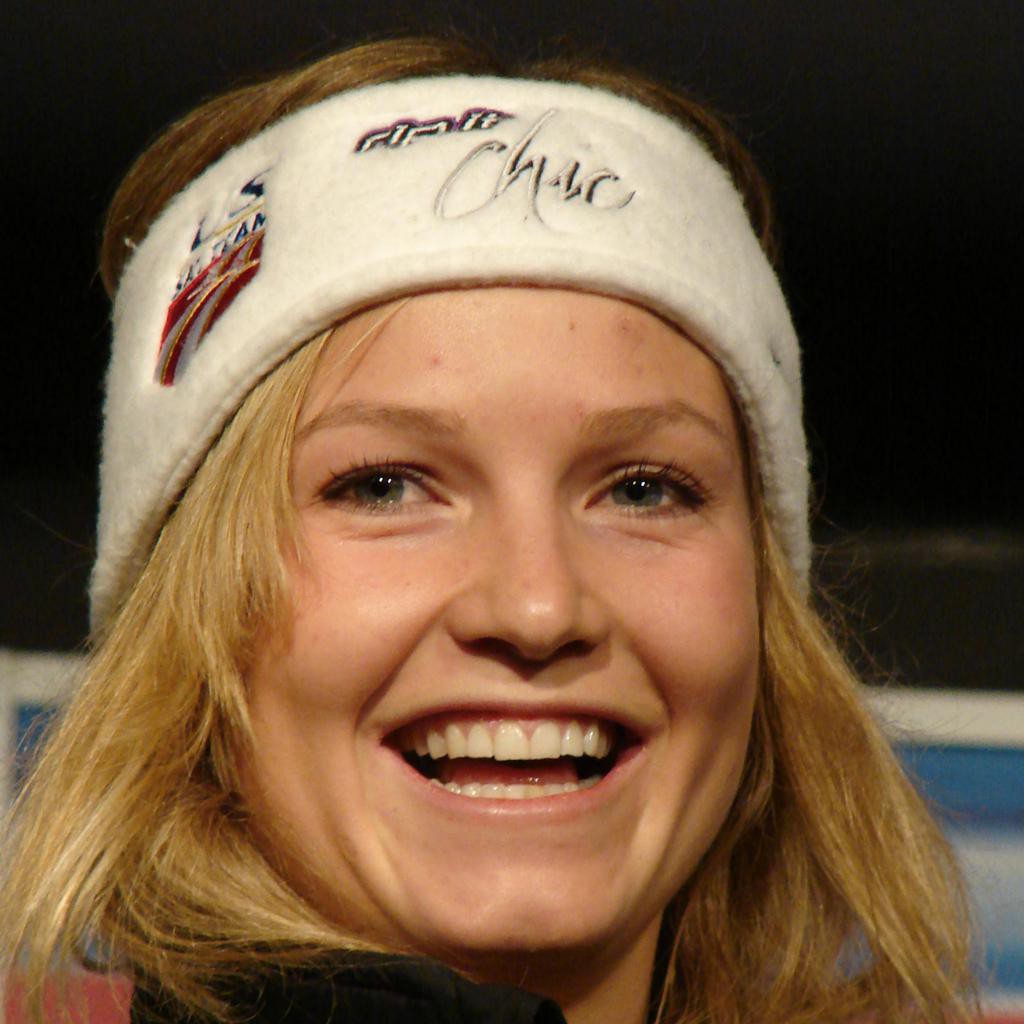} &
		\includegraphics[width=.135\linewidth]{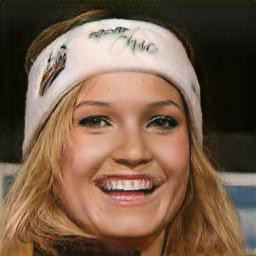} &
		\includegraphics[width=.135\linewidth]{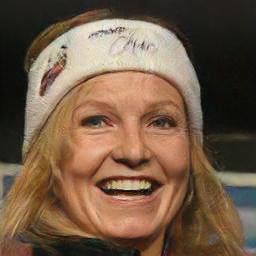} &
		\includegraphics[width=.135\linewidth]{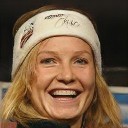} &
		\includegraphics[width=.135\linewidth]{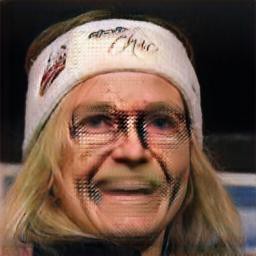} &
		\includegraphics[width=.135\linewidth]{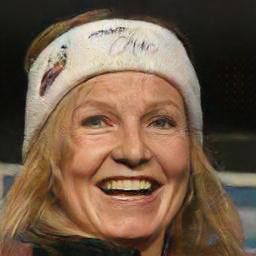} &
		\includegraphics[width=.135\linewidth]{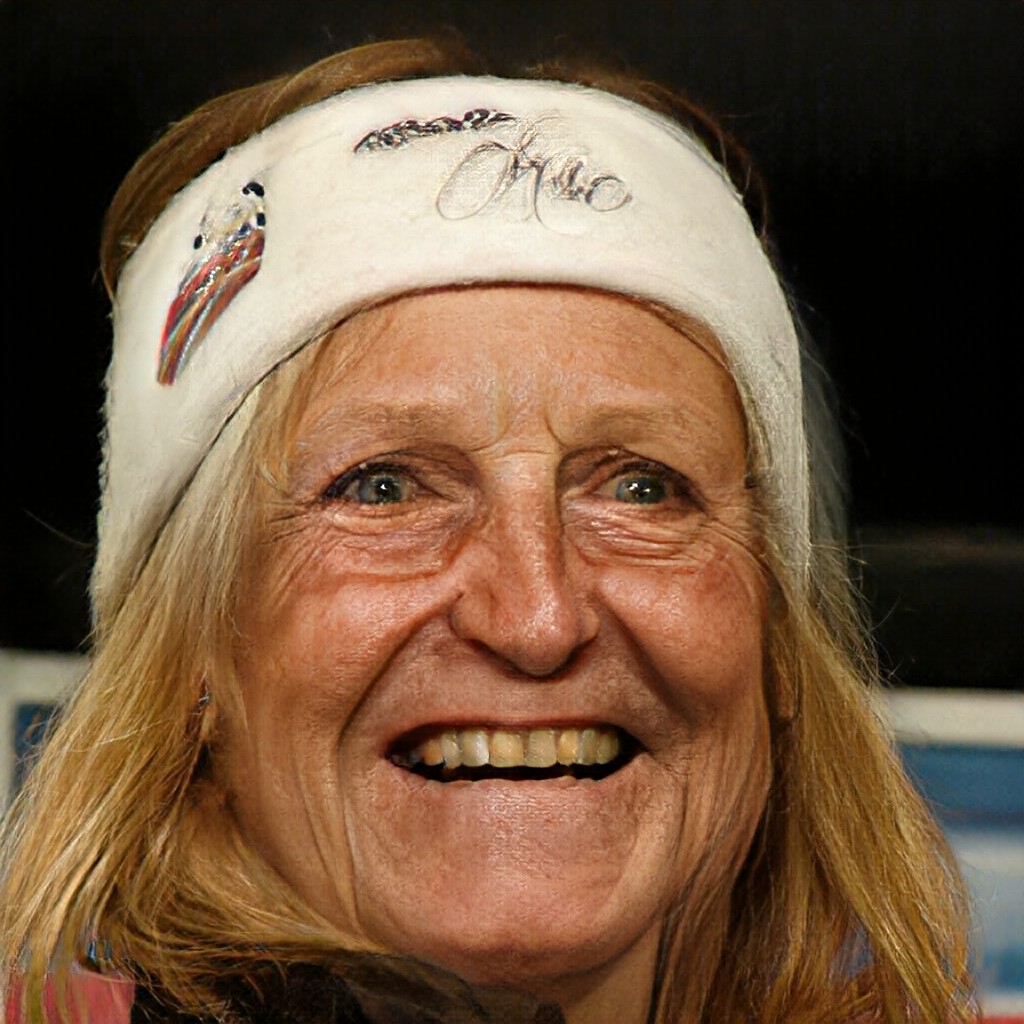} \\
	\end{tabular}\vspace{2mm}
	\caption{Qualitative evaluation with respect to state-of-the-art image-to-image face editing methods on 4 attributes, ``age'', ``bushy eyebrows'', ``eyeglasses'', and ``beard''. StarGAN* and AttGAN* represent the refined version using the cam-consistency loss~\cite{kim2021not}. Our method can produce high-resolution and semantically correct editing on these challenging cases. (Zoom in for better view.)
	}\vspace{-4mm}
	\label{fig:qualitative study}
\end{figure*}

The results are shown in Fig.~\ref{fig:compare with different encoders}. From the result we can see that all the encoding methods fail to retain the out-of-domain information. As for the first person wearing a blue hat and holding a fist, all encoders treat the hat as the hair and the fist as the background. Similar problem exists in the second person, in which his cap is inverted to hair and becomes white as he gets older. All these out-of-domain problems are addressed by our framework, regardless of their inversions and applied editings. The results show that our DA module and our deghosting network are encoder-independent and are robust enough to work with different types of inversion and editing methods.

\subsection{Comparison with SOTAs}
In order to prove the superiority of our model, we compare our model with other state-of-the-art facial attributes editing methods. Note that we do not make an additional comparison with StyleGAN inversion based editing method other than Fig.~\ref{fig:compare with different encoders}. This is due to that they would obviously fail on out-of-domain samples as the original GAN was not trained on them. To maintain fairness, here we mainly compare our results with those non-StyleGAN based image-to-image translation methods. In particular, we compare our model with StarGAN~\cite{choi2018stargan}, AttGAN~\cite{he2019attgan}, and STGAN~\cite{liu2019stgan}. Kim \etal ~\cite{kim2021not} propose to refine an image-to-image translation method by introducing a CAM-consistency loss to force the network to focus on attribute-relevant regions, and we also compare to the refined version of StarGAN and AttGAN
(denoted as StarGAN* and AttGAN*). All the results are generated with their official codes, except the refinement method of Kim \etal ~\cite{kim2021not} that is not publicly available and implemented by ourselves.

\textbf{Quantitative Evaluation.} To quantitatively compare our method with state-of-the-arts, we use the Fre´chet inception distance (FID)~\cite{heusel2017gans} and learned perceptual image patch similarity (LPIPS)~\cite{zhang2018unreasonable} metrics to measure the quality of the results. FID measures the distribution distance between the original image dataset and the manipulated dataset. We calculate FID metric for each model, and select 4 common attributes (``age'', ``bushy eyebrows'', ``eyeglasses'', and ``beard'') that can be modified by all of the models to generate the manipulated dataset. The final FID value of each model is obtained by averaging the FID values corresponding to each attribute. LPIPS metric measures the perceptual similarity between the two images. The smaller the value of LPIPS, the greater the similarity. We use it to evaluate non-edited region consistency.

The numerical results are shown in Table~\ref{tab:FID and LPIPS}. As can be seen, the proposed method shows the best FID and LPIPS scores among the competitors. This reveals that our model can better maintain the distribution of the original dataset, and a strong capability to preserve the image quality for non-edited regions.

\textbf{Qualitative Evaluation.} In order to demonstrate the superiority of our model, we conduct a qualitative study by contrasting the results generated from different models. Again, here we mainly compare with the image-to-image translation models. The results of editing 4 attributes by different models are shown in Fig.~\ref{fig:qualitative study}.
We can see that, the outputs of our model are of the highest quality compared to all other methods. Our outputs best change the attributes while retaining other irrelevant information. StarGAN and its refined version suffer from checkerboard-like artifacts. AttGAN* can achieve better editing than the original version, but it still produces blurry details and semantically incorrect editing (like the eyebrow on the left wrongly appears on the hair in the second example), indicating that using an additional mask-guided loss is not reliable for challenging cases. In contrast, our Diff-CAM mask driven framework obtains significantly preferable editing performance, not to mention the high-resolution features provided by StyleGAN-based editing.
\begin{table}[t]
	\centering
	{
		\begin{tabular}{ccccccc}
			\toprule
			& StarGAN & AttGAN & STGAN & StarGAN* & AttGAN* & Ours \\
			\midrule
			FID$\downarrow$ & 30.98 & 17.96 & 20.97 & 28.52 & 15.72 & \textbf{13.76} \\
			LPIPS$\downarrow$ & 0.208 & 0.107 & 0.178 & 0.138 & 0.099 & \textbf{0.094}\\
			\bottomrule
	\end{tabular}}\vspace{2mm}
	\caption{Quantitative comparison with state-of-the-art face editing methods. StarGAN* and AttGAN* represent the refined version using the cam-consistency loss~\cite{kim2021not}. Our method achieves the best numerical performance.
	}\vspace{-4mm}
	\label{tab:FID and LPIPS}
\end{table}

\subsection{Editing on non-facial attributes and domains}
\begin{figure}[t]\vspace{-2mm}
	\centering
	\rotatebox[origin=l]{90}{Car}	
	\captionsetup[subfigure]{labelformat=empty}	
	\begin{subfigure}{0.09\linewidth}
		\centering			
		\includegraphics[width=\linewidth]{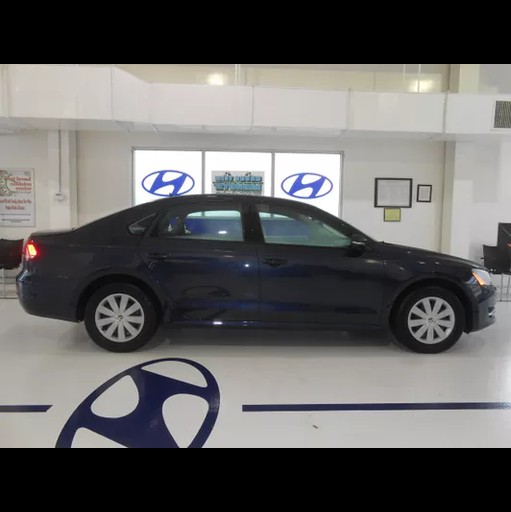}\\
		\includegraphics[width=\linewidth]{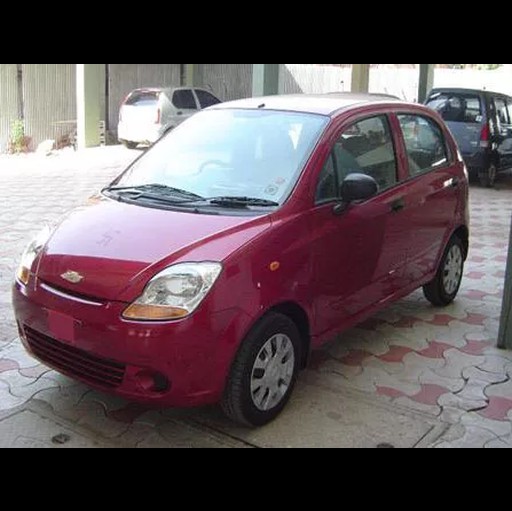}\\	
		\caption{Ori.}
	\end{subfigure}
	\hspace{-1.5mm}
	\begin{subfigure}{0.09\linewidth}
		\centering			
		\includegraphics[width=\linewidth]{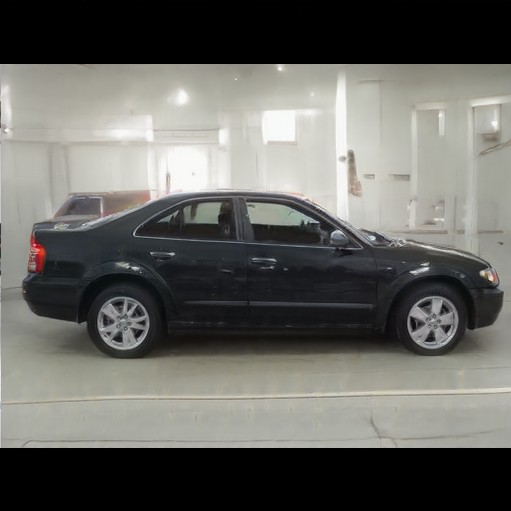}\\
		\includegraphics[width=\linewidth]{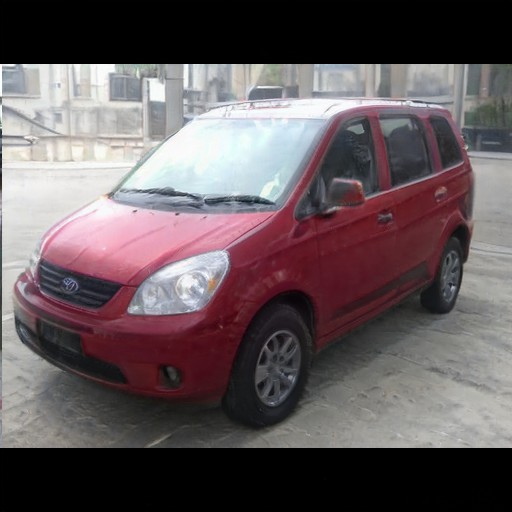}\\
		\caption{Inv.}
	\end{subfigure}
	\hspace{-1.5mm}
	\begin{subfigure}{0.09\linewidth}
		\centering			
		\includegraphics[width=\linewidth]{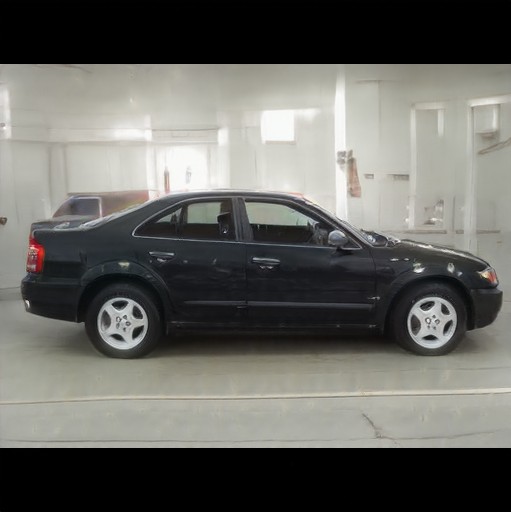}\\
		\includegraphics[width=\linewidth]{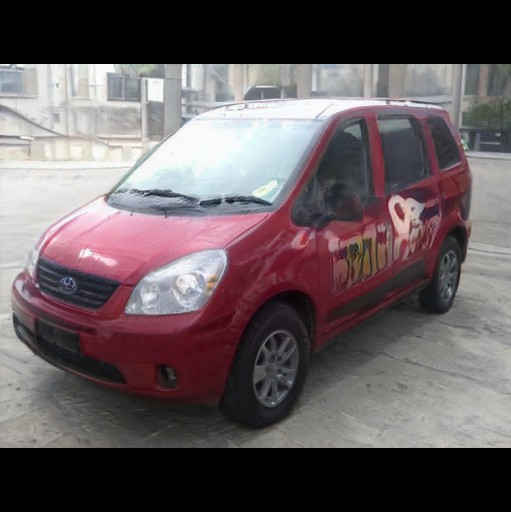}\\
		\caption{Edit.}
	\end{subfigure}
	\hspace{-1.5mm}
	\begin{subfigure}{0.09\linewidth}
		\centering			
		\includegraphics[width=\linewidth]{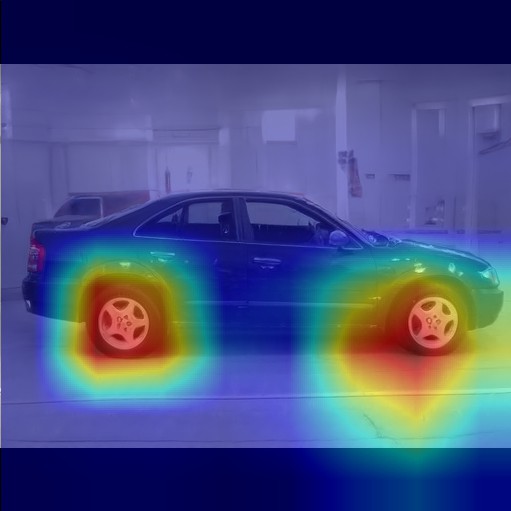}\\
		\includegraphics[width=\linewidth]{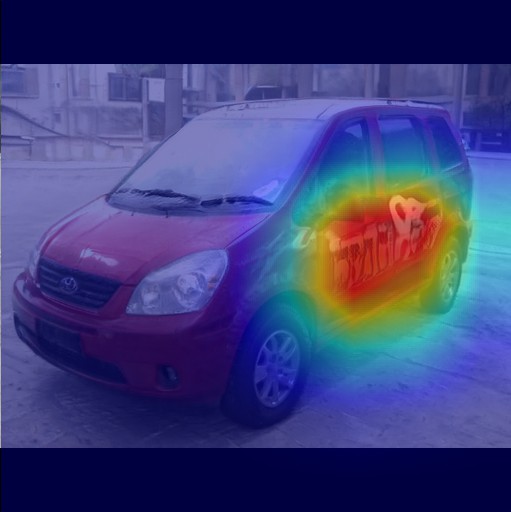}\\
		\caption{Diff.}
	\end{subfigure}
	\hspace{-1.5mm}
	\begin{subfigure}{0.09\linewidth}
		\centering			
		\includegraphics[width=\linewidth]{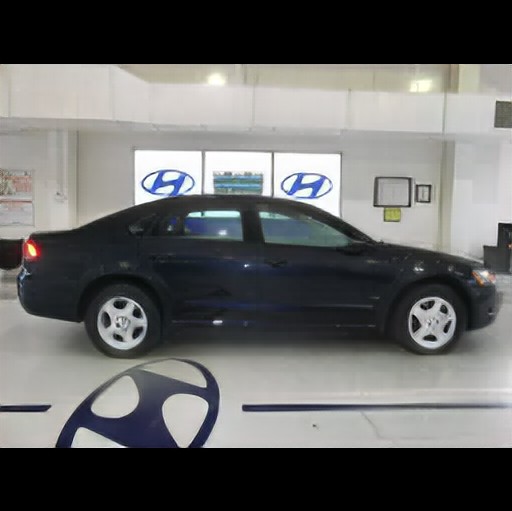}\\
		\includegraphics[width=\linewidth]{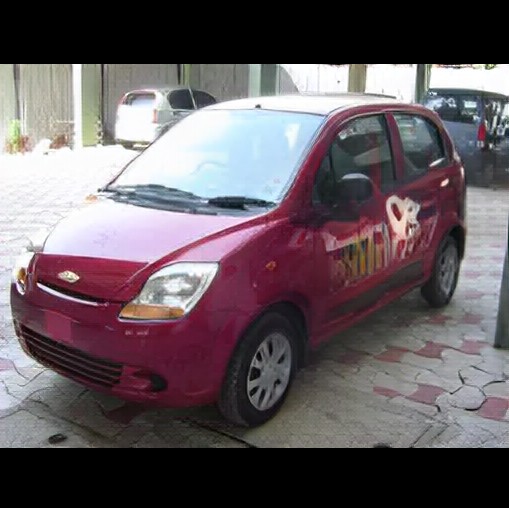}\\
		\caption{Ours}
	\end{subfigure}
	\rotatebox[origin=l]{90}{\hspace{-10mm}..............................}
	\begin{subfigure}{0.09\linewidth}
		\centering			
		\includegraphics[width=\linewidth]{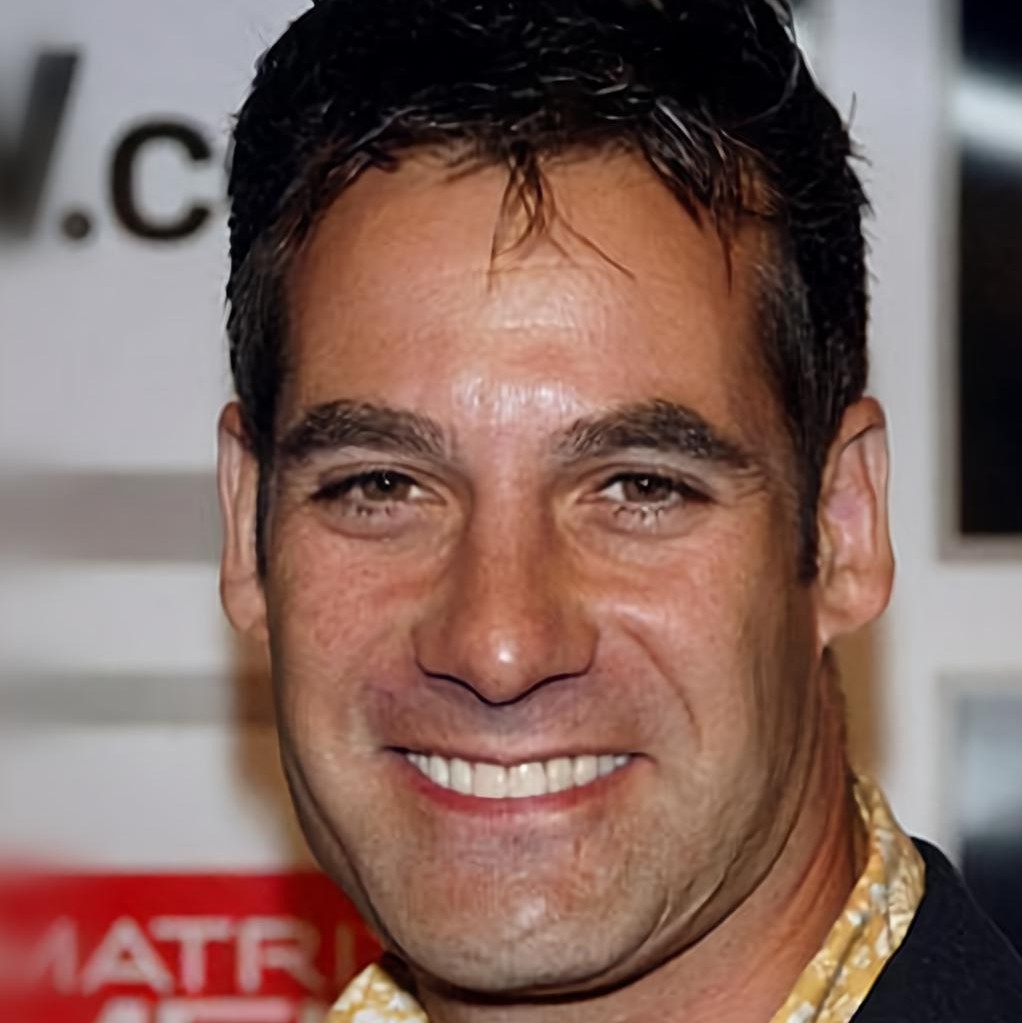}\\
		\includegraphics[width=\linewidth]{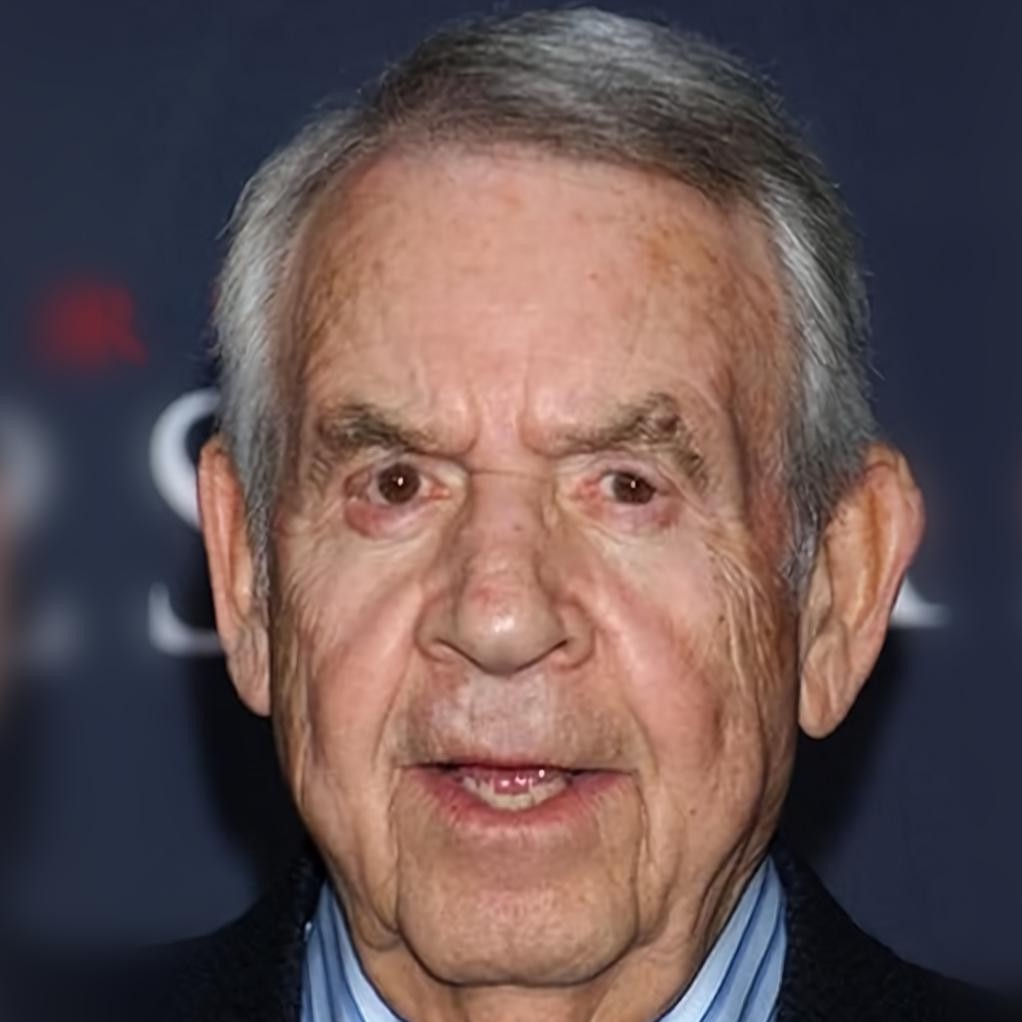}\\
		\caption{Ori.}
	\end{subfigure}
	\hspace{-1.5mm}
	\begin{subfigure}{0.09\linewidth}
		\centering			
		\includegraphics[width=\linewidth]{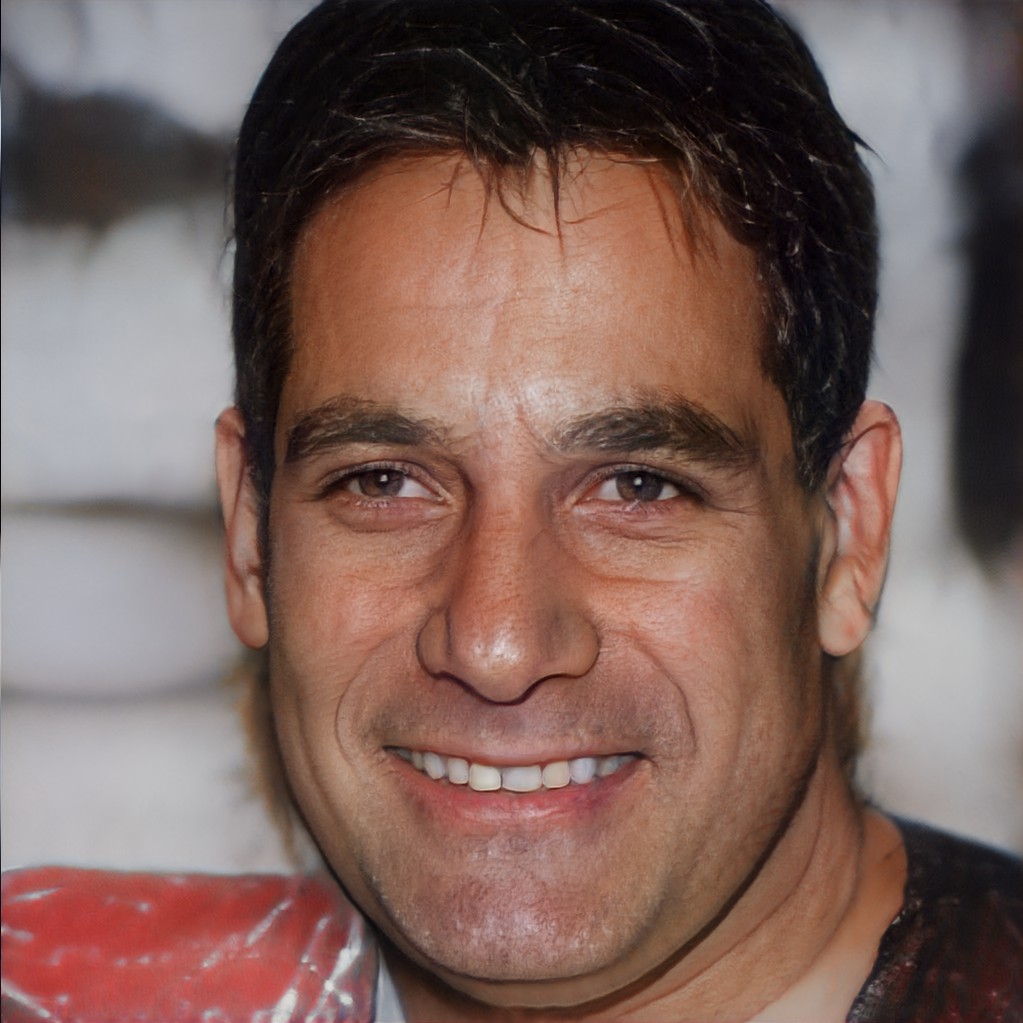}\\
		\includegraphics[width=\linewidth]{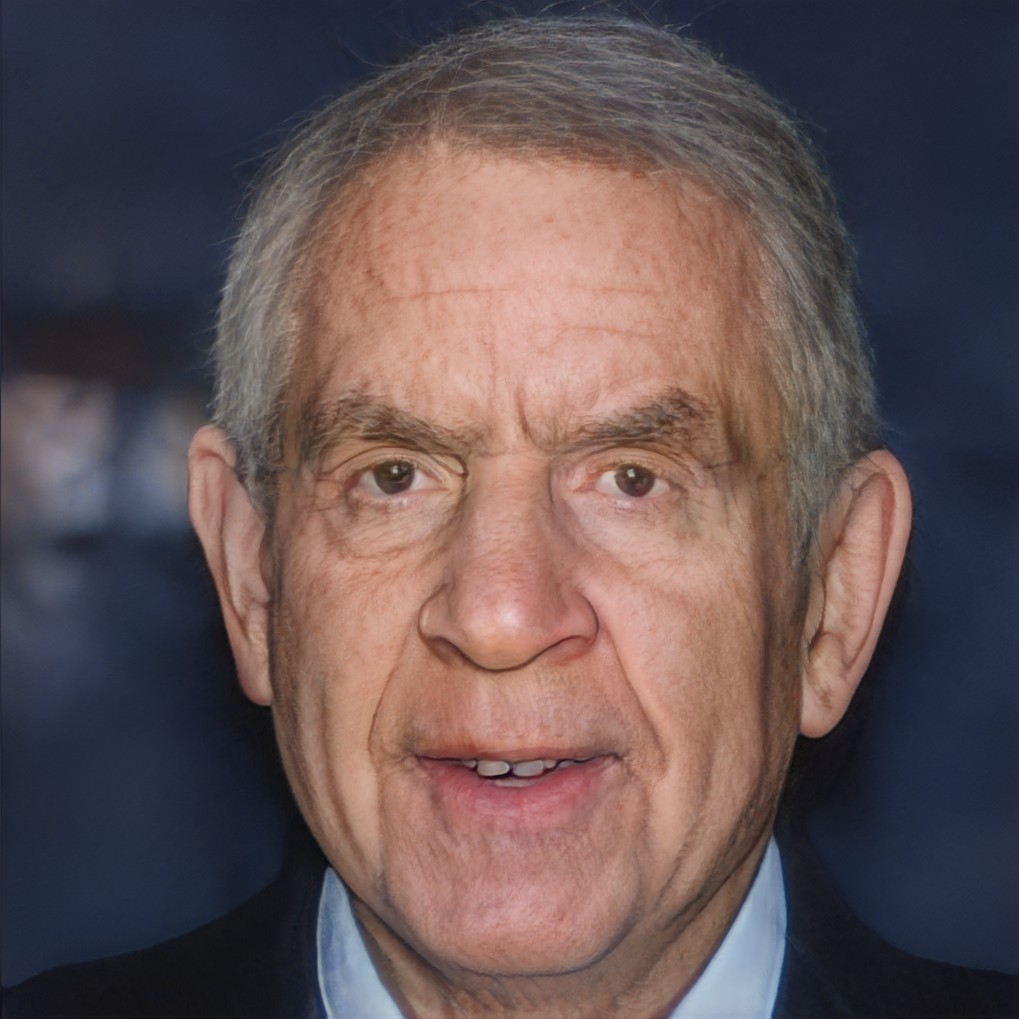}\\    	
		\caption{Inv.}
	\end{subfigure}
	\hspace{-1.5mm}
	\begin{subfigure}{0.09\linewidth}
		\centering			
		\includegraphics[width=\linewidth]{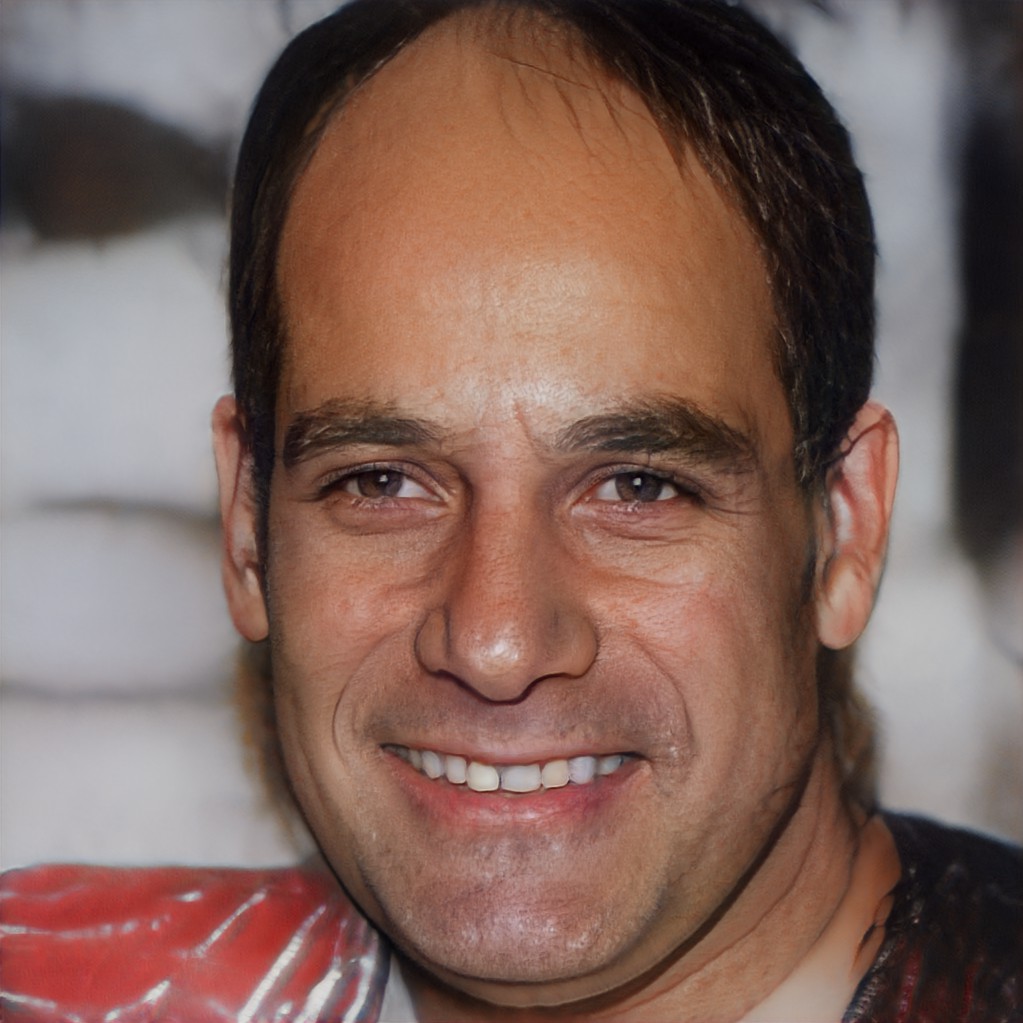}\\
		\includegraphics[width=\linewidth]{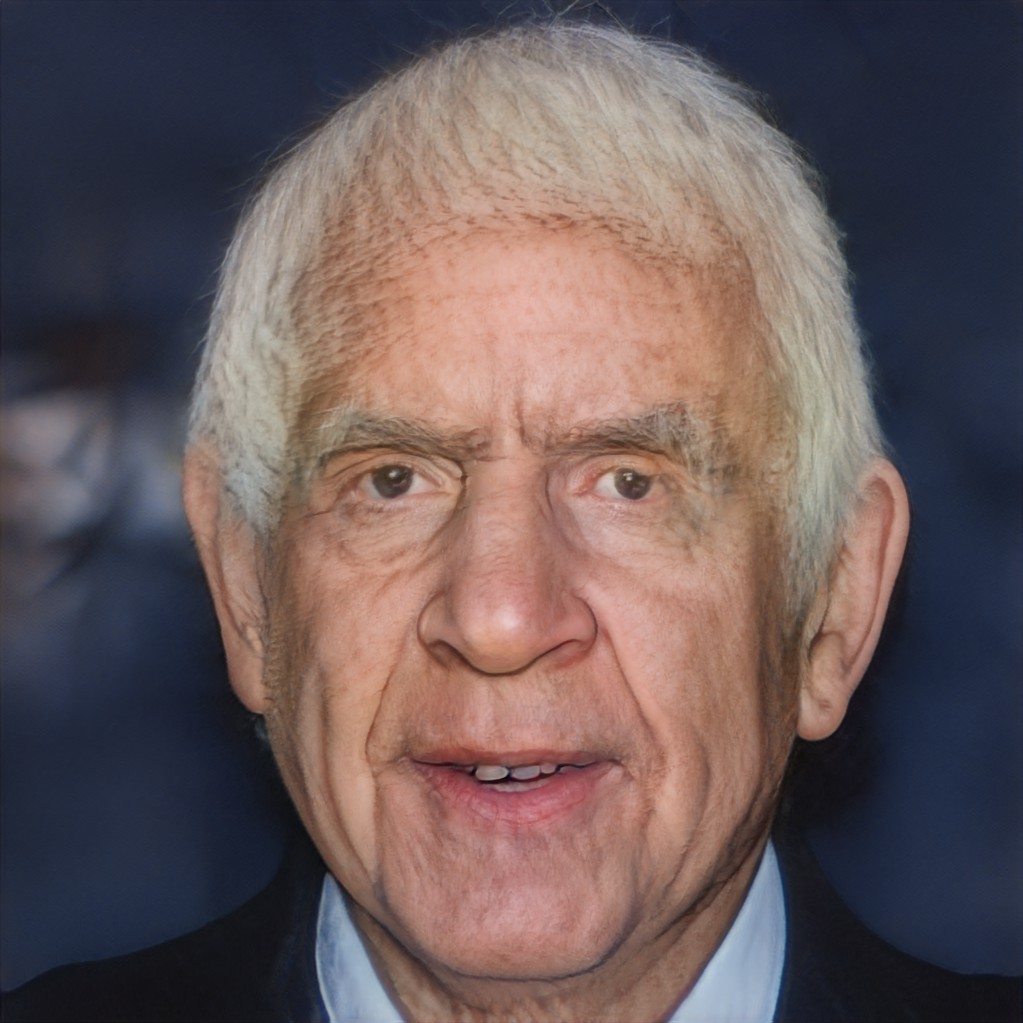}\\   	
		\caption{Edit.}
	\end{subfigure}
	\hspace{-1.5mm}
	\begin{subfigure}{0.09\linewidth}
		\centering			
		\includegraphics[width=\linewidth]{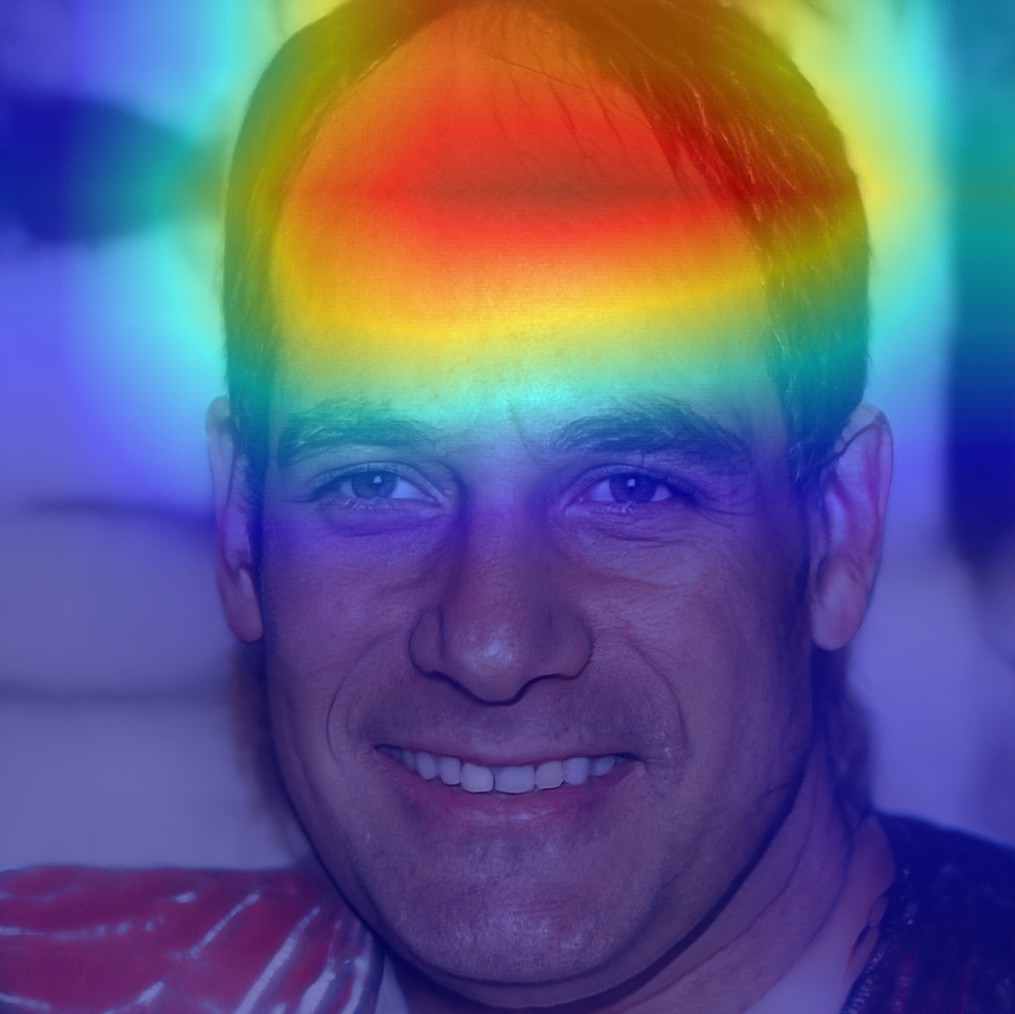}\\
		\includegraphics[width=\linewidth]{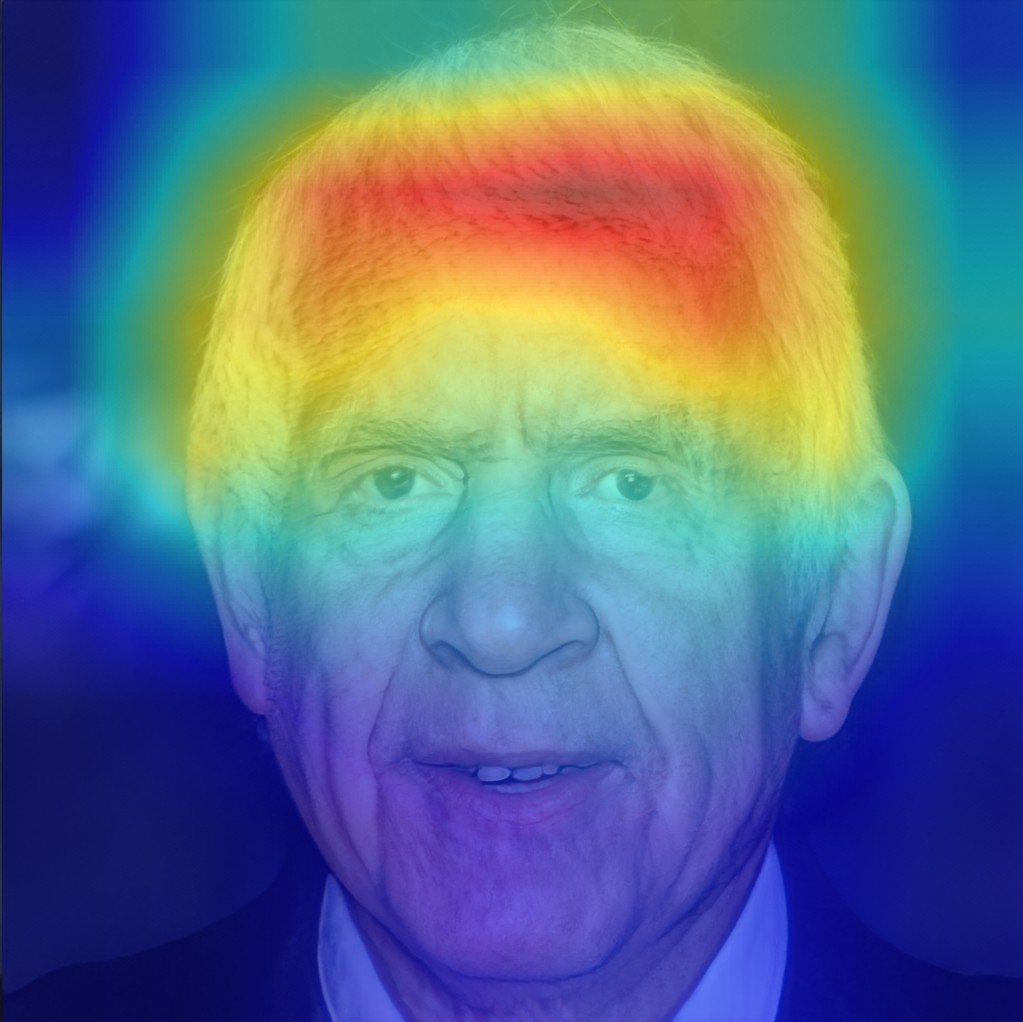}\\	
		\caption{Diff.}
	\end{subfigure}
	\hspace{-1.5mm}
	\begin{subfigure}{0.09\linewidth}
		\centering			
		\includegraphics[width=\linewidth]{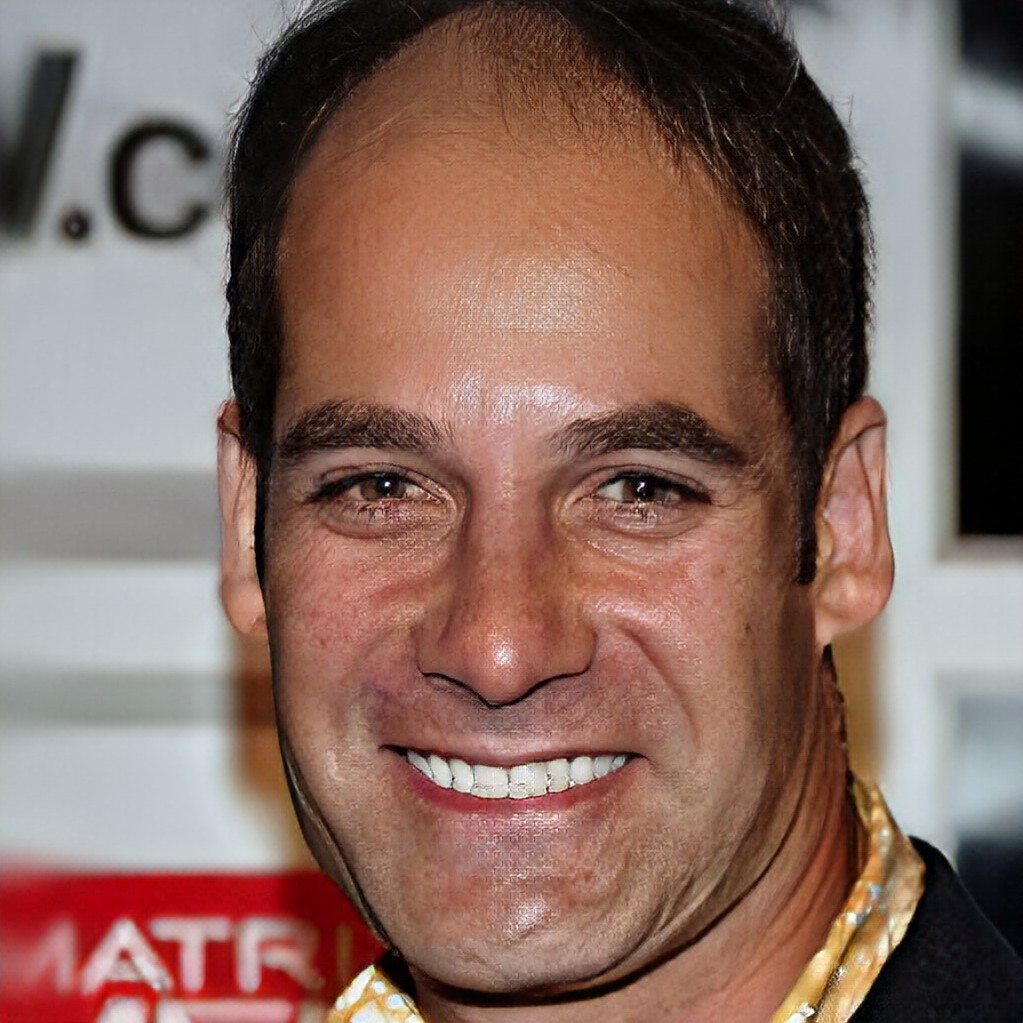}\\
		\includegraphics[width=\linewidth]{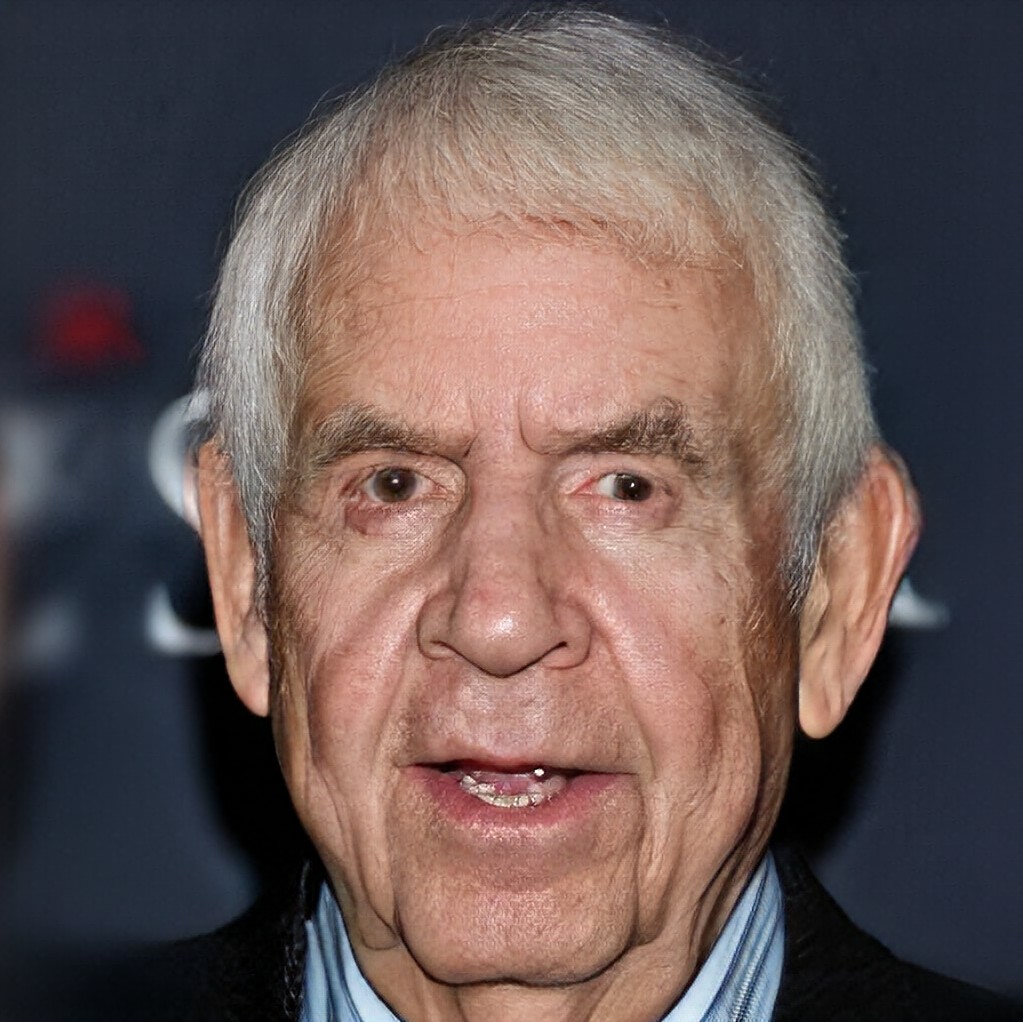}\\  	
		\caption{Ours}
	\end{subfigure}
	\rotatebox[origin=l]{90}{\hspace{-8mm}\tiny Hair color Hairstyle}	
	\vspace{-2mm}\caption{Evaluation on ``car'' domain and non-facial attributes ``hairstyle'', ``hair color''. (Zoom in for better view.)}
	\label{fig}\vspace{-6mm}
\end{figure}

We also verify the generalization ability of our model and display the qualitative results in Fig.~\ref{fig}. In particular,   regarding non-facial attributes (like ``hair color'' or ``hairstyle'') and other domains (like ``car''), the modifications may no longer happen in the center region like most of those occured in facial attributes editing, \eg, logo, wheels, and hair. Nevertheless, our Diff-CAM can always precisely distinguish the edited and non-edited areas. And our framework can thus be robust to small regions and different attributes for editing.

\subsection{Multi-attribute Editing}
\begin{figure}[t]
	\centering
	\begin{subfigure}[t]{0.22\linewidth}
		\centering
		\includegraphics[width=\textwidth]{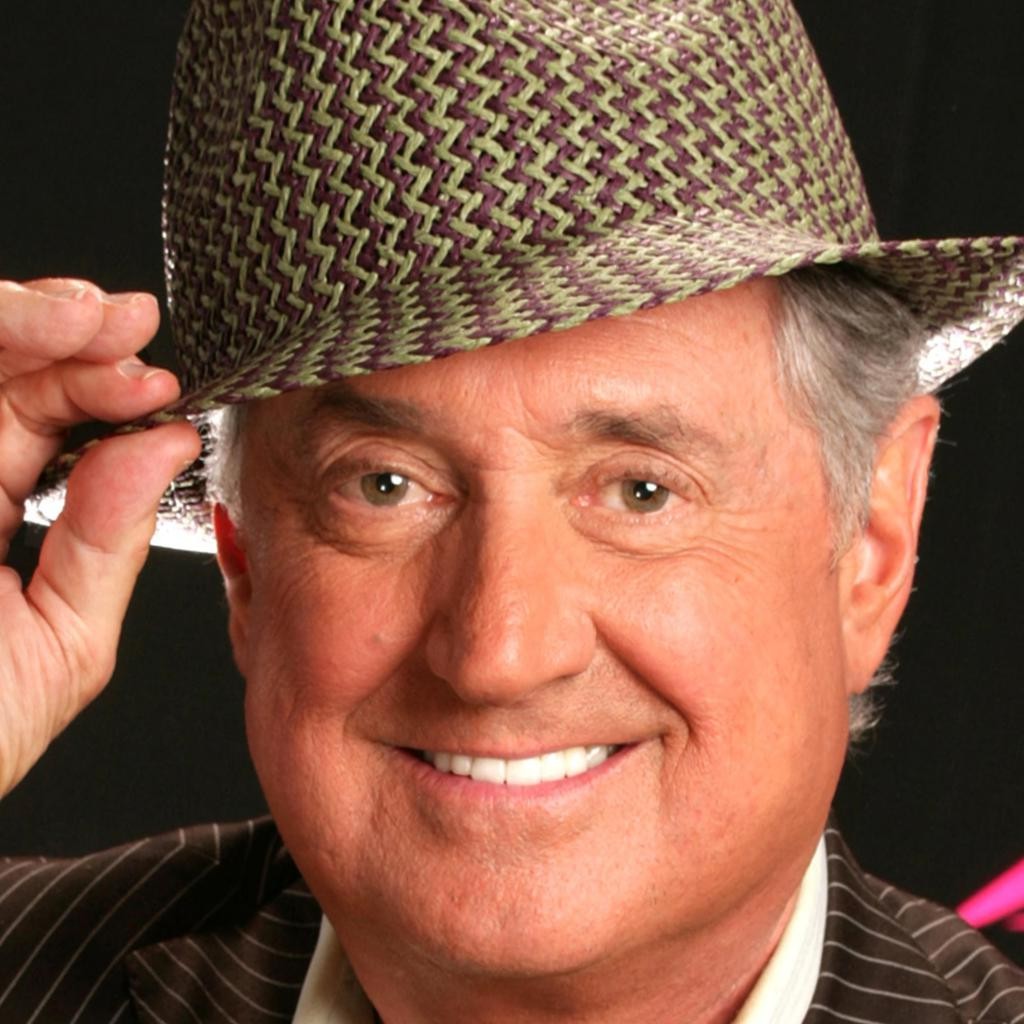}\vspace{1mm}
		\includegraphics[width=\textwidth]{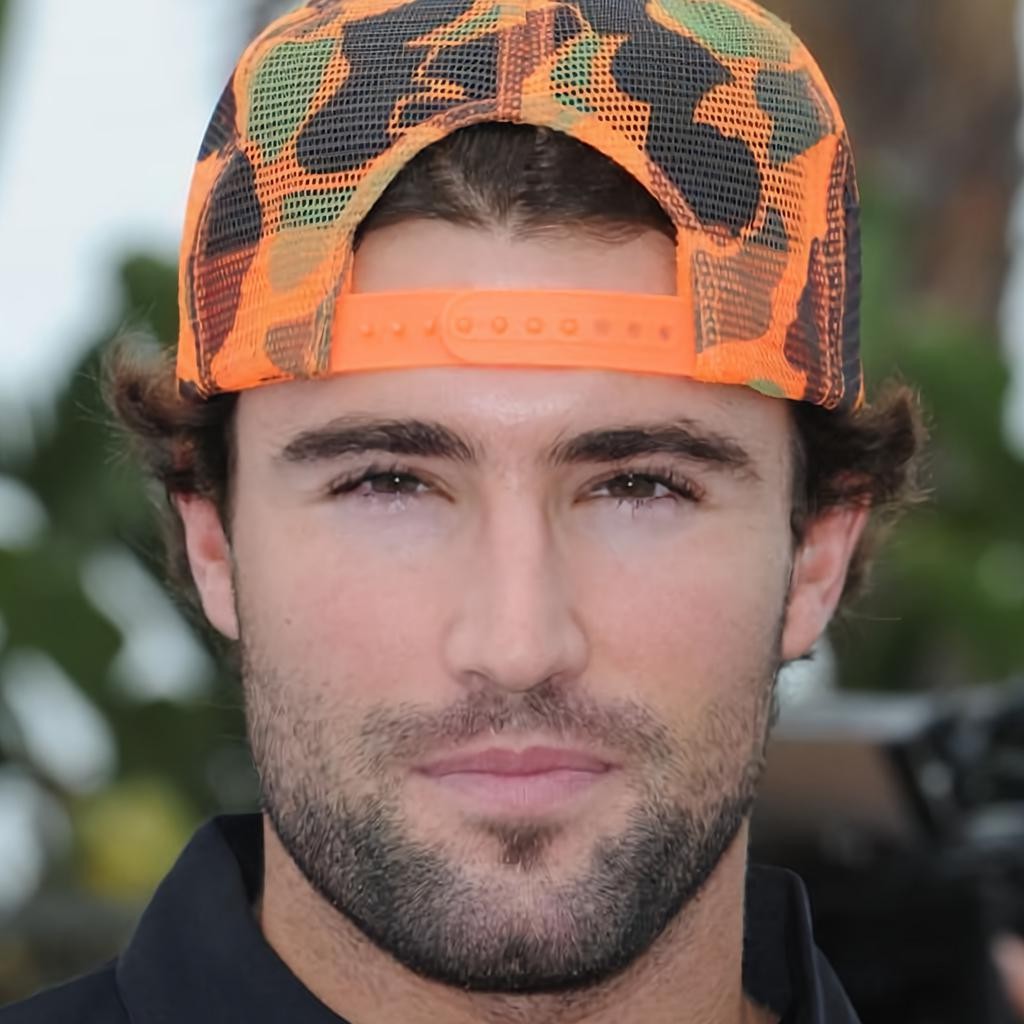}
		\caption{Input}
	\end{subfigure}
	\begin{subfigure}[t]{0.22\linewidth}
		\centering
		\includegraphics[width=\textwidth]{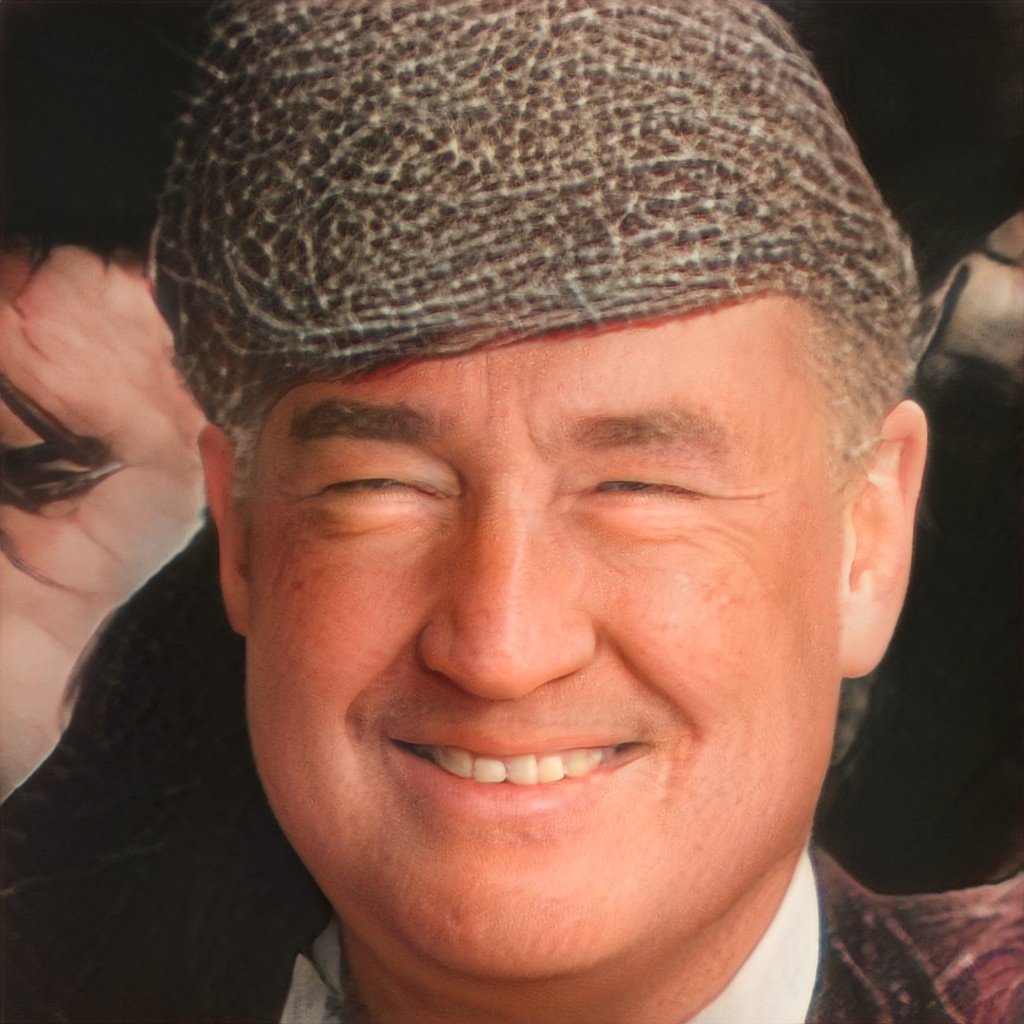}\vspace{1mm}
		\includegraphics[width=\textwidth]{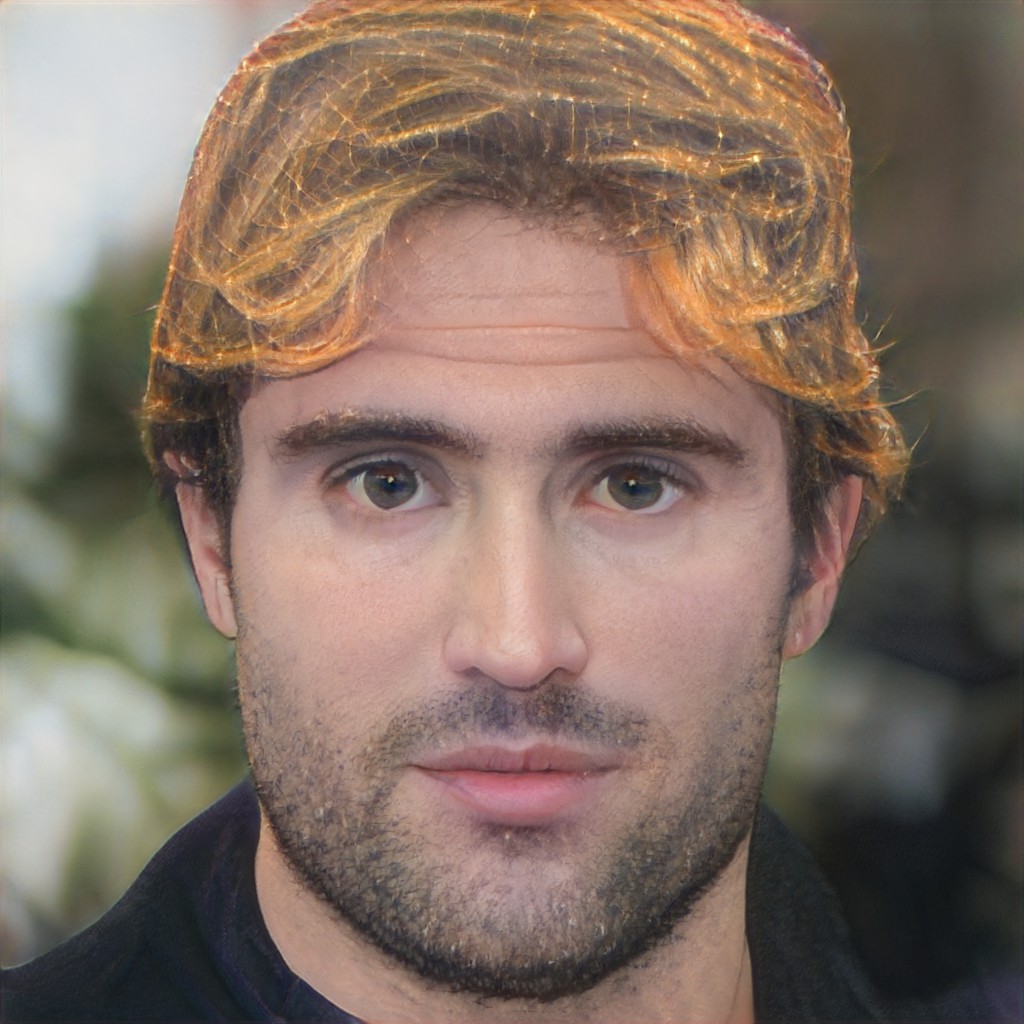}
		\caption{``Eyes open''}
	\end{subfigure}
	\begin{subfigure}[t]{0.22\linewidth}
		\centering
		\includegraphics[width=\textwidth]{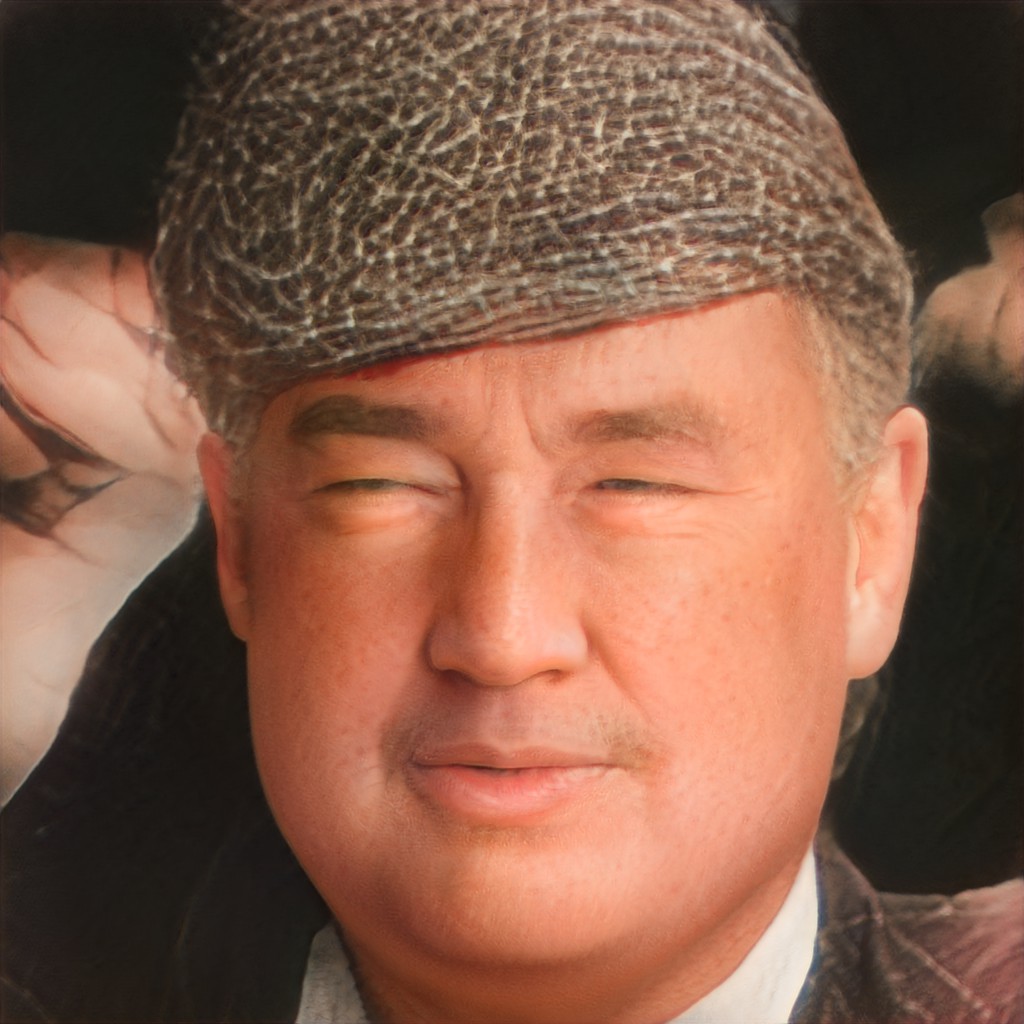}\vspace{1mm}
		\includegraphics[width=\textwidth]{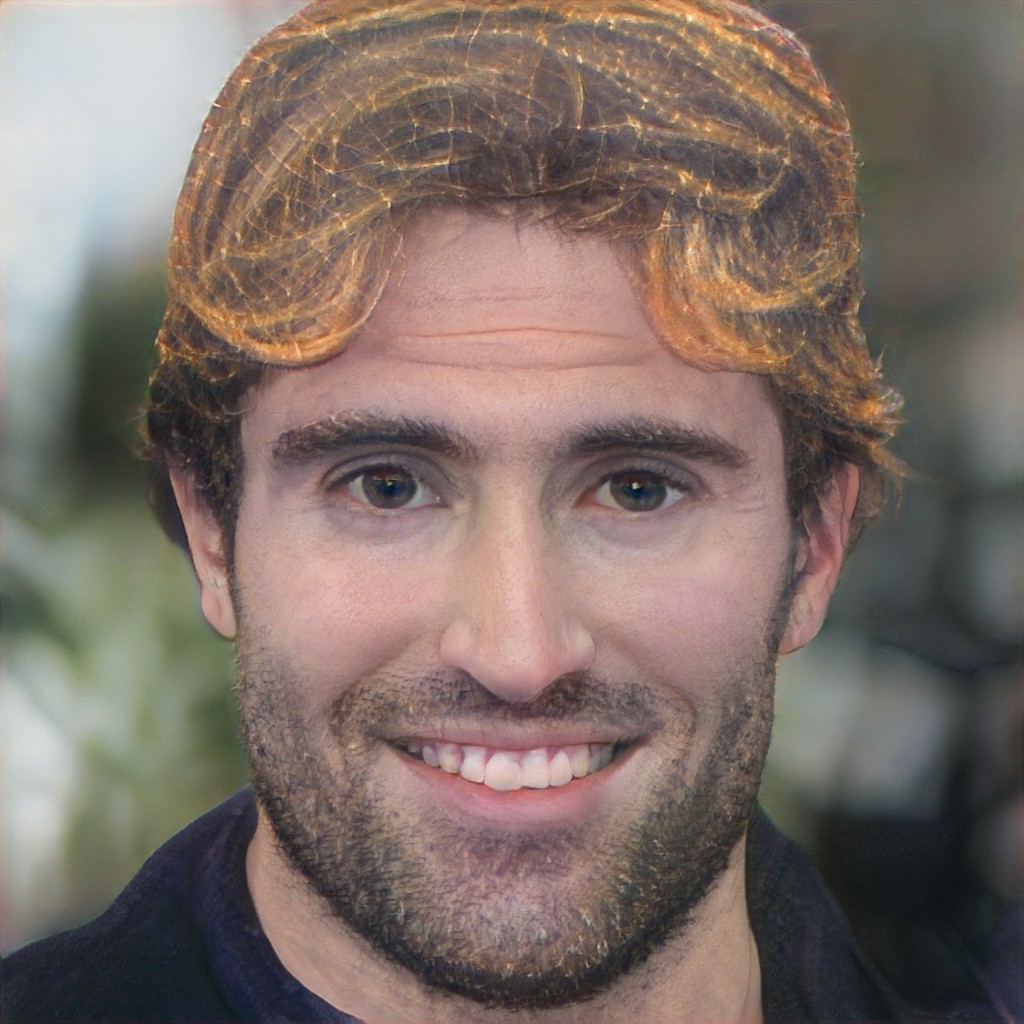}
		\caption{``Smile''}
	\end{subfigure}
	\begin{subfigure}[t]{0.22\linewidth}
		\centering
		\includegraphics[width=\textwidth]{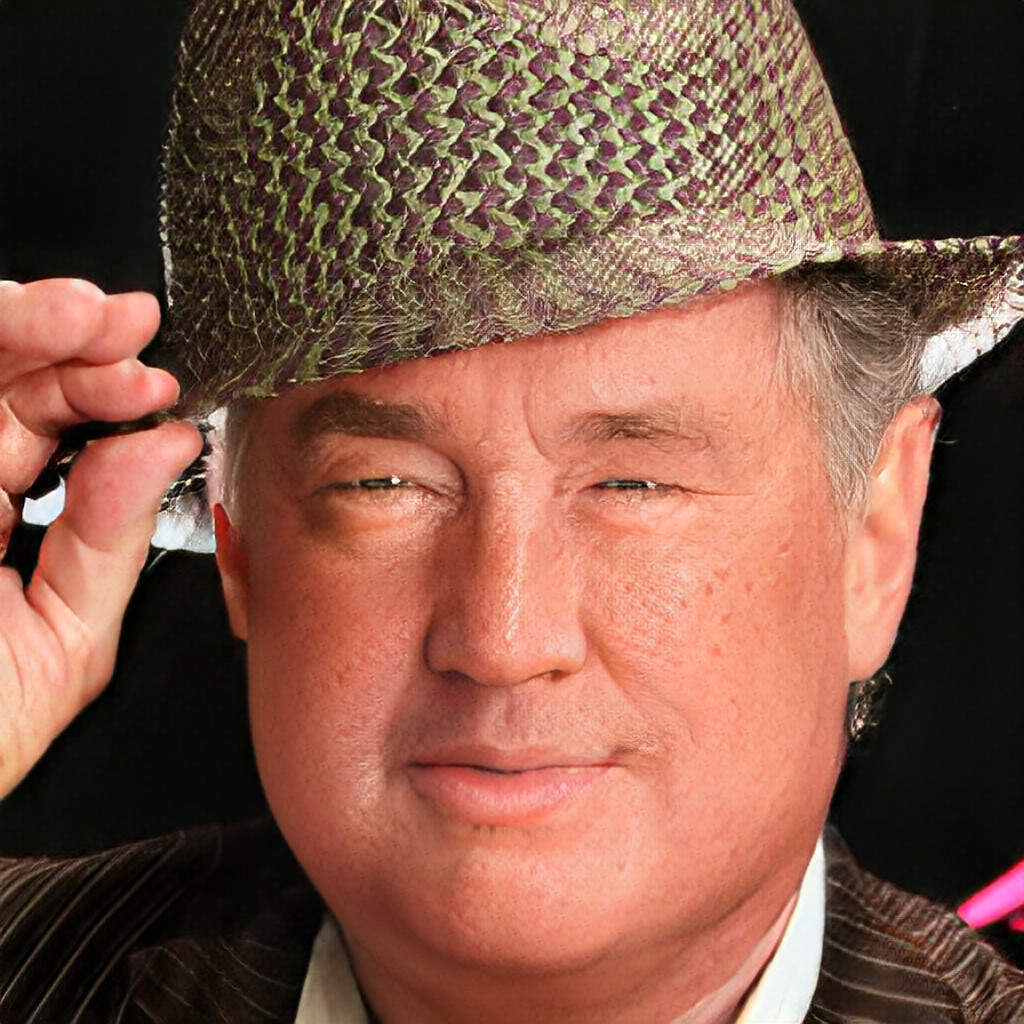}\vspace{1mm}
		\includegraphics[width=\textwidth]{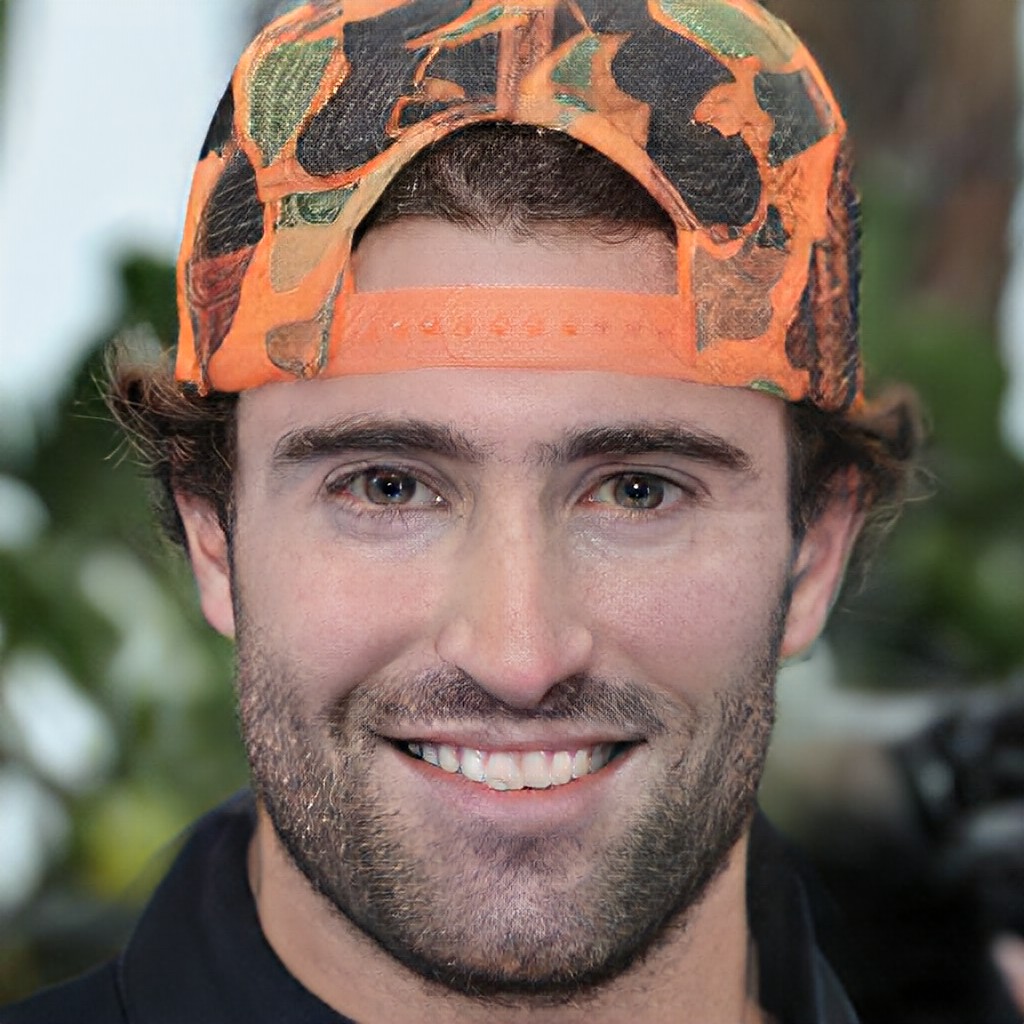}
		\caption{Ours}
	\end{subfigure}
	\vspace{-2mm}\caption{Multi-attribute editing results. Our method can successfully edit multiple attributes one by one, while still retaining the out-of-domain regions.}\vspace{-2mm}
	\label{fig:multi-attributes editing}
\end{figure}

In addition to the single-attribute editing, our model shows its high degree of flexibility by also supporting editing multiple attributes. The process of modifying multiple attributes is completed by modifying the attributes one by one as is described in Sec.~\ref{method:multi}. Fig.~\ref{fig:multi-attributes editing} shows two examples of editing two attributes, ``eyes open'' and ``smile''. The final outputs of our model successfully introduce the changes involved in the two editing steps and also manage to maintain the uninvertible information such as the hats and the fingers.

\subsection{Limitation}
\begin{figure}[t]
	\centering
	\rotatebox[origin=l]{90}{\hspace{-5mm}Thickness}	
	\captionsetup[subfigure]{labelformat=empty}	
	\begin{subfigure}{0.11\linewidth}
		\centering			
		\includegraphics[width=\textwidth]{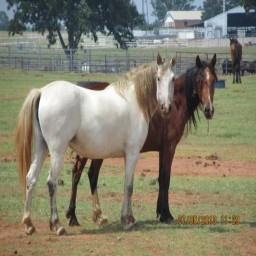}\\		
		\caption{Ori.}
	\end{subfigure}
	\begin{subfigure}{0.11\linewidth}
		\centering			
		\includegraphics[width=\textwidth]{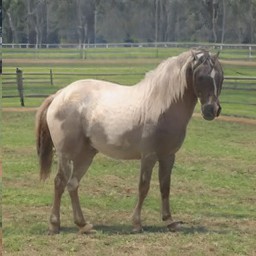}\\	
		\caption{Inv.}
	\end{subfigure}
	\begin{subfigure}{0.11\linewidth}
		\centering			
		\includegraphics[width=\textwidth]{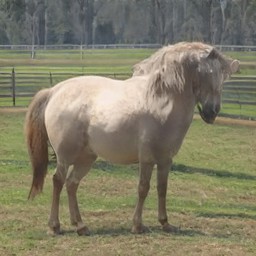}\\		
		\caption{Edit.}
	\end{subfigure}
	\begin{subfigure}{0.11\linewidth}
		\centering			
		\includegraphics[width=\textwidth]{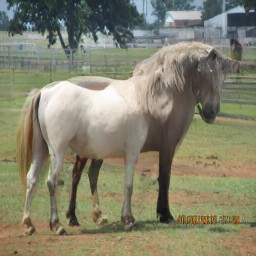}\\		
		\caption{Ours}
	\end{subfigure}
	\rotatebox[origin=l]{90}{\hspace{-6mm}.....................}
	\begin{subfigure}{0.11\linewidth}
		\centering			
		\includegraphics[width=\textwidth]{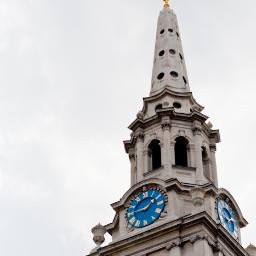}\\	
		\caption{Ori.}
	\end{subfigure}	
	\begin{subfigure}{0.11\linewidth}
		\centering			
		\includegraphics[width=\textwidth]{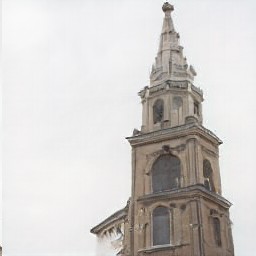}\\		
		\caption{Inv.}
	\end{subfigure}
	\hspace{-1.5mm}
	\begin{subfigure}{0.11\linewidth}
		\centering			
		\includegraphics[width=\textwidth]{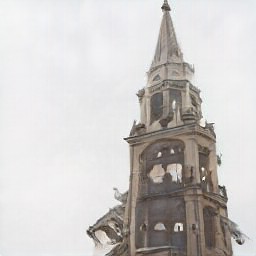}\\  	
		\caption{Edit.}
	\end{subfigure}
	\begin{subfigure}{0.11\linewidth}
		\centering			
		\includegraphics[width=\textwidth]{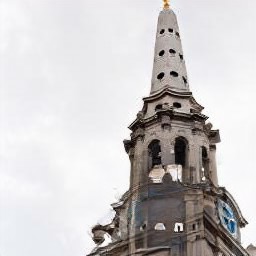}\\	
		\caption{Ours}
	\end{subfigure}
	\rotatebox[origin=l]{90}{\hspace{-3mm}Window}	
	\vspace{-2mm}\caption{Challenging cases of our model. The two examples from different domains show the editing results when the inversion cannot faithfully reconstruct the original image.}\vspace{-2mm}
   \label{fig:challenging cases}
\end{figure}

Although our model has achieved promising performance on the editing of facial or non-facial attributes and other domains, its ability to handle attribute changing is not unlimited. Fig.~\ref{fig:challenging cases} shows the examples of our limitation. Our model is heavily relied on the performance of GAN inversion methods and only introduces changes covered by the mask from the DA module. Therefore, if the inverted result cannot faithfully reconstruct the original image which is likely occurred in the case of non-human domains, serious distortion and ghosting artifacts will be existed in the final result.

%
%
%
%
%

\section{Conclusion}
In this paper, we propose a novel GAN prior based editing technique to tackle the out-of-domain inversion problem with a composition-decomposition paradigm. 
We introduce a differential activation mechanism to track semantic changes before and after editing. With the aid of the calculated Diff-CAM mask, a coarse reconstruction can be obtained by the composition of the edited image and the original input. We further present a deghosting network to mitigate the ghosting effect in the coarse result. 
Both qualitative and quantitative evaluations validate the superiority of our method.


\section*{Acknowledgement}
\vspace{-4mm}This project is supported by the National Natural Science Foundation of China (62102381, U1706218, 41927805, 61972162); Shandong Natural Science Foundation (ZR2021QF035); Fundamental Research Funds for the Central Universities (202113035); the National Key R\&D Program of China (2018AAA0100600); the China Postdoctoral Science Foundation (2020M682240, 2021T140631);  Guangdong International Science and Technology Cooperation Project (No. 2021A0505030009); Guangdong Natural Science Foundation (2021A1515012625); Guangzhou Basic and Applied Research Project (202102021074); and CCF-Tencent Open Research fund (RAGR20210114).

\clearpage
%
%
\bibliographystyle{splncs04}
\bibliography{outofdomain}
\end{document}